\documentclass[sigconf]{acmart}

\AtBeginDocument{%
  }

\setcopyright{acmlicensed}
\copyrightyear{2024}
\acmYear{2024}
\acmDOI{XXXXXXX.XXXXXXX}

\acmConference[RelKD 2024]
    {The Second International Workshop on Resource-Efficient Learning for Knowledge Discovery}
    {August 25--29, 2024}
    {Barcelona, Spain}
\acmISBN{978-1-4503-XXXX-X/18/06}

\usepackage[utf8]{inputenc} % allow utf-8 input
\usepackage[T1]{fontenc}    % use 8-bit T1 fonts
\usepackage{url}            % simple URL typesetting
\usepackage{booktabs}       % professional-quality tables
\usepackage{nicefrac}       % compact symbols for 1/2, etc.
\usepackage{microtype}      % microtypography

\usepackage{graphicx}
\usepackage[skip=1ex]{caption}
\usepackage{subcaption}
\usepackage{float}
\begin{document}

\title[Data-Centric Approach to Constrained Machine Learning]{Data-Centric Approach to Constrained Machine Learning:\\A Case Study on Conway's Game of Life}


\author{Anton Bibin}
\authornote{Equal contribution. Authors listed alphabetically.}
\affiliation{%
    \institution{Skoltech}
    \department{Skoltech Agro}
    \city{Moscow}
    \country{Russia}
}
\email{a.bibin@skoltech.ru}

\author{Anton Dereventsov}
\authornotemark[1]
\affiliation{%
    \institution{Lirio LLC}
    \department{Behavioral Reinforcement Learning Lab}
    \city{Knoxville}
    \state{TN}
    \country{USA}
}
\email{adereventsov@lirio.com}

%\renewcommand{\shortauthors}{Bibin et al.}

\begin{abstract}
This paper focuses on a data-centric approach to machine learning applications in the context of Conway's Game of Life.
Specifically, we consider the task of training a minimal architecture network to learn the transition rules of Game of Life for a given number of steps ahead, which is known to be challenging due to restrictions on the allowed number of trainable parameters.
An extensive quantitative analysis showcases the benefits of utilizing a strategically designed training dataset, with its advantages persisting regardless of other parameters of the learning configuration, such as network initialization weights or optimization algorithm.
Importantly, our findings highlight the integral role of domain expert insights in creating effective machine learning applications for constrained real-world scenarios.
\end{abstract}

\begin{CCSXML}
<ccs2012>
   <concept>
       <concept_id>10010147.10010257.10010258.10010259.10010264</concept_id>
       <concept_desc>Computing methodologies~Supervised learning by regression</concept_desc>
       <concept_significance>500</concept_significance>
       </concept>
   <concept>
       <concept_id>10010147.10010178.10010205.10010208</concept_id>
       <concept_desc>Computing methodologies~Continuous space search</concept_desc>
       <concept_significance>500</concept_significance>
       </concept>
   <concept>
       <concept_id>10003752.10003809.10010031.10010032</concept_id>
       <concept_desc>Theory of computation~Pattern matching</concept_desc>
       <concept_significance>300</concept_significance>
       </concept>
   <concept>
       <concept_id>10003752.10003809.10003716.10011138.10011140</concept_id>
       <concept_desc>Theory of computation~Nonconvex optimization</concept_desc>
       <concept_significance>100</concept_significance>
       </concept>
   <concept>
       <concept_id>10002951.10002952.10003219.10003215</concept_id>
       <concept_desc>Information systems~Extraction, transformation and loading</concept_desc>
       <concept_significance>100</concept_significance>
       </concept>
 </ccs2012>
\end{CCSXML}

\ccsdesc[500]{Computing methodologies~Supervised learning by regression}
\ccsdesc[500]{Computing methodologies~Continuous space search}
\ccsdesc[300]{Theory of computation~Pattern matching}
\ccsdesc[100]{Theory of computation~Nonconvex optimization}
\ccsdesc[100]{Information systems~Extraction, transformation and loading}

\keywords{Data-Centric ML, Supervised Learning, Data Design, Game of Life}



\maketitle

%%%%%%%%%%%%%%%%%%%%%%%%%%%%%%%%%%%%%%%%%%%%%%%%%%%%%%%%%%%%%%%%%%%%%%%%%%%%%%%%%%%%%%%%%%%%%%%%%%%
\section{Introduction}
In this work we consider the problem of learning the rules of Conway's Game of Life, posed as an image-to-image translation task.
Such a setting is inspired by~\cite{springer2021s}, where the authors investigate the ability of neural networks to learn the rules of Conway's Game of Life from the given state transition images.
The authors have observed that such a task is often not achievable by conventional machine learning approaches and is only attainable under a sufficient network overparameterization (about $5-10$ times for a successful $1$- or $2$-step prediction).
We restrict the setting by only working with minimally sufficient architectures and do not allow network overparameterization, which effectively puts us in a domain of \textit{constrained machine learning}.
To offset the lack of usual machine learning leniency, we allow for meticulous control of the training data.
Specifically, we design a training board and compare the efficiency of training neural networks on this constructed board, rather than learning on randomly generated boards.
We observe that even a single properly-crafted training board offers a significant increase in convergence rate and speed, especially on more challenging multi-step prediction tasks, regardless of the choice of a network initialization and an optimization algorithm.
We use this observation to highlight the vital importance of the training data in constrained settings, which is well-aligned with the increasingly prominent concept of \textit{data-centric machine learning}.
We believe that this example can be easily translated to many real-world applications where machine learning approaches are restricted by the practical constraints like ethical, legal, security, or hardware limitations.

\begin{figure}[t]
    \centering
    \begin{subfigure}{.3\linewidth}
        \includegraphics[width=\linewidth]{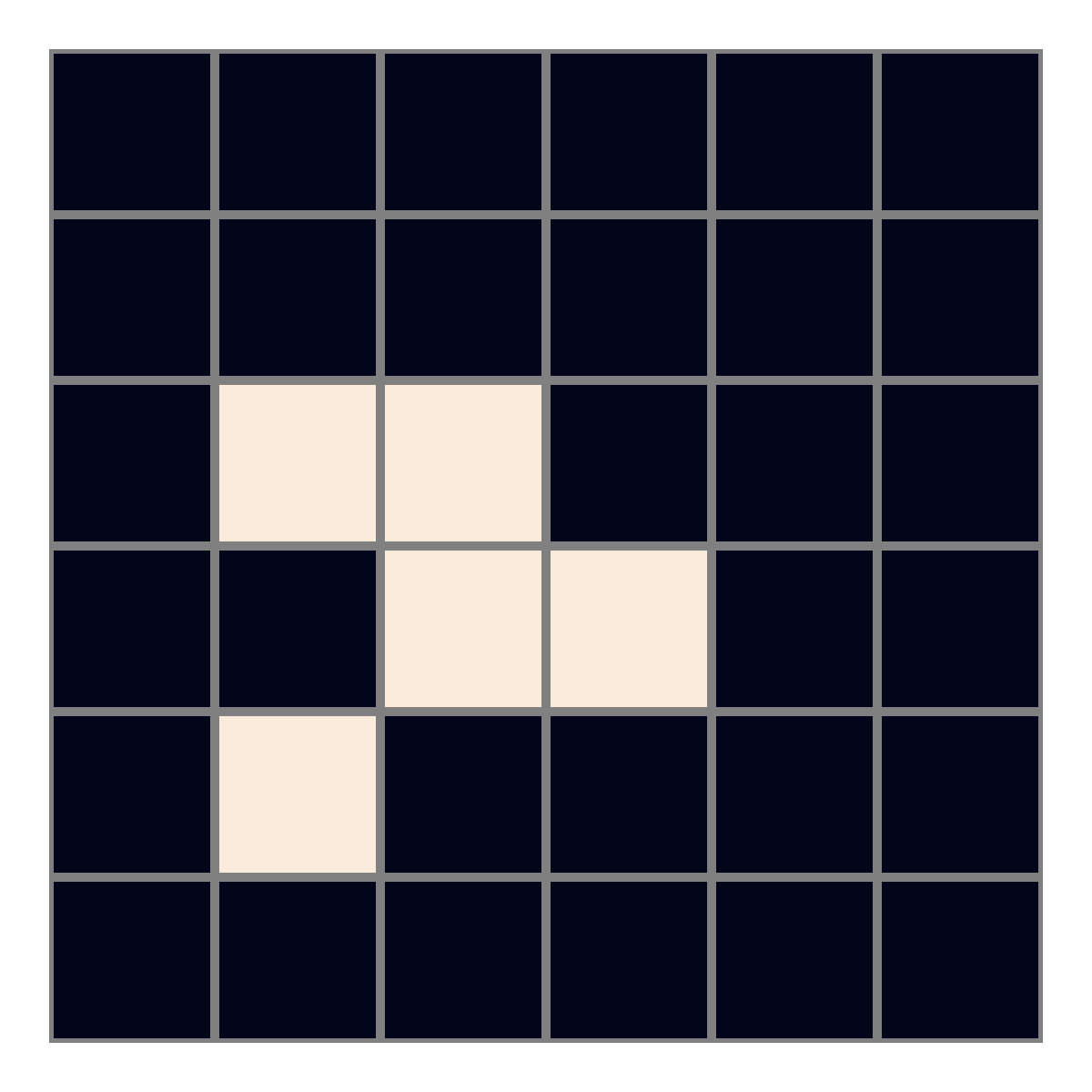}
        \caption{State at time 0}
    \end{subfigure}
    \begin{subfigure}{.3\linewidth}
        \includegraphics[width=\linewidth]{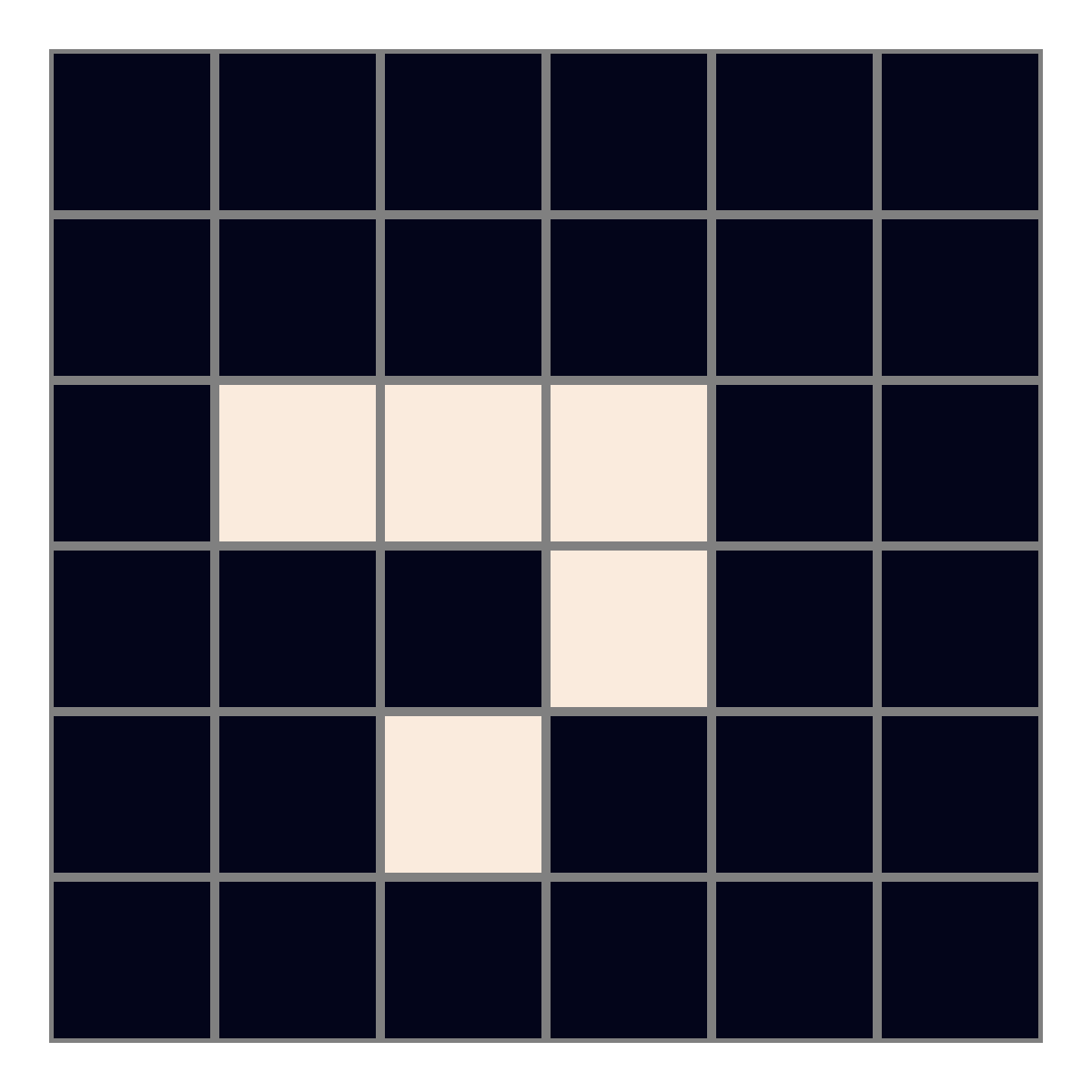}
        \caption{State at time 1}
    \end{subfigure}
    \begin{subfigure}{.3\linewidth}
        \includegraphics[width=\linewidth]{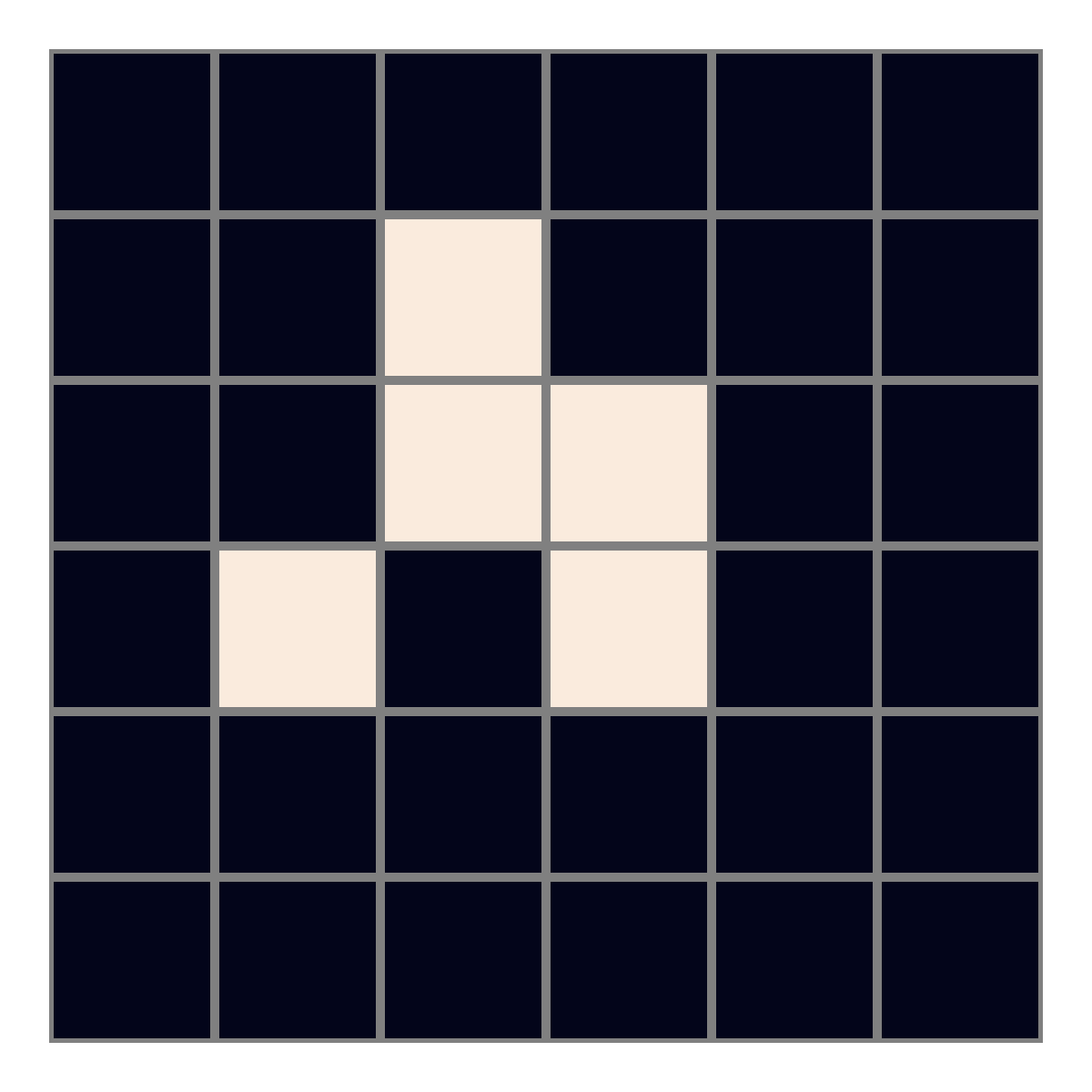}
        \caption{State at time 2}
    \end{subfigure}
    \\
    \begin{subfigure}{.3\linewidth}
        \includegraphics[width=\linewidth]{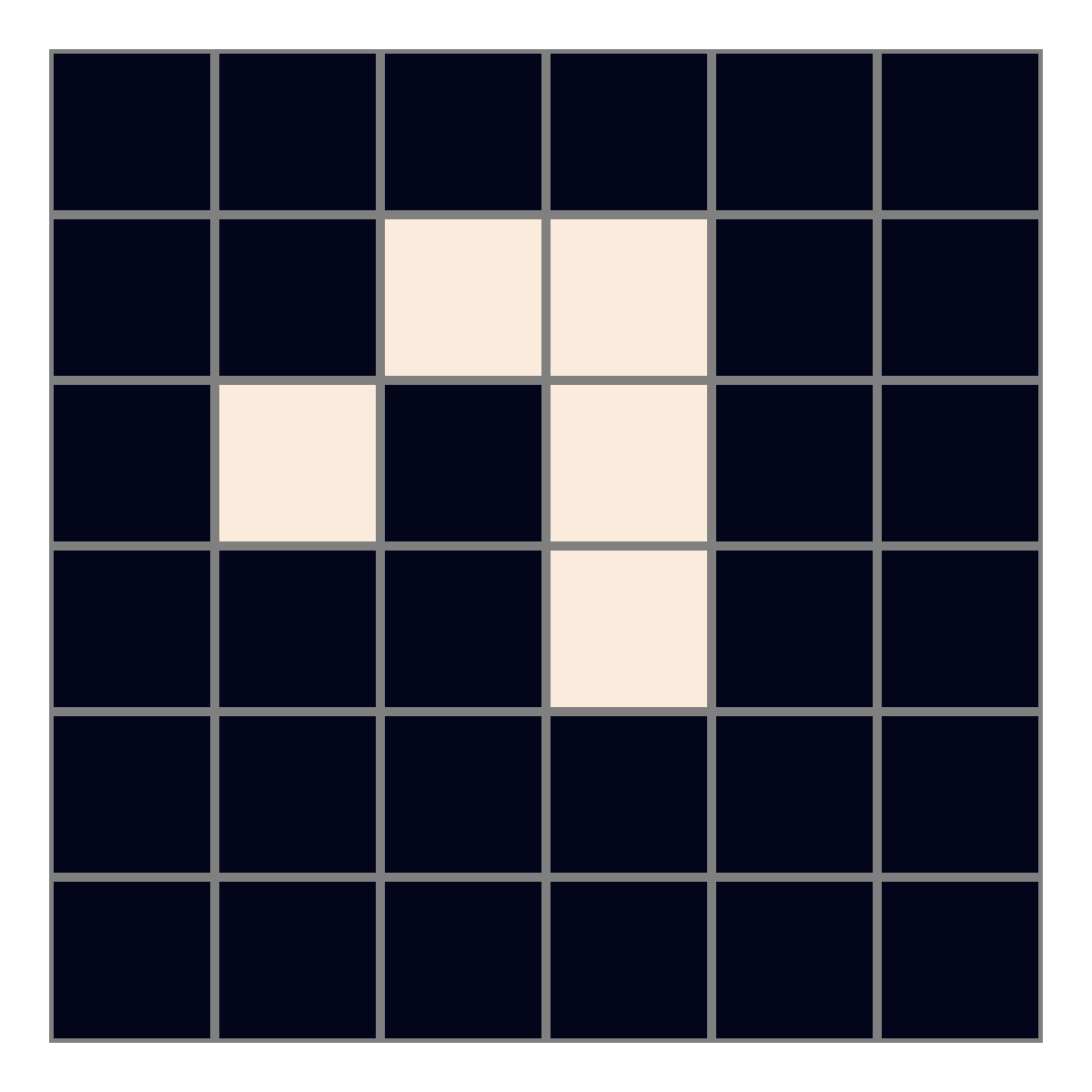}
        \caption{State at time 3}
    \end{subfigure}
    \begin{subfigure}{.3\linewidth}
        \includegraphics[width=\linewidth]{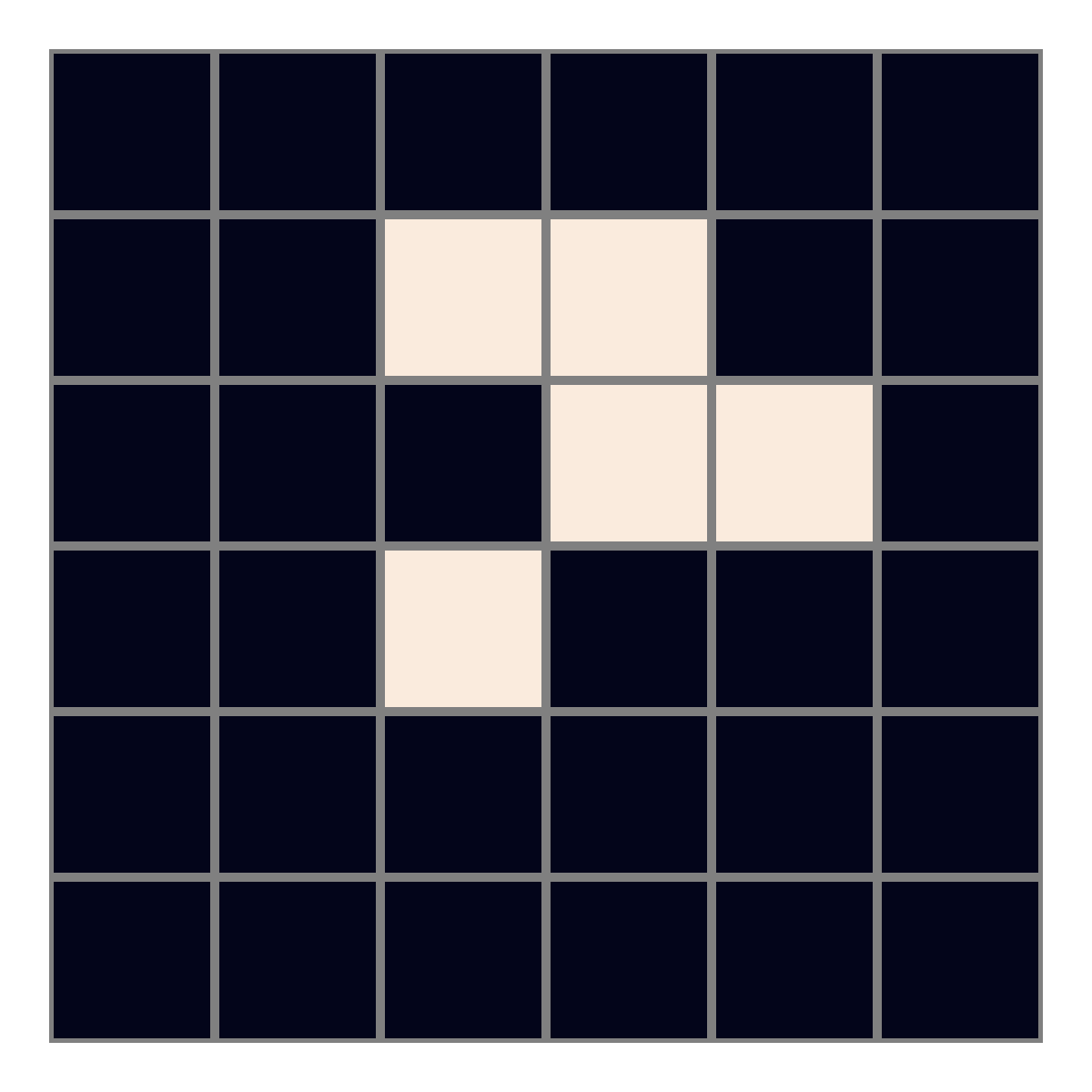}
        \caption{State at time 4}
    \end{subfigure}
    \begin{subfigure}{.3\linewidth}
        \includegraphics[width=\linewidth]{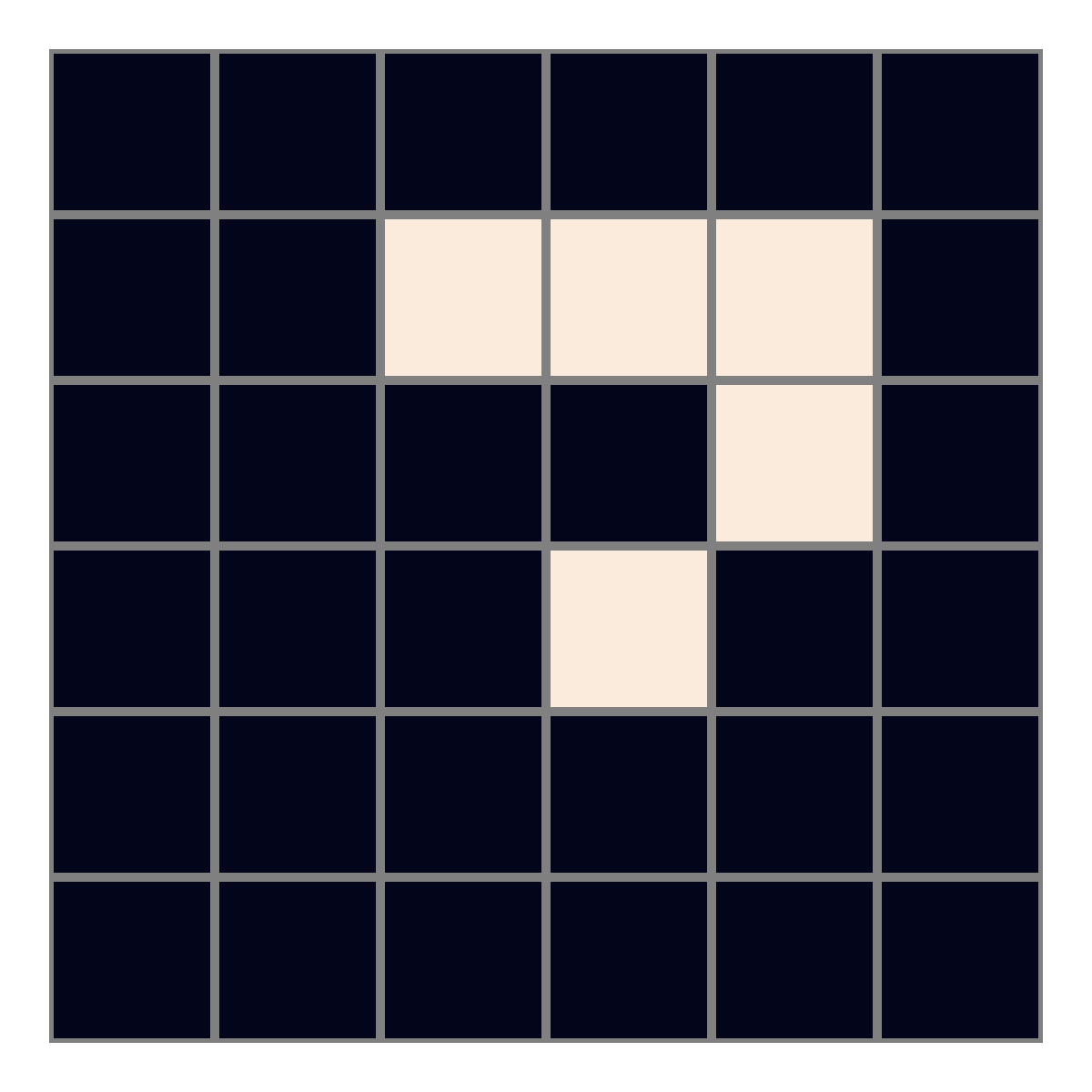}
        \caption{State at time 5}
    \end{subfigure}
    \caption{An example of a state trajectory in the Game of Life.
             Alive cells are white and dead cells are black.}
    \label{fig:gol_states}
\end{figure}

Our contributions are the following:
\begin{itemize}
    \item Develop multiple ways of representing the Game of Life as a neural network with minimal architecture;
    \item Explain the process of constructing the training data by analyzing the environment;
    \item Demonstrate the significance of the training data choice via an extensive qualitative analysis;
    \item Establish a simple yet challenging task that can be used by practitioners as a benchmark.
\end{itemize}

\subsection{Conway's Game of Life}

Conway's Game of Life, created by mathematician John Conway in 1970, is a two-dimensional cellular automaton where cells are either alive or dead.
The evolution of the system is determined by rules based on the states of neighboring cells, as shown in Figure~\ref{fig:gol_states}.
Despite its simplicity, it remains significant in mathematics and machine learning due to its inherent complexity and applications~\cite{rennard2002implementation, rendell2011universal}.

The Game of Life models how simple rules can produce complex patterns.
This has led to studies on its mathematical properties and its use in developing neural networks to simulate and predict system evolution~\cite{adamatzky2010game, hirte2022john, krechetov2021game, grattarola2021learning}.

In this work we explore the role of the training data on a task of learning the multi-step Game of Life with the minimal conventional architecture.
Specifically, we use a convolutional neural network (CNN) to learn the rules of the game and predict the next state of the board.
Training on our custom-made board outperforms the traditional paradigm in terms of accuracy and speed, and showcases the advantage of a data-centric approach in constrained machine learning applications.

\subsection{Constrained Machine Learning}

Machine learning has become essential in many fields, but traditional methods often ignore constraints necessary for ethical, reliable, and safe decision-making.
Constrained machine learning integrates explicit constraints (either mathematical, logical, or rule-based) derived from domain knowledge, into the learning process~\cite{perez2021constrained, yao2021power, gori2023machine}.

This approach is crucial in areas like healthcare~\cite{ahmad2018interpretable, chen2021ethical, nguyen2021budget}, autonomous vehicles~\cite{gonzalez2015review, dalal2018safe}, finance~\cite{bae2022constrained, al2022optimization}, and natural language processing~\cite{ammanabrolu2020graph, yang2021safe, zhang2022survey}.

It is important to note that incorporating additional constraints can increase the implementation, training, and deployment complexity of machine learning models.
But despite the potential challenges, constrained machine learning offers a promising avenue for improving the behavior and performance of machine learning models in critical real-life applications.

\subsection{Data-Centric Machine Learning}

Data-centric machine learning focuses on the quality, diversity, and relevance of data rather than just model complexity and hyperparameter tuning.
High-quality and abundant data significantly affect model performance and generalization~\cite{pan2022data, majeed2023data}.
Data-centric machine learning finds application in numerous real-world scenarios across various industries, including healthcare~\cite{zahid2021systematic, emmert2022digital, dritsas2022data}, finance~\cite{liu2022finrl, horvatha2023harnessing}, and environmental sciences~\cite{devarajan2021dlio, li2022big}.

Challenges include ensuring data quality and accessibility, handling imbalanced or noisy data, and addressing privacy and ethical concerns~\cite{roh2019survey, jo2020lessons}.
Domain expert insights are crucial for evaluating data relevance and guiding preprocessing steps to ensure effective training and validation of machine learning models~\cite{gennatas2020expert}.
Addressing these challenges is essential for realizing the full potential of data-centric machine learning~\cite{anik2021data, miranda2021towards, seedat2022dc, zha2023data}.

\begin{figure*}[t]
    \centering
    \includegraphics[width=.8\linewidth]{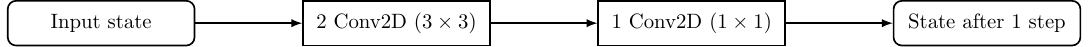}
    \caption{Minimal CNN architecture for the Game of Life network.}
    \label{fig:gol_model_1_step}
\end{figure*}

\subsection{Related Work}
We chose Game of Life as a challenging environment that is complicated for conventional machine learning algorithms to learn.
Such a setting is inspired by~\cite{springer2021s}, where the authors investigated how network overparameterization affects performance.
In contrast, we only consider the minimal network architectures.

In this work we only consider the classical discrete formulation of Game of Life, however numerous other generalizations have been proposed and studied, see e.g.~\cite{adachi2004game, rafler2011generalization, aach2021generalization}.

Training networks of the smallest feasible architecture is considered in~\cite{nye2018efficient}, where the authors observe the inability of learning the parity function and fast Fourier transform with the minimal networks.
Research on the training of small networks is relevant, see e.g.~\cite{winoto2020small} where the authors consider slim networks for deployment on mobile devices.
Additionally, small, or even minimal, networks can be obtained by pruning, which is another prominent area of research, see e.g.~\cite{blalock2020state}.

We consider the setting of constrained machine learning, where the main constraint is the size of the network.
Depending on the application, researchers are interested in other practical constraints, such as the scarce amount of data~\cite{dereventsov2021offline, fu2021benchmarks}, skewed distribution of historical data~\cite{aslanides2017universal, bojun2020steady}, and other practical limitations~\cite{tennenholtz2020off, dulac2021challenges}.
Due to the restricted nature of our setting, we employ full-CNN networks to perform image-to-image translation.
However, more sophisticated techniques are available if the condition of minimal architecture is relaxed, see e.g.~\cite{isola2017image, murez2018image, pang2021image}.

One of our main results demonstrates that even a single state observation could be sufficient to learn the Game of Life.
Similar findings are presented in~\cite{motamedi2021data}, where the authors achieve high performance in an image classification task by removing a significant part of the training data and only keeping the highest quality images.
The process of designing or creating training data itself is an actively developing area that has seen significant progress in different areas of machine learning, see e.g.~\cite{ratner2016data, abufadda2021survey}.

%%%%%%%%%%%%%%%%%%%%%%%%%%%%%%%%%%%%%%%%%%%%%%%%%%%%%%%%%%%%%%%%%%%%%%%%%%%%%%%%%%%%%%%%%%%%%%%%%%%
\section{Game of Life as a Neural Network}
In the Game of Life, the state of the board changes each turn, depending on the current board configuration, according to the following rules:
\begin{enumerate}
    \item Any live cell with fewer than two live neighbors dies, as if by underpopulation.
    \item Any live cell with two or three live neighbors lives on to the next generation.
    \item Any live cell with more than three live neighbors dies, as if by overpopulation.
    \item Any dead cell with exactly three live neighbors becomes a live cell, as if by reproduction.
\end{enumerate}

Following the established convention, we represent a state of the board in Game of Life by a grayscale image, see Figure~\ref{fig:gol_states}, where a pixel value of $1$ indicates that the cell is alive and a value of $0$ indicates that the cell is dead.
Thus at a time $t \ge 0$ the state is given as a binary matrix $s_t \in \mathbb{R}^{M \times N}$, where each value $c \in s_t$ represents the condition of a cell at time $t$ with $c = 1$ indicating that the cell is alive and $c = 0$ indicating that the cell is dead.
Therefore the Game of Life transition can be viewed as an operator $\mathcal{G}: \mathbb{R}^{M \times N} \to \mathbb{R}^{M \times N}$ acting on the current state $s_t$ and outputting the next state $s_{t+1}$, i.e.
\[
    \mathcal{G}(s_t) = s_{t+1}.
\]
To establish a neural network representation of the operator $\mathcal{G}$, note that a single step $s_t \to s_{t+1}$ of Game of Life consists of transforming each cell $c \to c^\prime$ as
\begin{equation}\label{eq:gol_rules}
    c^\prime
    = \left\{\begin{array}{rl}
        1 & \text{if the $3 \times 3$ patch centered at $c$}
            \\
            & \text{has at least $3$ living cells and}
            \\
            & \text{$c$ has at most $3$ living neighbors},
        \\
        0 & \text{otherwise}.
    \end{array}\right.
\end{equation}
Therefore in order to make a transition $s_t \to s_{t+1}$, for each cell $c$ one has to know two pieces of information: the condition of the cell $c$ itself and the conditions of the cells in the $3 \times 3$ patch centered at cell $c$.
A conventional architecture that can efficiently extract this information with just a few trainable parameters is the \textit{Convolutional Neural Network}, see e.g.~\cite{goodfellow2016convolutional}.

Indeed, the Game of Life operator $\mathcal{G}: \mathbb{R}^{M \times N} \to \mathbb{R}^{M \times N}$ can be represented as a 2-layer convolutional neural network with $23$ trainable parameters, see Figure~\ref{fig:gol_model_1_step}.
Conceptually, the first layer contains two $3 \times 3$ filters, extracting the information on the number of alive neighbors and the condition of the cell itself.
This layer can be followed by any reasonable activation function, so we consider ReLU and Tanh~--- the most popular choices among practitioners.
The second layer combines these two features to make a prediction on the cell condition on the next step.
Unlike the first layer, the choice of activation function here is more restrictive since the output of $\mathcal{G}$ should only consist of $0$ and $1$, regardless of the input.
As such, we only consider the ReLU activation after the second layer.

Below we provide two minimal configurations with conventional architectures that are capable of capturing the transition rules~\eqref{eq:gol_rules}.
We note that while it is technically possible to construct a smaller network recreating the rules for Game of Life, it would require either a use of residual connections, unconventional layer structures, or activation functions, which is beyond the scope of the current work, and hence we only consider the fully-CNN architectures.

\begin{figure*}[t]
    \centering
    \includegraphics[width=.8\linewidth]{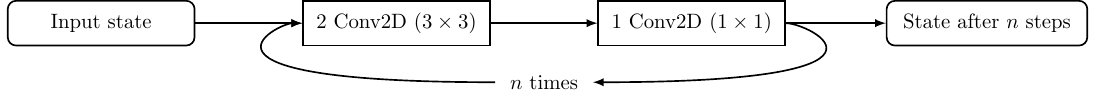}
    \caption{Minimal recursive CNN architecture for the multi-step Game of Life network.}
    \label{fig:gol_model_recursive}
\end{figure*}

\begin{figure*}[t]
    \centering
    \includegraphics[width=.8\linewidth]{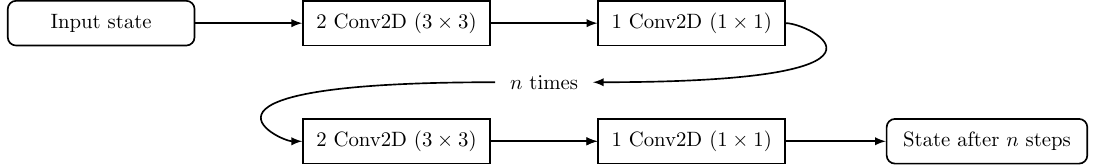}
    \caption{Minimal sequential CNN architecture for the multi-step Game of Life network.}
    \label{fig:gol_model_sequential}
\end{figure*}

\subsection{ReLU Activation}
Consider a $2$-layer convolutional neural network with the ReLU activation after each layer:
\[
    s_t \ \to\ 
    2\ \text{Conv2D}(3 \times 3) \ \xrightarrow{\text{ReLU}}\ 
    1\ \text{Conv2D}(1 \times 1) \ \xrightarrow{\text{ReLU}}\ 
    s_{t+1}
\]
The network consists of convolutional filters with the following weights and biases:
\begin{align*}
    W_{1,1} &= \begin{pmatrix}
        1 & 1 & 1
        \\
        1 & 0 & 1
        \\
        1 & 1 & 1
    \end{pmatrix}
    &
    b_{1,1} &= -3
    \\
    W_{1,2} &= \begin{pmatrix}
        -1 & -1 & -1
        \\
        -1 & -1 & -1
        \\
        -1 & -1 & -1
    \end{pmatrix}
    &
    b_{1,2} &= 3
    \\
    W_{2,1} &= \begin{pmatrix} -1 \\ -1 \end{pmatrix}
    &
    b_{2,1} &= 1
\end{align*}
where $W_{i,j}$ and $b_{i,j}$ are the $j$-th weights and bias of the $i$-th layer of the network.
Note that for a state $s_t \in \mathbb{R}^{M \times N}$ we get the following output of the first layer:
\begin{multline*}
    \text{ReLU}(W_{1,1} s_t + b_{1,1})
    = [c_{ij}^\prime]_{i=1,j=1}^{M,N}
    \\
    = \left\{\begin{array}{rl}
        0 & \text{if $c_{ij}$ has $\le 3$ living neighbors,}
        \\
        > 0 & \text{otherwise.}
    \end{array}\right.
\end{multline*}
\begin{multline*}
    \text{ReLU}(W_{1,2} s + b_{1,2})
    = [c_{ij}^\prime]_{i=1,j=1}^{M,N}
    \\
    = \left\{\begin{array}{rl}
        0 & \text{if a $3 \times 3$ patch at $c_{ij}$ has $\ge 2$ living cells,}
        \\
        > 0 & \text{otherwise.}
    \end{array}\right.
\end{multline*}
Therefore, after the second layer, the value of a cell $c^\prime$ of the output state $s_{t+1}$ is equal $1$ if a $3 \times 3$ patch centered at a cell $c$ has at least $2$ living cells and $c$ has at most $3$ living neighbors, or $0$ otherwise, which correspond to the Game of Life rules~\eqref{eq:gol_rules}.

\subsection{Tanh Activation}
Consider a $2$-layer convolutional neural network with the Tanh activation after the first layer:
\[
    s_t \ \to\ 
    2\ \text{Conv2D}(3 \times 3) \ \xrightarrow{\text{Tanh}}\ 
    1\ \text{Conv2D}(1 \times 1) \ \xrightarrow{\text{ReLU}}\ 
    s_{t+1}
\]
The network consists of convolutional filters with the following weights and biases:
\begin{align*}
    W_{1,1} &= \begin{pmatrix}
        1 & 1 & 1
        \\
        1 & \nicefrac{3}{5} & 1
        \\
        1 & 1 & 1
    \end{pmatrix}
    &
    b_{1,1} &= -2.4
    \\
    W_{1,2} &= \begin{pmatrix}
        1 & 1 & 1
        \\
        1 & \nicefrac{2}{5} & 1
        \\
        1 & 1 & 1
    \end{pmatrix}
    &
    b_{1,2} &= -3.6
    \\
    W_{2,1} &= \begin{pmatrix} 2 \\ -2 \end{pmatrix}
    &
    b_{2,1} &= -1
\end{align*}
where $W_{i,j}$ and $b_{i,j}$ are the $j$-th weights and bias of the $i$-th layer of the network.
Then for any cell $c \in \{0;1\}$ with $n \in \{0,1,2,3,4,5,6,7,8\}$ alive neighbors the output of the second layer before the ReLU activation is the following:
\begin{gather*}
    \begin{array}{c|cc}
        & c = 0 & c = 1
        \\\hline
        n = 0 & -0.9703 & -0.9002
        \\
        n = 1 & -0.7926 & -0.3766
        \\
        n = 2 & \phantom{-}0.0834 & \phantom{-}1.0621
        \\
        n = 3 & \phantom{-}1.1482 & \phantom{-}1.0621
        \\
        n = 4 & \phantom{-}0.0834 & -0.3766
        \\
        n = 5 & -0.7926 & -0.9002
        \\
        n = 6 & -0.9703 & -0.9862
        \\
        n = 7 & -0.9960 & -0.9981
        \\
        n = 8 & -0.9995 & -0.9997
    \end{array}
\end{gather*}
Thus, after applying the ReLU activation and predicting the condition of a cell $c^\prime$, we obtain a transition corresponding to the Game of Life rules~\eqref{eq:gol_rules}.
We note that while technically this network does not achieve zero loss, it is capable of reaching $100\%$ accuracy and is easier to train in practice, likely due to a more gradual feature transfer of the Tanh function.

\subsection{Multi-step Game of Life}
In order to evaluate the benefits of a data-centric approach in more challenging scenarios, we consider the task of learning $n$-steps Game of Life, which becomes increasingly more complex with the increase of the number of steps $n$.
In this formulation, the training data is given in the form of pairs $(x,y)$, where $y$ is the state $x$ after $n$ steps of the Game of Life. %, see e.g. Figure~\ref{fig:gol_board_random_0}, \ref{fig:gol_board_random_2} and Figure~\ref{fig:gol_board_fixed_0}, \ref{fig:gol_board_fixed_2} for $n = 2$.
In this case the multi-step operator $\mathcal{G}_n$ can be represented in either of the following ways:
\begin{itemize}
    \item $1$-step network, recursively fed into itself $n$ times, which results in a total of $23$ trainable parameters, same as the single-step network, see Figure~\ref{fig:gol_model_recursive};
    \item $n$ instances of the $1$-step network connected sequentially, which results in a total of $23n$ trainable parameters, see Figure~\ref{fig:gol_model_sequential}.
\end{itemize}
In our numerical experiments we consider both options.
We also note that in this paper we only consider $1$- and $2$-step formulations of Game of Life learning, since for $n \ge 3$ none of the learning configurations managed to learn the environment regardless of the choices of initialization weight, optimization algorithm, and hyperparameters.
This is due to the constrained nature of our setting as we are only considering networks of minimal architecture.

%%%%%%%%%%%%%%%%%%%%%%%%%%%%%%%%%%%%%%%%%%%%%%%%%%%%%%%%%%%%%%%%%%%%%%%%%%%%%%%%%%%%%%%%%%%%%%%%%%%
\section{Training Data Design}
A state in Game of Life is represented by a \textit{board}~--- a $2$-dimensional binary matrix where a value of $0$/$1$ represents a dead/alive cell respectively~--- and a training set is given as a collection of boards.
When constructing a training set, one has to decide on the number of training boards, size of each board, and an average board \textit{density}~--- the percentage of alive cells on the board.
Below we explain the choice of these parameters for each of two datasets we use in this paper: the "random" dataset and the "fixed" dataset.

A conventional way of obtaining training data for a classification task typically involves generating a sufficiently large number of pairs $(x,y)$ via the methods appropriate for the current scenario~\cite{roh2019survey}.
In the case of $n$-step Game of Life, $x$ represents the input board and $y$ represents the corresponding output board same state after $n$ steps.
In practice, such a dataset can be obtained by fixing some initial states and computing the corresponding outputs.

\subsection{Random Dataset}
\begin{figure}[h!]
    \centering
    \includegraphics[width=.95\linewidth]{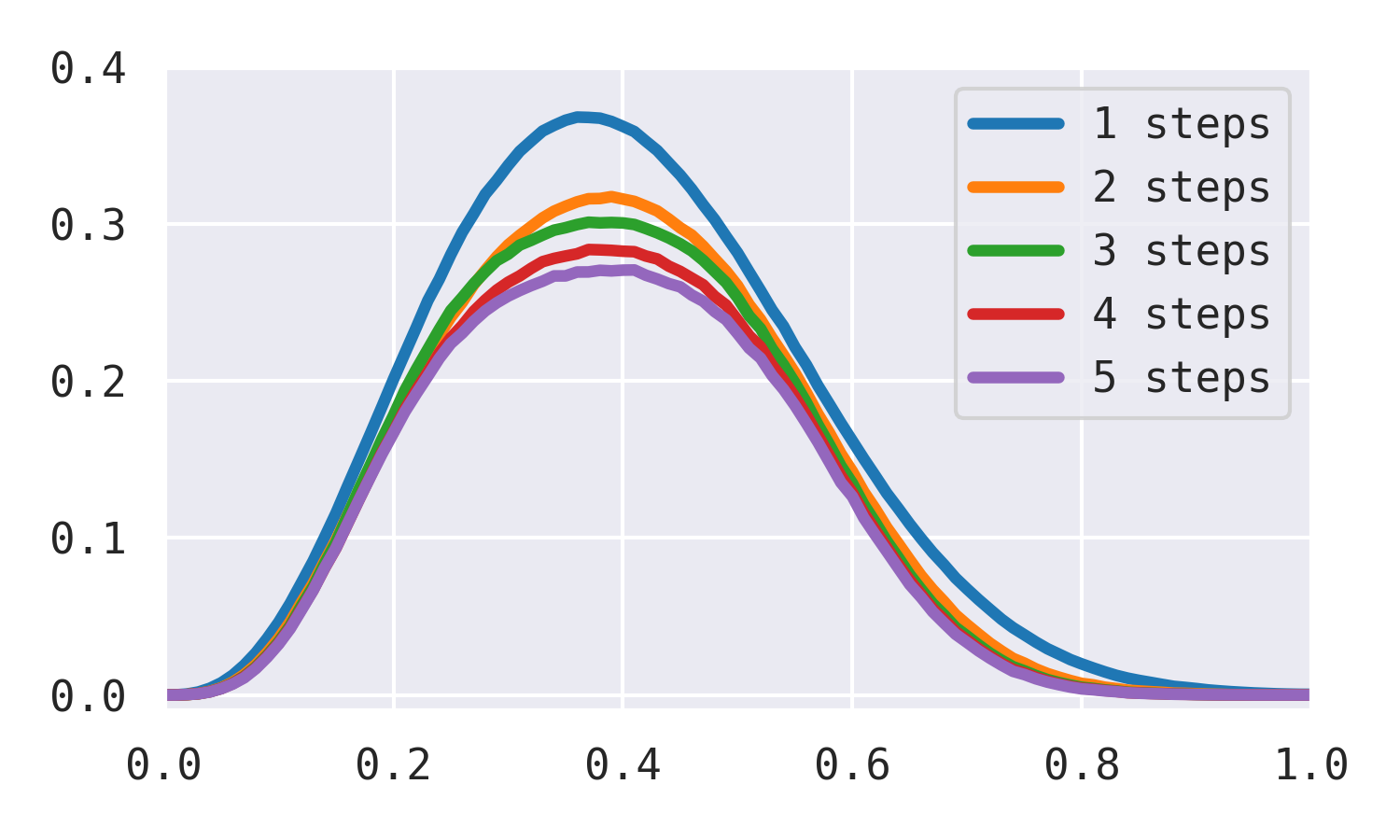}
    \caption{Average board density after multiple steps of the Game of Life.}
    \label{fig:gol_board_density}
\end{figure}

While there are multiple viable choices for each of these parameters, in the interest of being consistent with the existing literature, we replicate the data collection choices employed in~\cite{springer2021s}.
Specifically, we implement a data generator to randomly sample boards with an average density of $38\%$, as such a density minimizes the distribution shift between the inputs and outputs, see Figure~\ref{fig:gol_board_density}, and also maximizes the learning success rate, see~\cite[Section 3.4]{springer2021s}.

\begin{figure}[t]
    \centering
    \includegraphics[width=.78\linewidth]{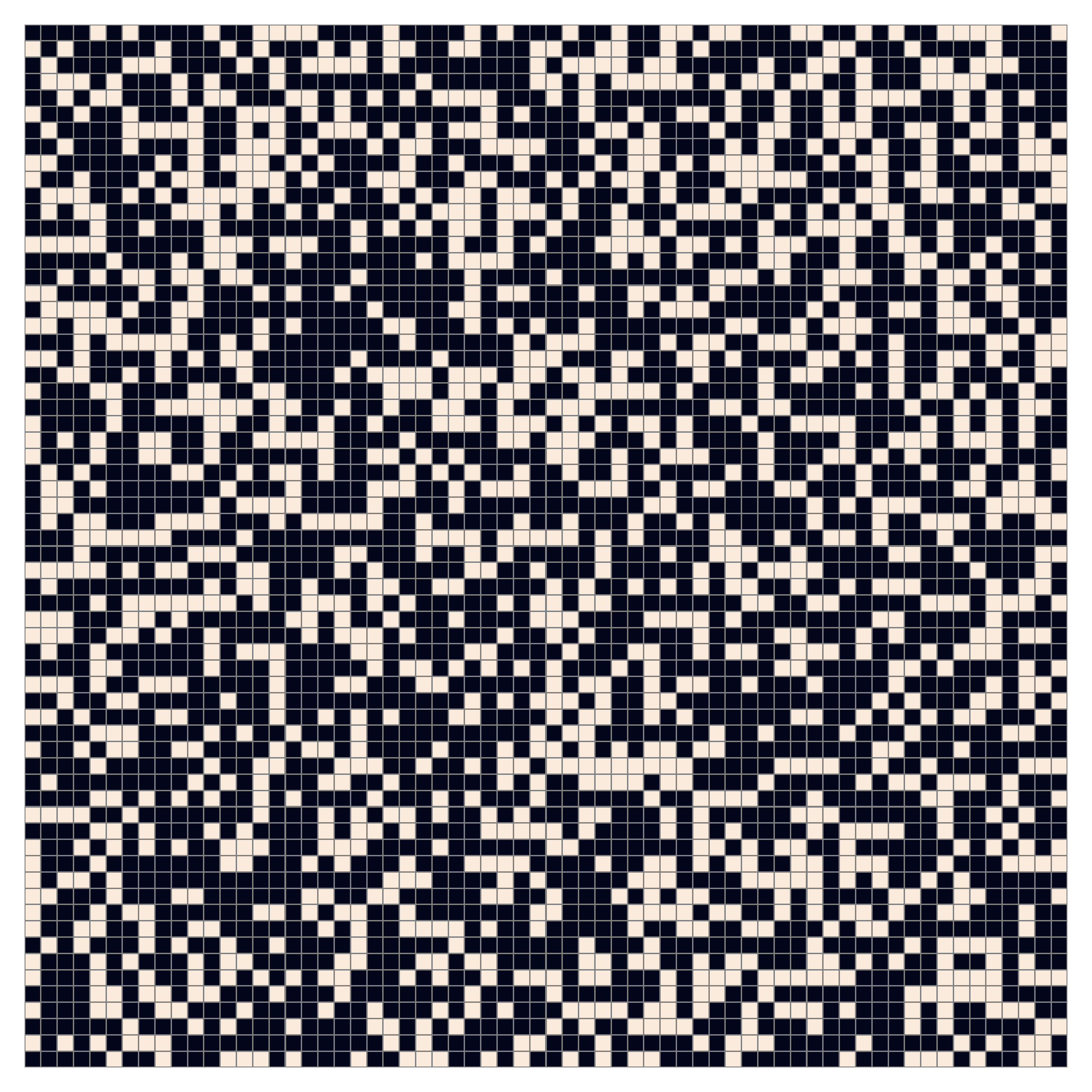}
    \caption{Example of a generated training board.}
    \label{fig:gol_board_random}
\end{figure}

The size of each training board is set to $64 \times 64$ to match the size of our manually designed training board.
In order to mitigate the risk of learning on a poorly generated board, we sample a new training board on each iteration of the algorithm when training on the "random" dataset.
An example of the generated training board is presented in Figure~\ref{fig:gol_board_random}.

\subsection{Fixed Dataset}
A comprehensive answer on the most appropriate training dataset design is highly application-specific and is yet to be found in general.
In the context of Game of Life, the main challenges are the redundancy of the patterns in random board that dilute the knowledge and the undecidability of the state dynamics.
To address these challenges, we utilize two main concepts in the process of board construction: \textit{symmetry} and meaningful diversity of \textit{patterns}.

As it follows from the rules~\eqref{eq:gol_rules}, in order to make a prediction on the condition of the current state, one does not have to distinguish the neighboring cells; thus, one can deduce that the filters of the first layer should be symmetric about the center.
However, it is non-trivial to achieve symmetry of the filters by utilizing a randomly sampled dataset.
To endorse the symmetric structure of the convolutional filters, we make our board symmetric by performing horizontal and vertical reflections.
Such an arrangement essentially means that each $3 \times 3$ patch is seen by the network $4$ times with different orientations, thus promoting symmetry of the filters.

\begin{figure}[h]
    \centering
    \includegraphics[width=.24\linewidth]{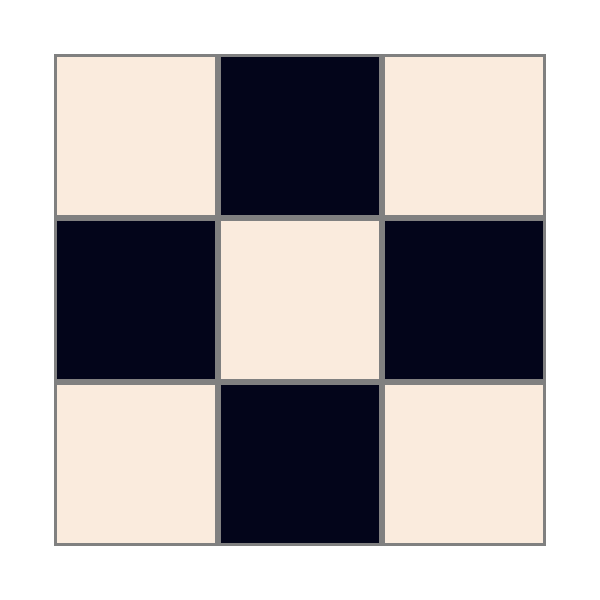}
    \includegraphics[width=.24\linewidth]{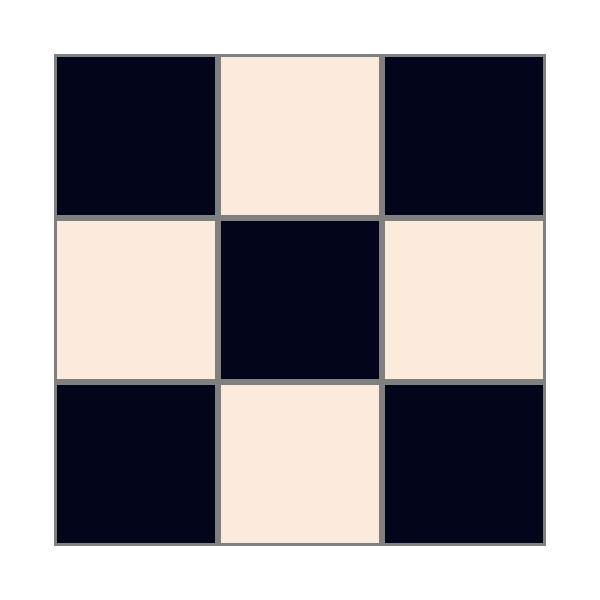}
    \includegraphics[width=.24\linewidth]{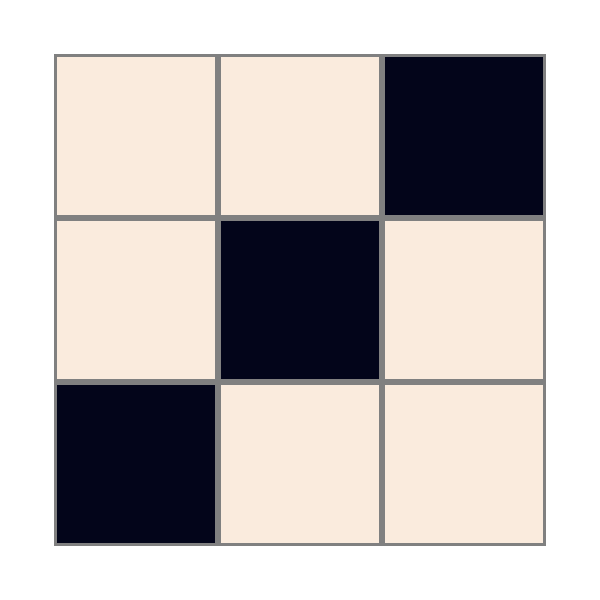}
    \includegraphics[width=.24\linewidth]{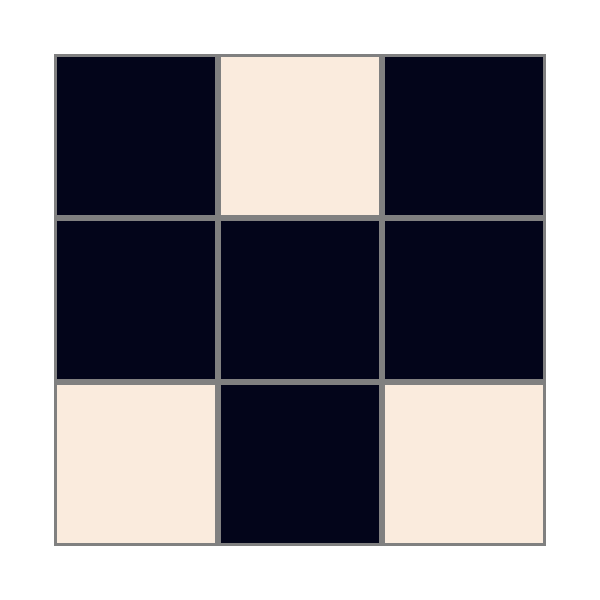}
    \\
    \includegraphics[width=.24\linewidth]{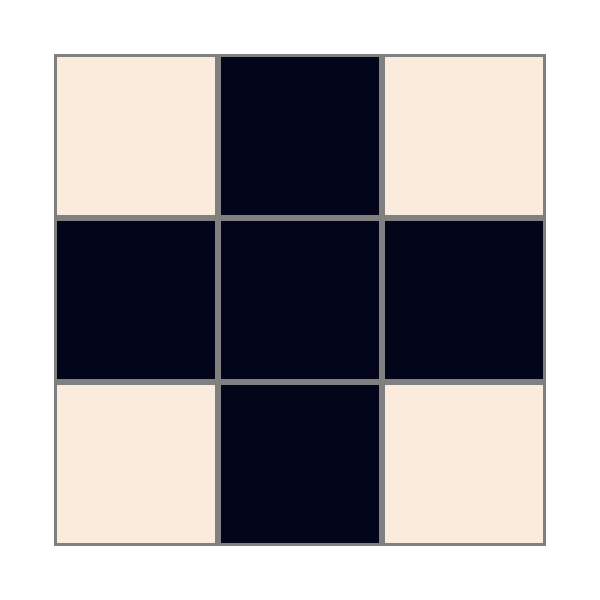}
    \includegraphics[width=.24\linewidth]{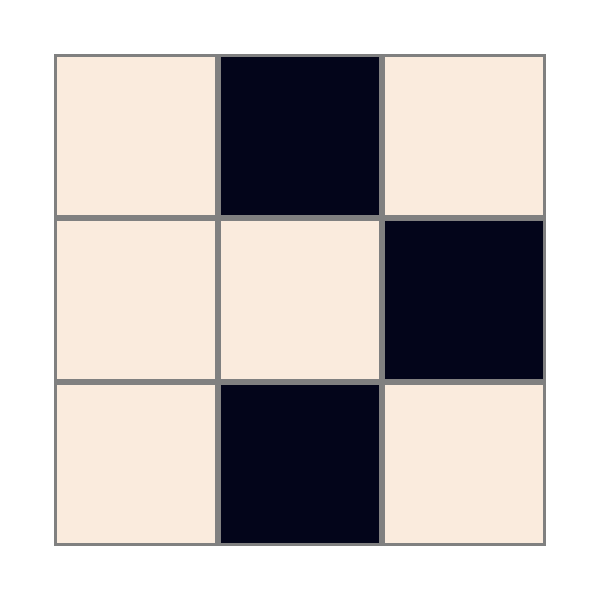}
    \includegraphics[width=.24\linewidth]{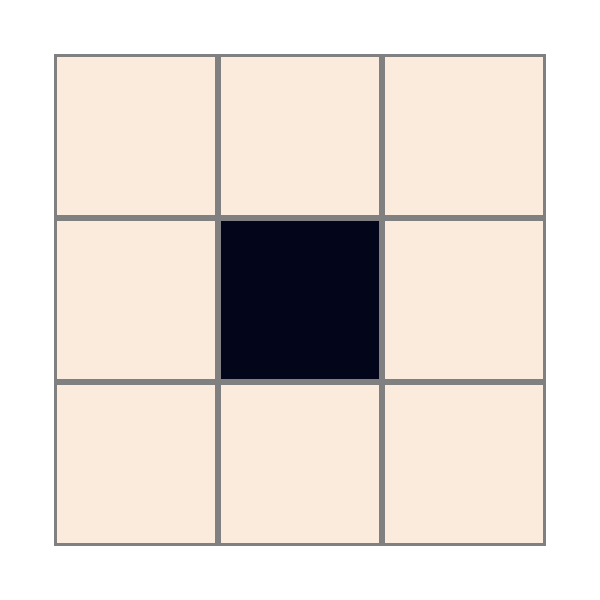}
    \includegraphics[width=.24\linewidth]{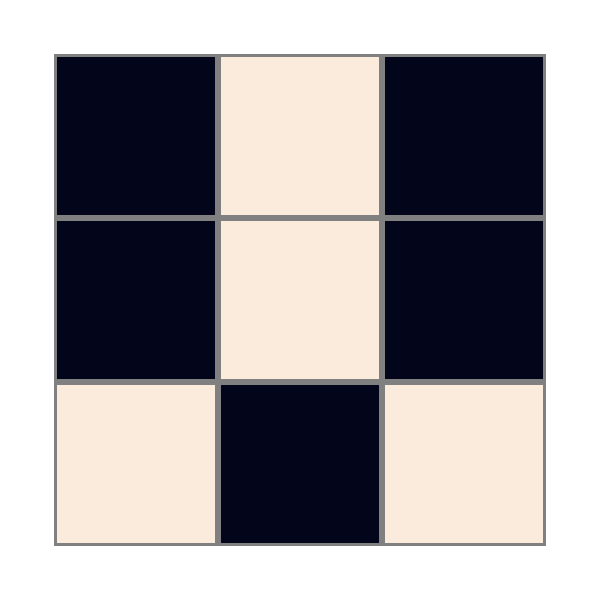}
    \caption{Examples of patterns utilized in construction of the "fixed" training board for the Game of Life.}
    \label{fig:gol_patterns}
\end{figure}

It is evident that the training data must contain a variety of $3 \times 3$ patches from which the rules~\eqref{eq:gol_rules} can be learned.
While considering every possible $3 \times 3$ patch is hardly feasible, as it results in $2^9 = 512$ different configurations, it is also not necessary, as most of them are redundant due to the indistinguishability of the neighbors and the symmetry of the filters.
As such, one can prioritize putting the more important patterns in the training board and thereby avoid excessive redundancy, which is the approach we pursue.
Some of the patterns we use are presented in Figure~\ref{fig:gol_patterns}.

\begin{figure}[t]
    \centering
    \includegraphics[width=.78\linewidth]{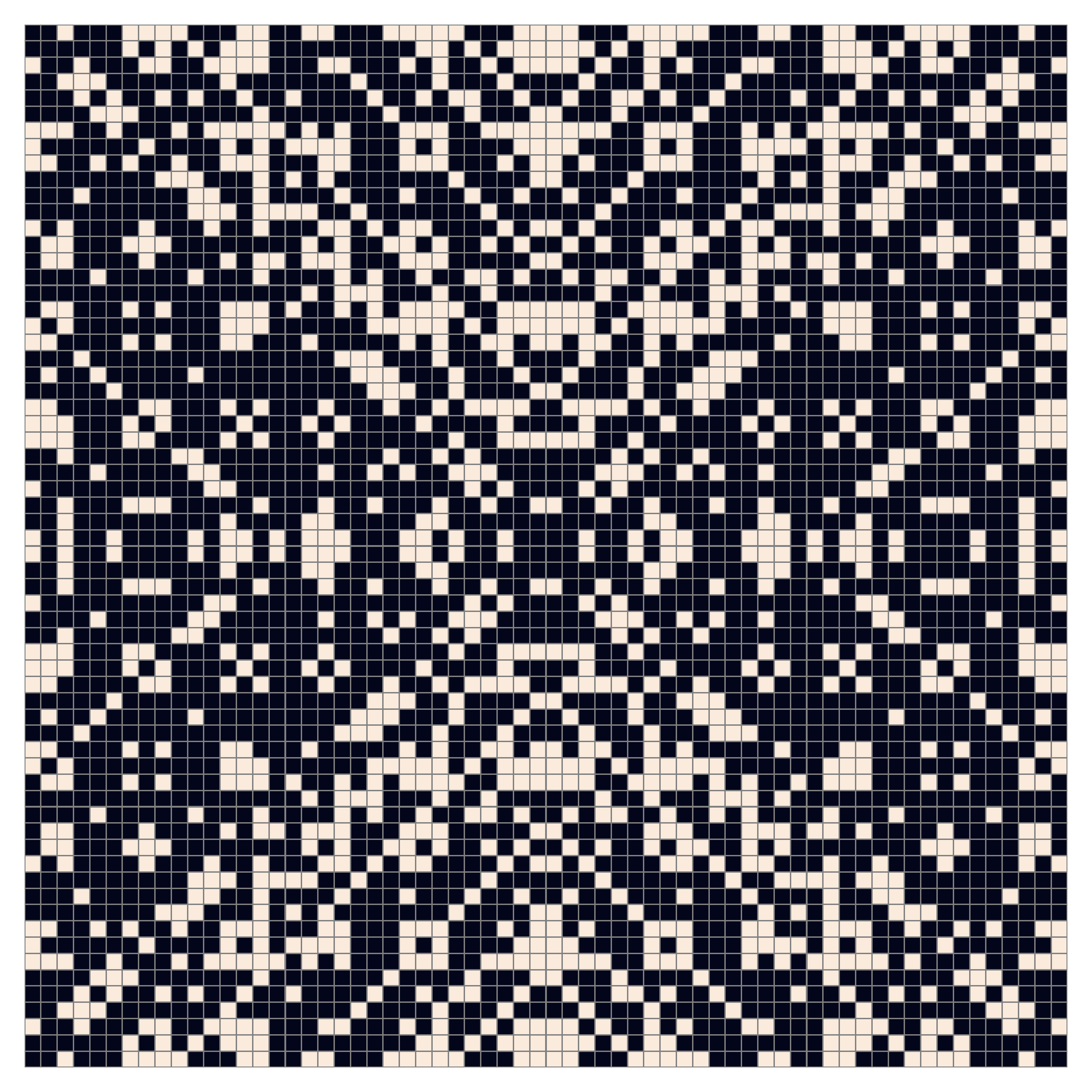}
    \caption{Manually constructed training board.}
    \label{fig:gol_board_fixed}
\end{figure}

To accommodate the above concepts, we first construct a $32 \times 32$ board that contains the critical patterns, and then reflect it horizontally and vertically to harness the power of symmetry, which results in a $64 \times 64$ board.
Note that the reflection of the board automatically creates additional variants of the patterns that are not explicitly included in the original $32 \times 32$ board, which allows us to train on a single board.
The constructed board that we use in our numerical examples is presented in Figure~\ref{fig:gol_board_fixed}.

% \begin{figure*}[t]
%     \centering
%     \begin{subfigure}{.3\linewidth}
%         \includegraphics[width=\linewidth]{./images/gol_board_random_0.png}
%         \caption{Input state}
%         \label{fig:gol_board_random_0}
%     \end{subfigure}
%     \begin{subfigure}{.3\linewidth}
%         \includegraphics[width=\linewidth]{./images/gol_board_random_1.png}
%         \caption{State after $1$ step}
%         \label{fig:gol_board_random_1}
%     \end{subfigure}
%     \begin{subfigure}{.3\linewidth}
%         \includegraphics[width=\linewidth]{./images/gol_board_random_2.png}
%         \caption{State after $2$ steps}
%         \label{fig:gol_board_random_2}
%     \end{subfigure}
%     \caption{Example of a generated training board for the Game of Life.}
%     \label{fig:gol_board_random}
% \end{figure*}

% \begin{figure*}[t]
%     \centering
%     \begin{subfigure}{.3\linewidth}
%         \includegraphics[width=\linewidth]{./images/gol_board_fixed_0.png}
%         \caption{Input state}
%         \label{fig:gol_board_fixed_0}
%     \end{subfigure}
%     \begin{subfigure}{.3\linewidth}
%         \includegraphics[width=\linewidth]{./images/gol_board_fixed_1.png}
%         \caption{State after $1$ step}
%         \label{fig:gol_board_fixed_1}
%     \end{subfigure}
%     \begin{subfigure}{.3\linewidth}
%         \includegraphics[width=\linewidth]{./images/gol_board_fixed_2.png}
%         \caption{State after $2$ steps}
%         \label{fig:gol_board_fixed_2}
%     \end{subfigure}
%     \caption{Manually constructed training board for the Game of Life. \clr{(x13g3)}}
%     \label{fig:gol_board_fixed}
% \end{figure*}

\begin{figure*}[t]
    \centering
    \includegraphics[width=.49\linewidth]{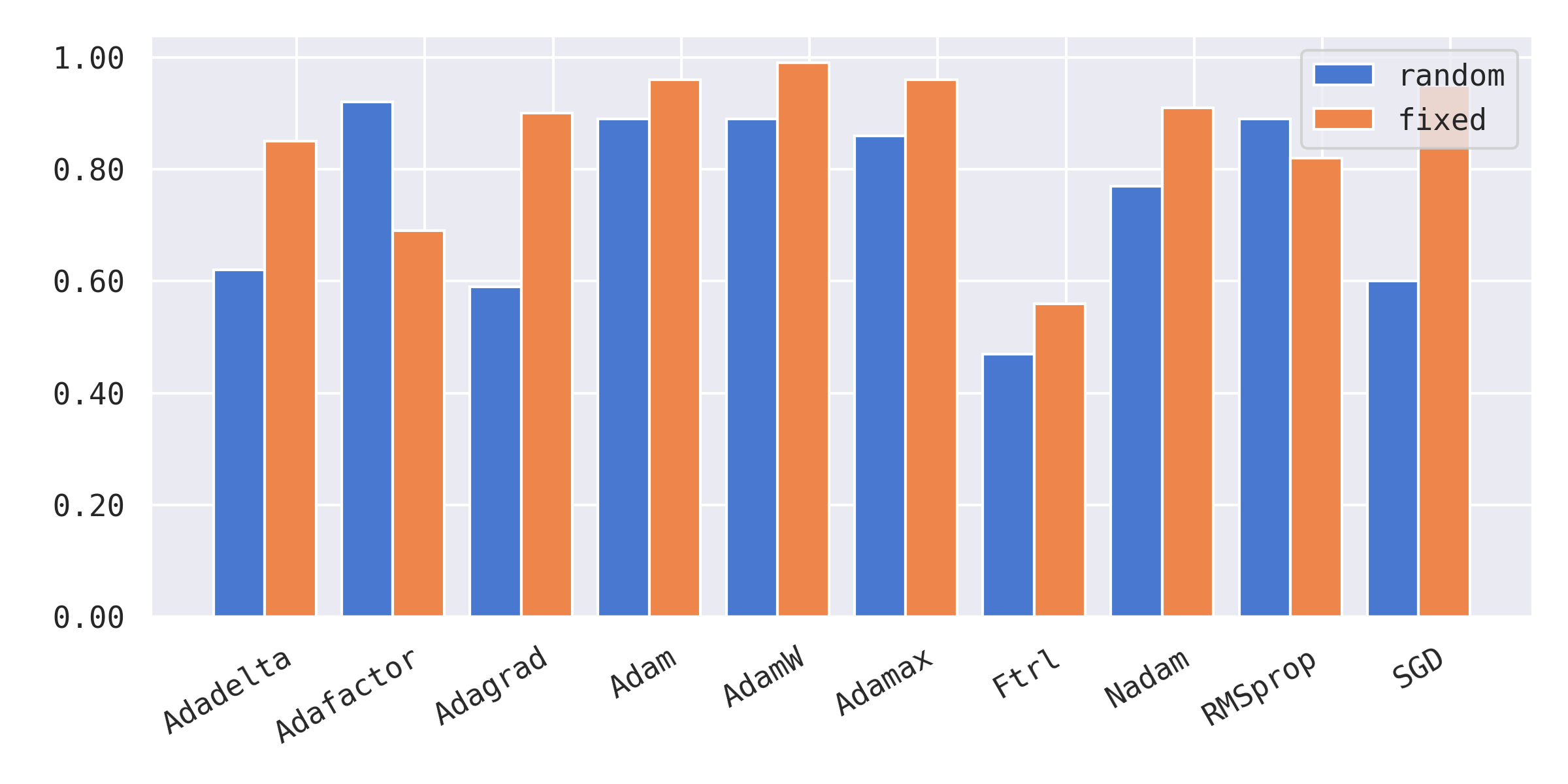}
    \includegraphics[width=.49\linewidth]{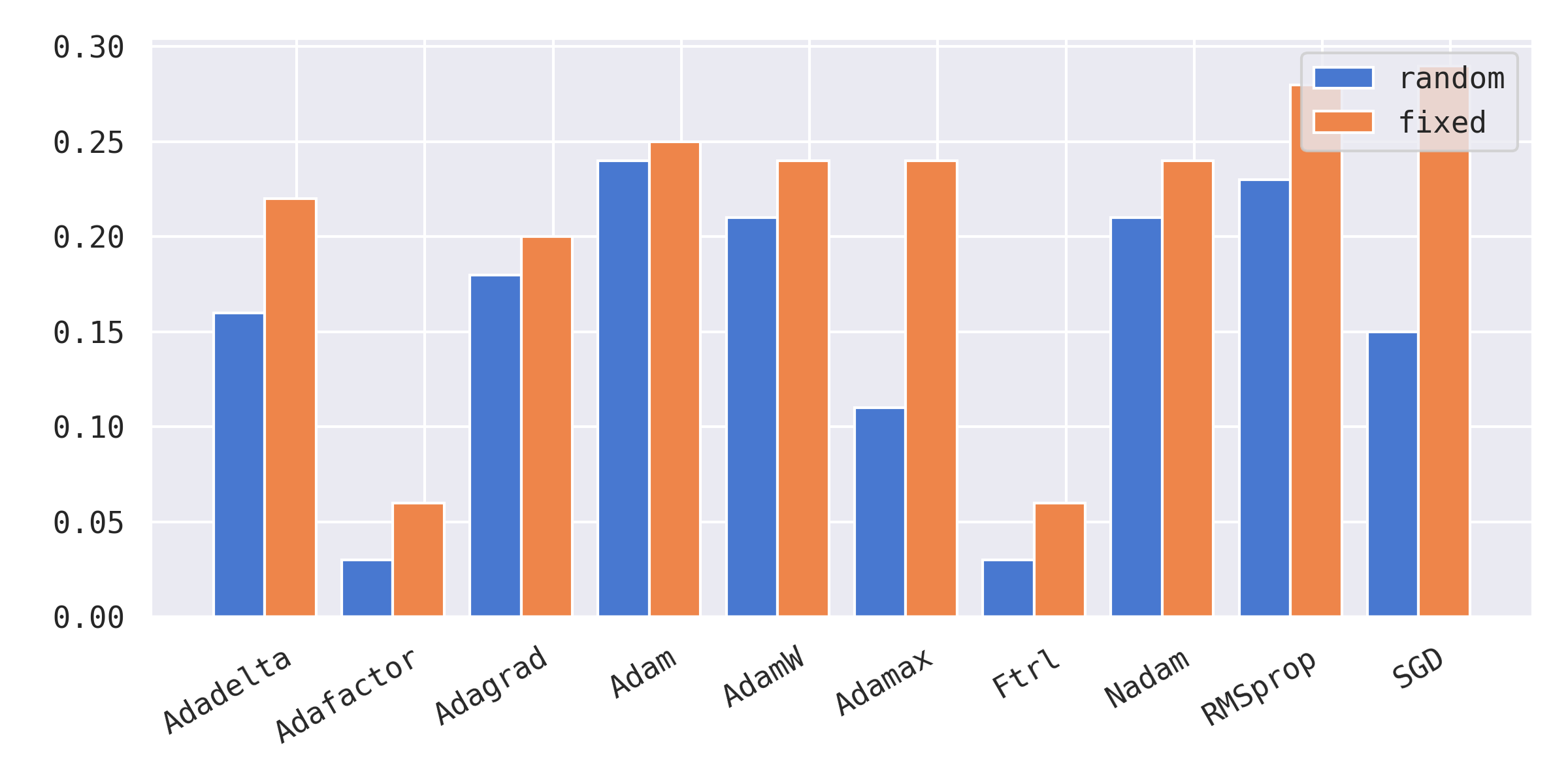}
    \caption{Success rates on 1-step Game of Life for networks with Tanh (left) / ReLU (right) activations.}
    \label{fig:1_step_success}
\end{figure*}

\begin{table*}[t]
    \caption{Results on 1-step Game of Life for networks with Tanh (left) / ReLU (right) activations.}
    \label{tab:1_step}
    \centering\fontsize{8}{9.6}\selectfont
    \begin{tabular}{lrrrrrr}
        \toprule
        & \multicolumn{3}{c}{Success rate} & \multicolumn{3}{c}{Number of epochs}
        \\\cmidrule(lr){2-4}\cmidrule(lr){5-7}
        algorithm & random & fixed & change & random & fixed & change
        \\\midrule
        Adadelta & 0.61 & 0.85 & +39\% & 7449 & 5022 & +33\%
        \\
        Adafactor & 0.92 & 0.69 & -25\% & 2403 & 2590 & -8\%
        \\
        Adagrad & 0.59 & 0.90 & +53\% & 2366 & 1174 & +50\%
        \\
        Adam & 0.89 & 0.96 & +8\% & 402 & 966 & -140\%
        \\
        AdamW & 0.89 & 0.99 & +11\% & 313 & 8992 & -2771\%
        \\
        Adamax & 0.86 & 0.96 & +12\% & 294 & 3078 & -946\%
        \\
        Ftrl & 0.47 & 0.56 & +19\% & 6992 & 1112 & +84\%
        \\
        Nadam & 0.77 & 0.91 & +18\% & 2119 & 1665 & +21\%
        \\
        RMSprop & 0.89 & 0.82 & -8\% & 2414 & 747 & +69\%
        \\
        SGD & 0.60 & 0.95 & +58\% & 866 & 1420 & -64\%
        \\\bottomrule
    \end{tabular}
    \hfill
    \begin{tabular}{lrrrrrr}
        \toprule
        & \multicolumn{3}{c}{Success rate} & \multicolumn{3}{c}{Number of epochs}
        \\\cmidrule(lr){2-4}\cmidrule(lr){5-7}
        algorithm & random & fixed & change & random & fixed & change
        \\\midrule
        Adadelta & 0.16 & 0.22 & +38\% & 5425 & 4657 & +14\%
        \\
        Adafactor & 0.03 & 0.06 & +100\% & 3530 & 5127 & -45\%
        \\
        Adagrad & 0.18 & 0.20 & +11\% & 655 & 1176 & -79\%
        \\
        Adam & 0.24 & 0.25 & +4\% & 1608 & 830 & +48\%
        \\
        AdamW & 0.21 & 0.24 & +14\% & 4975 & 714 & +86\%
        \\
        Adamax & 0.11 & 0.24 & +118\% & 7384 & 807 & +89\%
        \\
        Ftrl & 0.03 & 0.06 & +100\% & 2835 & 1380 & +51\%
        \\
        Nadam & 0.21 & 0.24 & +14\% & 3869 & 2310 & +40\%
        \\
        RMSprop & 0.23 & 0.28 & +22\% & 633 & 9718 & -1436\%
        \\
        SGD & 0.15 & 0.29 & +93\% & 2470 & 1570 & +36\%
        \\\bottomrule
    \end{tabular}
\end{table*}

%%%%%%%%%%%%%%%%%%%%%%%%%%%%%%%%%%%%%%%%%%%%%%%%%%%%%%%%%%%%%%%%%%%%%%%%%%%%%%%%%%%%%%%%%%%%%%%%%%%
\section{Numerical Experiments}\label{sec:numerics}
Our experiments are performed in \texttt{Python~3.8} on a personal laptop.
Training of all the neural networks is done with \texttt{TensorFlow~2.12}.
The source code reproducing the presented experiments is available at~\url{https://github.com/sukiboo/game_of_life}.

In this section we provide an extensive quantitative analysis to showcase the advantage of a data-centric approach.
To this end, we employ all available optimization methods currently implemented in \texttt{Tensorflow~2.12}.
Specifically, we deploy the following optimization algorithms: \texttt{Adadelta}~\cite{zeiler2012adadelta}, \texttt{Adafactor}~\cite{shazeer2018adafactor}, \texttt{Adagrad}~\cite{duchi2011adaptive}, \texttt{Adam} and \texttt{Adamax}~\cite{kingma2014adam}, \texttt{AdamW}~\cite{loshchilov2017decoupled}, \texttt{Ftrl}~\cite{mcmahan2013ad}, \texttt{Nadam}~\cite{dozat2016incorporating}, \texttt{RMSprop}~\cite{hinton2012neural}, and \texttt{SGD} with momentum~\cite{sutskever2013importance}.
The details for each algorithm and the exact values of the hyperparameters are given in Appendix.

In the presented experiments, for each environment and algorithm we perform $100$ learning simulations.
Each simulation consists of deploying an algorithm to train a neural network by minimizing the mean square loss on the provided training set~--- either a sequence of randomly generated boards or a single fixed board.
Each training session is run for $10,000$ epochs, which for our environments we found to be sufficient for an algorithm to either converge or plateau.
All the algorithms are evaluated on the same test set consisting of $100$ randomly generated $100 \times 100$ boards and their corresponding states after $n$ steps of Game of Life.
A simulation is considered successful if the network's prediction accuracy on the test set achieves $100\%$.

For each algorithm we measure the percentage of successful simulations and the average number of epochs required to learn the environment.
In order to evaluate the advantage of this data-centric approach over the conventional ones, for each experiment we report the relative change in \textit{success} (given by the number of successful simulations) and \textit{efficacy} (given by the average number of epochs required to reach convergence).

\paragraph{1-step Game of Life.}
The results of learning $1$-step Game of Life are presented in Figure~\ref{fig:1_step_success} and Table~\ref{tab:1_step}.

\begin{figure*}[t]
    \centering
    \includegraphics[width=.49\linewidth]{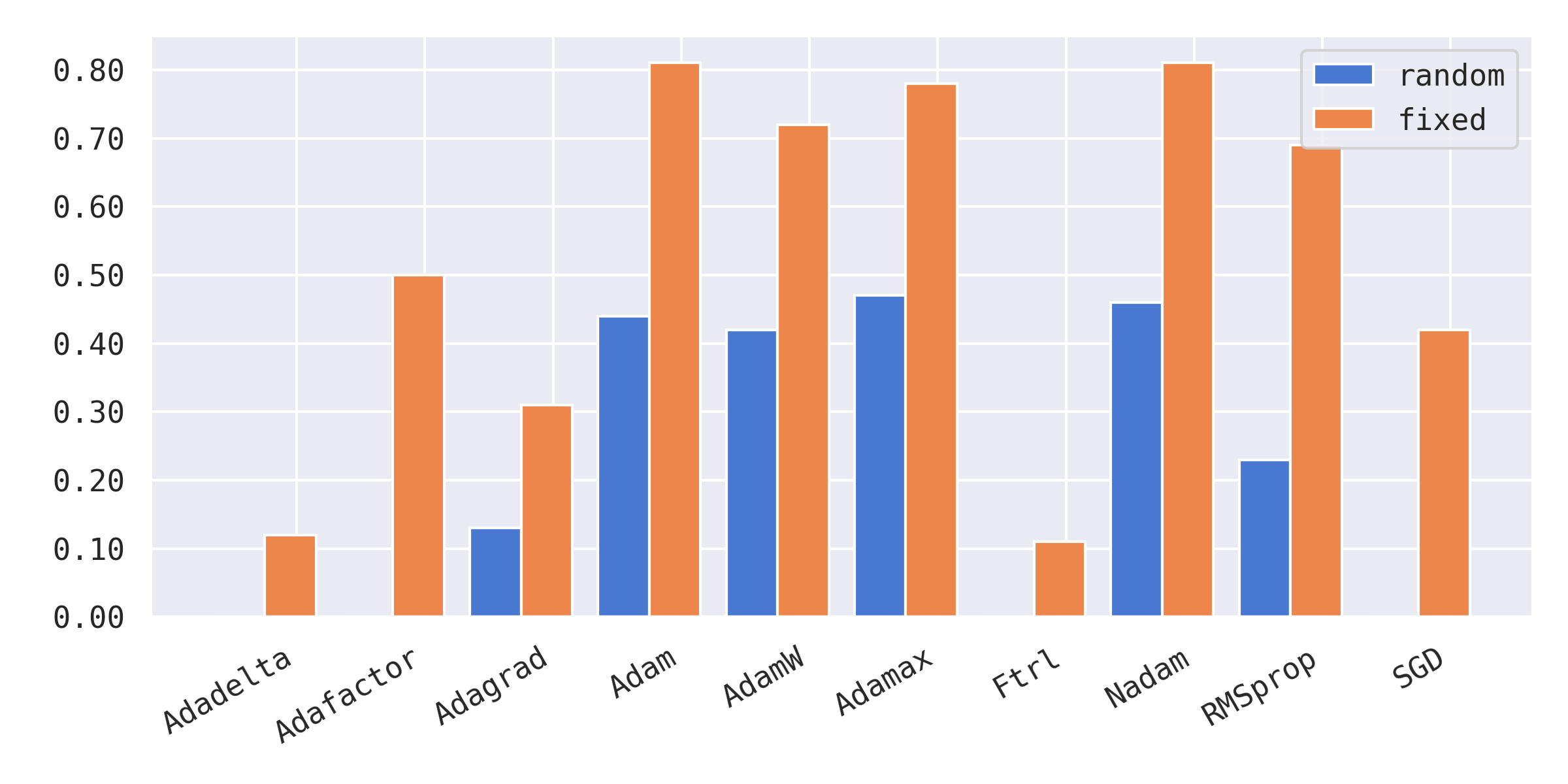}
    \includegraphics[width=.49\linewidth]{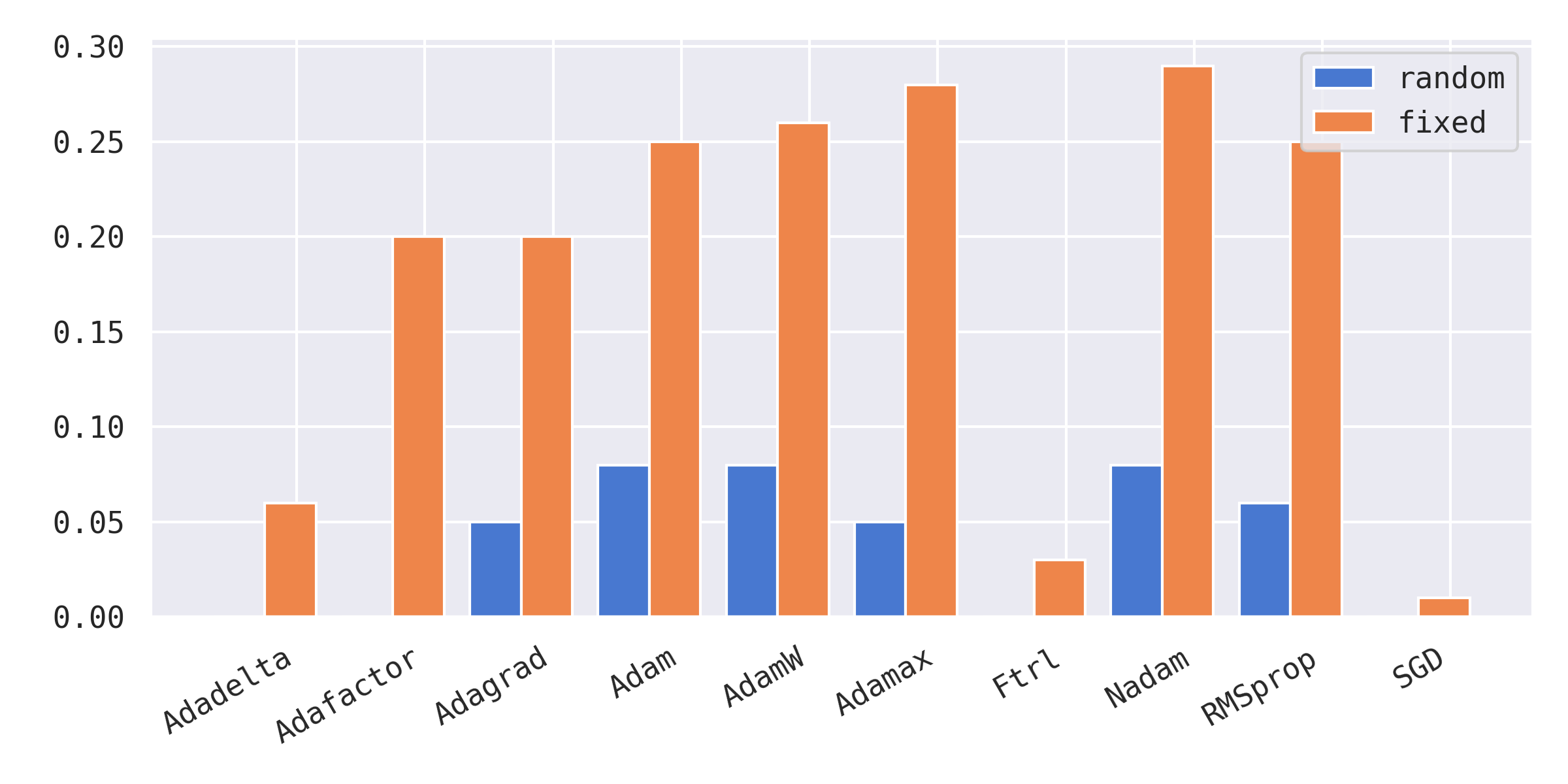}
    \caption{Success rates on 2-step Game of Life with recursive networks with Tanh (left) / ReLU (right) activations.}
    \label{fig:2_step_success_rec}
\end{figure*}

\paragraph{2-step Game of Life with Recursive Networks.}
The results of learning $2$-step Game of Life with recursive (see Figure~\ref{fig:gol_model_recursive}) networks are presented in Figure~\ref{fig:2_step_success_rec} and Table~\ref{tab:2_step_rec}.

\begin{table*}[t]
    \caption{Results on 2-step Game of Life with recursive network and Tanh (left) / ReLU (right) activations.}
    \label{tab:2_step_rec}
    \centering\fontsize{8}{9.6}\selectfont
    \begin{tabular}{lrrrrrr}
        \toprule
        & \multicolumn{3}{c}{Success rate} & \multicolumn{3}{c}{Number of epochs}
        \\\cmidrule(lr){2-4}\cmidrule(lr){5-7}
        algorithm & random & fixed & change & random & fixed & change
        \\\midrule
        Adadelta & --- & 0.19 & --- & --- & 9598 & ---
        \\
        Adafactor & --- & 0.52 & --- & --- & 6839 & ---
        \\
        Adagrad & 0.14 & 0.36 & +157\% & 9649 & 9134 & +5\%
        \\
        Adam & 0.44 & 0.81 & +84\% & 7753 & 4472 & +42\%
        \\
        AdamW & 0.43 & 0.72 & +67\% & 9547 & 3906 & +59\%
        \\
        Adamax & 0.47 & 0.78 & +66\% & 3935 & 5984 & -52\%
        \\
        Ftrl & --- & 0.26 & --- & --- & 9794 & ---
        \\
        Nadam & 0.46 & 0.82 & +78\% & 8441 & 7261 & +14\%
        \\
        RMSprop & 0.26 & 0.69 & +165\% & 9985 & 9981 & +0\%
        \\
        SGD & --- & 0.42 & --- & --- & 8535 & ---
        \\\bottomrule
    \end{tabular}
    \hfill
    \begin{tabular}{lrrrrrr}
        \toprule
        & \multicolumn{3}{c}{Success rate} & \multicolumn{3}{c}{Number of epochs}
        \\\cmidrule(lr){2-4}\cmidrule(lr){5-7}
        algorithm & random & fixed & change & random & fixed & change
        \\\midrule
        Adadelta & --- & 0.08 & --- & --- & 7425 & ---
        \\
        Adafactor & --- & 0.21 & --- & --- & 9119 & ---
        \\
        Adagrad & 0.05 & 0.20 & +300\% & 6023 & 5917 & +2\%
        \\
        Adam & 0.08 & 0.25 & +212\% & 5693 & 7399 & -30\%
        \\
        AdamW & 0.08 & 0.26 & +225\% & 7718 & 2803 & +64\%
        \\
        Adamax & 0.05 & 0.28 & +460\% & 3318 & 5536 & -67\%
        \\
        Ftrl & --- & 0.03 & --- & --- & 7947 & ---
        \\
        Nadam & 0.08 & 0.30 & +275\% & 5524 & 7653 & -39\%
        \\
        RMSprop & 0.06 & 0.25 & +317\% & 9845 & 7151 & +27\%
        \\
        SGD & --- & 0.01 & --- & --- & 4349 & ---
        \\\bottomrule
    \end{tabular}
\end{table*}

\begin{figure*}[t!]
    \centering
    \includegraphics[width=.49\linewidth]{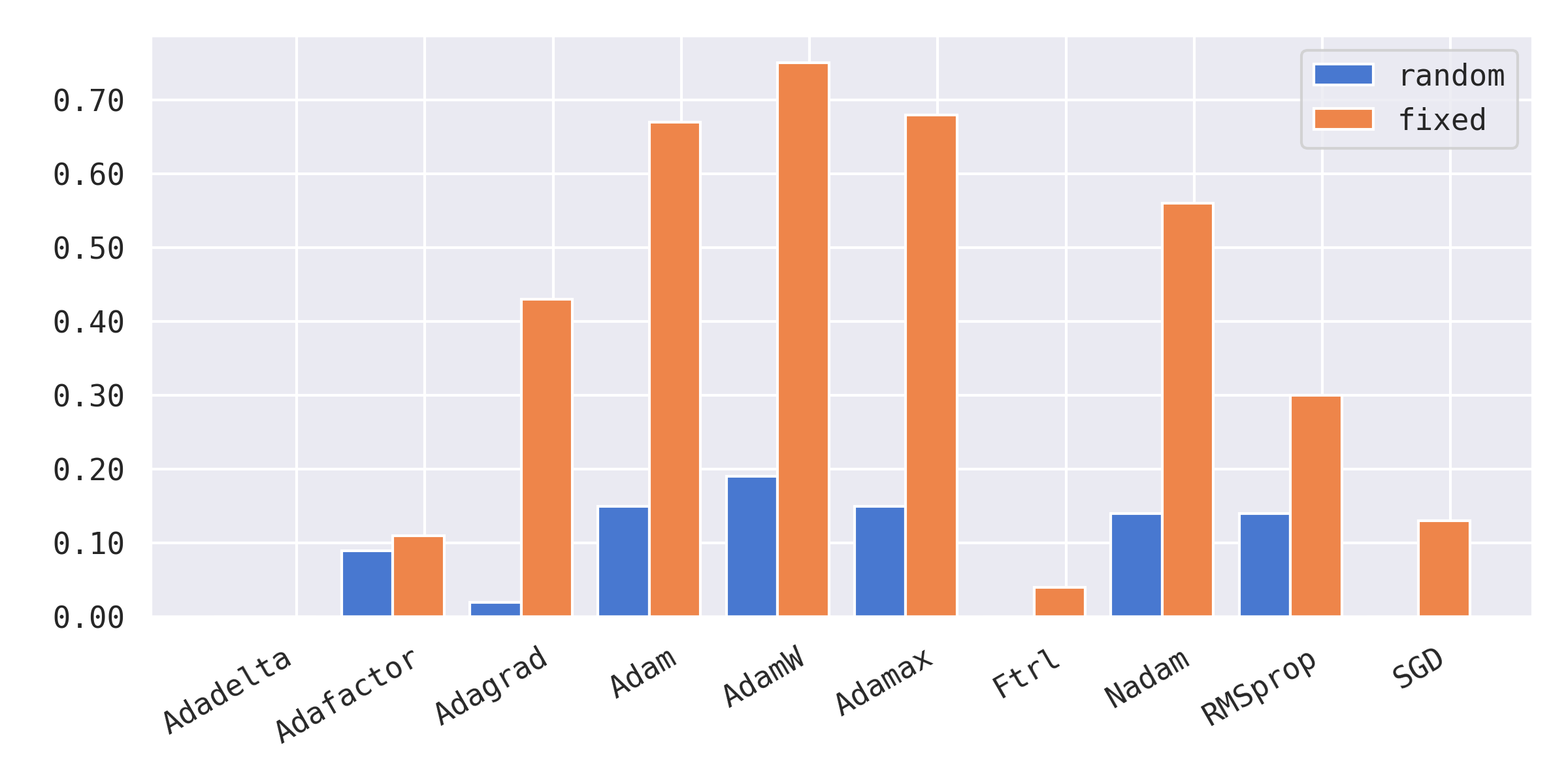}
    \includegraphics[width=.49\linewidth]{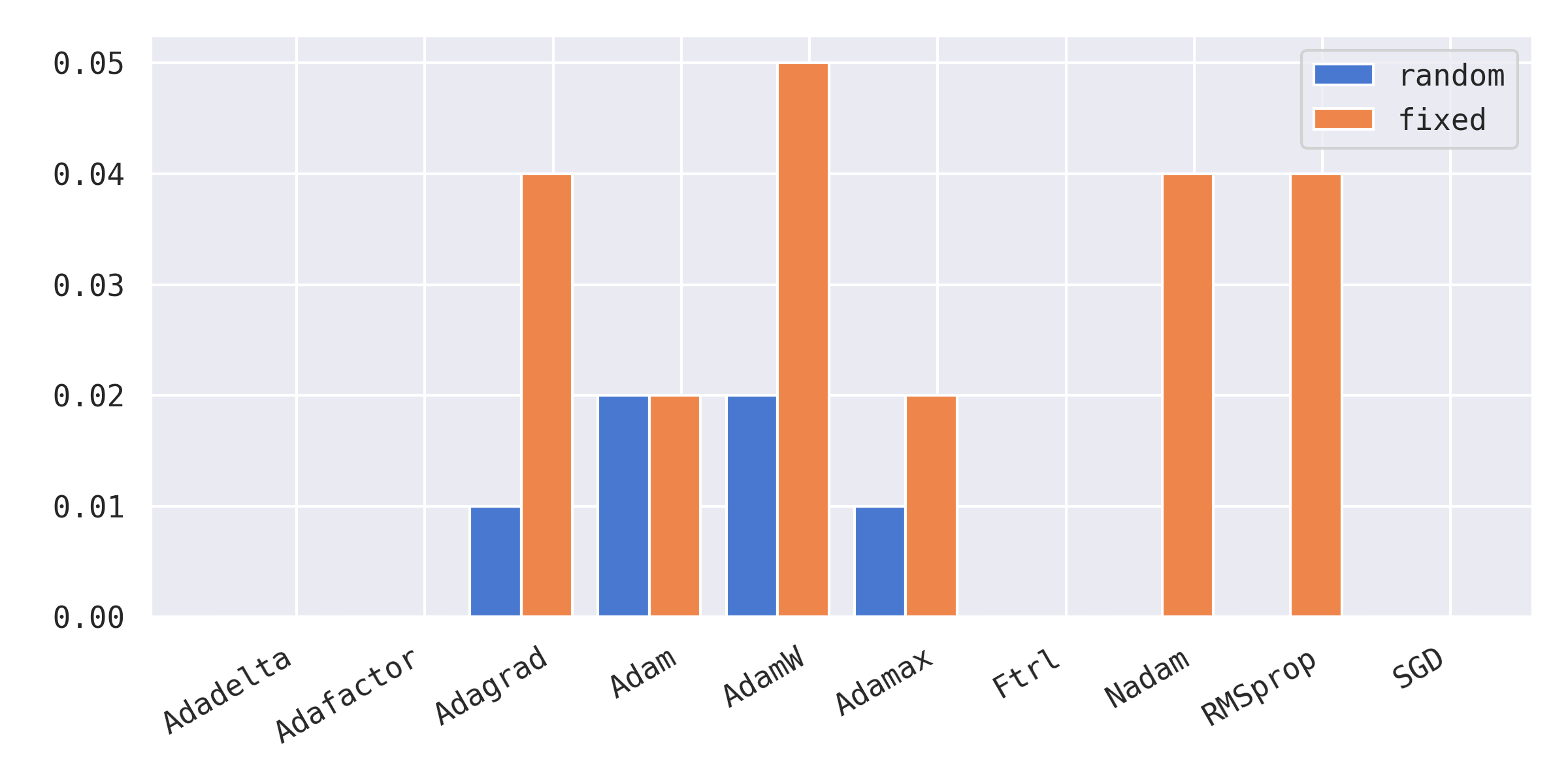}
    \caption{Success rates on 2-step Game of Life with sequential networks with Tanh (left) / ReLU (right) activations.}
    \label{fig:2_step_success_seq}
\end{figure*}

\paragraph{2-step Game of Life with Sequential Networks.}
The results of learning $2$-step Game of Life with sequential (see Figure~\ref{fig:gol_model_sequential}) networks are presented in Figure~\ref{fig:2_step_success_seq} and Table~\ref{tab:2_step_seq}.

\begin{table*}[t!]
    \caption{Results on 2-step Game of Life with sequential network and Tanh (left) / ReLU (right) activations.}
    \label{tab:2_step_seq}
    \centering\fontsize{8}{9.6}\selectfont
    \begin{tabular}{lrrrrrr}
        \toprule
        & \multicolumn{3}{c}{Success rate} & \multicolumn{3}{c}{Number of epochs}
        \\\cmidrule(lr){2-4}\cmidrule(lr){5-7}
        algorithm & random & fixed & change & random & fixed & change
        \\\midrule
        Adadelta & --- & --- & --- & --- & --- & ---
        \\
        Adafactor & 0.08 & 0.11 & +38\% & 9621 & 8969 & +7\%
        \\
        Adagrad & 0.03 & 0.44 & +1367\% & 9640 & 6703 & +30\%
        \\
        Adam & 0.15 & 0.68 & +353\% & 7994 & 7122 & +11\%
        \\
        AdamW & 0.18 & 0.77 & +328\% & 8076 & 4113 & +49\%
        \\
        Adamax & 0.15 & 0.68 & +353\% & 7151 & 3536 & +51\%
        \\
        Ftrl & --- & 0.04 & --- & --- & 9173 & ---
        \\
        Nadam & 0.14 & 0.56 & +300\% & 7976 & 6811 & +15\%
        \\
        RMSprop & 0.13 & 0.31 & +138\% & 9989 & 9270 & +7\%
        \\
        SGD & --- & 0.13 & --- & --- & 9637 & ---
        \\\bottomrule
    \end{tabular}
    \hfill
    \begin{tabular}{lrrrrrr}
        \toprule
        & \multicolumn{3}{c}{Success rate} & \multicolumn{3}{c}{Number of epochs}
        \\\cmidrule(lr){2-4}\cmidrule(lr){5-7}
        algorithm & random & fixed & change & random & fixed & change
        \\\midrule
        Adadelta & --- & --- & --- & --- & --- & ---
        \\
        Adafactor & --- & --- & --- & --- & --- & ---
        \\
        Adagrad & 0.01 & 0.04 & +300\% & 7068 & 7767 & -10\%
        \\
        Adam & 0.02 & 0.02 & +0\% & 5720 & 5694 & +0\%
        \\
        AdamW & 0.02 & 0.05 & +150\% & 8822 & 6780 & +23\%
        \\
        Adamax & 0.01 & 0.02 & +100\% & 6080 & 2936 & +52\%
        \\
        Ftrl & --- & --- & --- & --- & --- & ---
        \\
        Nadam & --- & 0.04 & --- & --- & 7165 & ---
        \\
        RMSprop & --- & 0.04 & --- & --- & 5350 & ---
        \\
        SGD & --- & --- & --- & --- & --- & ---
        \\\bottomrule
    \end{tabular}
\end{table*}

\subsection{Discussion}
We note that the advantage of designing a fixed training board is not necessarily evident in the 1-step prediction setting, and in some cases results in slower convergence.
We hypothesize that this is due to the relative simplicity of the environment, where even the conventional approaches perform well.
However, considering the 2-step prediction settings is where one observes the benefits of using custom designed training data, as is apparent from Figures~\ref{fig:2_step_success_rec}, \ref{fig:2_step_success_seq} and Tables~\ref{tab:2_step_rec}, \ref{tab:2_step_seq}.
Overall, our results highlight the importance of a data-centric approach on challenging tasks where conventional approaches struggle to achieve satisfactory performance.

%%%%%%%%%%%%%%%%%%%%%%%%%%%%%%%%%%%%%%%%%%%%%%%%%%%%%%%%%%%%%%%%%%%%%%%%%%%%%%%%%%%%%%%%%%%%%%%%%%%
\section{Conclusion}
In this work we study the importance of training data for the efficiency of machine learning algorithms in constrained environments, represented by the multi-step Game of Life prediction tasks.
We observe the advantage of a data-centric approach over its conventional counterpart, which becomes especially evident in more challenging environments.

Specifically, we demonstrate that training on a single strategically designed data point can be more beneficial than $10,000$ generated data points in terms of both convergence and efficiency.
Our findings highlight the essential role of data design in constrained machine learning, necessitating a collaborative approach between ML practitioners and domain experts for successful real-world applications.

\subsection{Limitations and Future Research}
In this paper we focus on learning the Game of Life in the minimal architecture, which posses certain limitations.
In particular, we were unable to learn a single instance of $n$-step Game of Life for $n \ge 3$, which is likely due to the \textit{unpredictability} of the state dynamics given by the rules~\eqref{eq:gol_rules}.
Exploring ways to overcome this limitation without sacrificing the minimality is one of the natural future research directions.

While the presented numerical experiments are performed only on the backpropagation algorithms, we observe the similar behavior on the genetic algorithms~\cite{shapiro2001genetic} and the family of the blackbox optimization algorithms based on Gaussian smoothing~\cite{zhang2020scalable}.
For the consistency of presentation we do not report these results in the current work.

We consider the canonical formulation of Conway's Game of Life to act as a benchmarking environment.
It would be of interest to replicate our findings on other versions or generalizations, or other cellular automaton games, e.g.~\cite{peper2010variations}.

%%%%%%%%%%%%%%%%%%%%%%%%%%%%%%%%%%%%%%%%%%%%%%%%%%%%%%%%%%%%%%%%%%%%%%%%%%%%%%%%%%%%%%%%%%%%%%%%%%%
\bibliographystyle{ACM-Reference-Format}
\bibliography{references}


\begin{thebibliography}{67}

%%% the Web address.

\ifx \showCODEN    \undefined \def \showCODEN     #1{\unskip}     \fi
\ifx \showDOI      \undefined \def \showDOI       #1{#1}\fi
\ifx \showISBNx    \undefined \def \showISBNx     #1{\unskip}     \fi
\ifx \showISBNxiii \undefined \def \showISBNxiii  #1{\unskip}     \fi
\ifx \showISSN     \undefined \def \showISSN      #1{\unskip}     \fi
\ifx \showLCCN     \undefined \def \showLCCN      #1{\unskip}     \fi
\ifx \shownote     \undefined \def \shownote      #1{#1}          \fi
\ifx \showarticletitle \undefined \def \showarticletitle #1{#1}   \fi
\ifx \showURL      \undefined \def \showURL       {\relax}        \fi
% invisible to TeX
\providecommand\bibfield[2]{#2}
\providecommand\bibinfo[2]{#2}
\providecommand\natexlab[1]{#1}
\providecommand\showeprint[2][]{arXiv:#2}

\bibitem[Aach et~al\mbox{.}(2021)]%
        {aach2021generalization}
\bibfield{author}{\bibinfo{person}{Marcel Aach}, \bibinfo{person}{Jens~Henrik
  Goebbert}, {and} \bibinfo{person}{Jenia Jitsev}.}
  \bibinfo{year}{2021}\natexlab{}.
\newblock \showarticletitle{Generalization over different cellular automata
  rules learned by a deep feed-forward neural network}.
\newblock \bibinfo{journal}{\emph{arXiv preprint arXiv:2103.14886}}
  (\bibinfo{year}{2021}).
\newblock


\bibitem[Abufadda and Mansour(2021)]%
        {abufadda2021survey}
\bibfield{author}{\bibinfo{person}{Mohammad Abufadda} {and}
  \bibinfo{person}{Khalid Mansour}.} \bibinfo{year}{2021}\natexlab{}.
\newblock \showarticletitle{A survey of synthetic data generation for machine
  learning}. In \bibinfo{booktitle}{\emph{2021 22nd international arab
  conference on information technology (ACIT)}}. IEEE, \bibinfo{pages}{1--7}.
\newblock


\bibitem[Adachi et~al\mbox{.}(2004)]%
        {adachi2004game}
\bibfield{author}{\bibinfo{person}{Susumu Adachi}, \bibinfo{person}{Ferdinand
  Peper}, {and} \bibinfo{person}{Jia Lee}.} \bibinfo{year}{2004}\natexlab{}.
\newblock \showarticletitle{The Game of Life at finite temperature}.
\newblock \bibinfo{journal}{\emph{Physica D: Nonlinear Phenomena}}
  \bibinfo{volume}{198}, \bibinfo{number}{3-4} (\bibinfo{year}{2004}),
  \bibinfo{pages}{182--196}.
\newblock


\bibitem[Adamatzky(2010)]%
        {adamatzky2010game}
\bibfield{author}{\bibinfo{person}{Andrew Adamatzky}.}
  \bibinfo{year}{2010}\natexlab{}.
\newblock \bibinfo{booktitle}{\emph{Game of life cellular automata}}.
  Vol.~\bibinfo{volume}{1}.
\newblock \bibinfo{publisher}{Springer}.
\newblock


\bibitem[Ahmad et~al\mbox{.}(2018)]%
        {ahmad2018interpretable}
\bibfield{author}{\bibinfo{person}{Muhammad~Aurangzeb Ahmad},
  \bibinfo{person}{Carly Eckert}, {and} \bibinfo{person}{Ankur Teredesai}.}
  \bibinfo{year}{2018}\natexlab{}.
\newblock \showarticletitle{Interpretable machine learning in healthcare}. In
  \bibinfo{booktitle}{\emph{Proceedings of the 2018 ACM international
  conference on bioinformatics, computational biology, and health
  informatics}}. \bibinfo{pages}{559--560}.
\newblock


\bibitem[Al~Janabi(2022)]%
        {al2022optimization}
\bibfield{author}{\bibinfo{person}{Mazin~AM Al~Janabi}.}
  \bibinfo{year}{2022}\natexlab{}.
\newblock \showarticletitle{Optimization algorithms and investment portfolio
  analytics with machine learning techniques under time-varying liquidity
  constraints}.
\newblock \bibinfo{journal}{\emph{Journal of Modelling in Management}}
  \bibinfo{volume}{17}, \bibinfo{number}{3} (\bibinfo{year}{2022}),
  \bibinfo{pages}{864--895}.
\newblock


\bibitem[Ammanabrolu and Hausknecht(2020)]%
        {ammanabrolu2020graph}
\bibfield{author}{\bibinfo{person}{Prithviraj Ammanabrolu} {and}
  \bibinfo{person}{Matthew Hausknecht}.} \bibinfo{year}{2020}\natexlab{}.
\newblock \showarticletitle{Graph constrained reinforcement learning for
  natural language action spaces}.
\newblock \bibinfo{journal}{\emph{arXiv preprint arXiv:2001.08837}}
  (\bibinfo{year}{2020}).
\newblock


\bibitem[Anik and Bunt(2021)]%
        {anik2021data}
\bibfield{author}{\bibinfo{person}{Ariful~Islam Anik} {and}
  \bibinfo{person}{Andrea Bunt}.} \bibinfo{year}{2021}\natexlab{}.
\newblock \showarticletitle{Data-centric explanations: explaining training data
  of machine learning systems to promote transparency}. In
  \bibinfo{booktitle}{\emph{Proceedings of the 2021 CHI Conference on Human
  Factors in Computing Systems}}. \bibinfo{pages}{1--13}.
\newblock


\bibitem[Aslanides et~al\mbox{.}(2017)]%
        {aslanides2017universal}
\bibfield{author}{\bibinfo{person}{John Aslanides}, \bibinfo{person}{Jan
  Leike}, {and} \bibinfo{person}{Marcus Hutter}.}
  \bibinfo{year}{2017}\natexlab{}.
\newblock \showarticletitle{Universal reinforcement learning algorithms: Survey
  and experiments}.
\newblock \bibinfo{journal}{\emph{arXiv preprint arXiv:1705.10557}}
  (\bibinfo{year}{2017}).
\newblock


\bibitem[Bae et~al\mbox{.}(2022)]%
        {bae2022constrained}
\bibfield{author}{\bibinfo{person}{Hyeong-Ohk Bae}, \bibinfo{person}{Seung-Yeal
  Ha}, \bibinfo{person}{Myeongju Kang}, \bibinfo{person}{Hyuncheul Lim},
  \bibinfo{person}{Chanho Min}, {and} \bibinfo{person}{Jane Yoo}.}
  \bibinfo{year}{2022}\natexlab{}.
\newblock \showarticletitle{A constrained consensus based optimization
  algorithm and its application to finance}.
\newblock \bibinfo{journal}{\emph{Appl. Math. Comput.}}  \bibinfo{volume}{416}
  (\bibinfo{year}{2022}), \bibinfo{pages}{126726}.
\newblock


\bibitem[Blalock et~al\mbox{.}(2020)]%
        {blalock2020state}
\bibfield{author}{\bibinfo{person}{Davis Blalock}, \bibinfo{person}{Jose~Javier
  Gonzalez~Ortiz}, \bibinfo{person}{Jonathan Frankle}, {and}
  \bibinfo{person}{John Guttag}.} \bibinfo{year}{2020}\natexlab{}.
\newblock \showarticletitle{What is the state of neural network pruning?}
\newblock \bibinfo{journal}{\emph{Proceedings of machine learning and systems}}
   \bibinfo{volume}{2} (\bibinfo{year}{2020}), \bibinfo{pages}{129--146}.
\newblock


\bibitem[Bojun(2020)]%
        {bojun2020steady}
\bibfield{author}{\bibinfo{person}{Huang Bojun}.}
  \bibinfo{year}{2020}\natexlab{}.
\newblock \showarticletitle{Steady state analysis of episodic reinforcement
  learning}.
\newblock \bibinfo{journal}{\emph{Advances in Neural Information Processing
  Systems}}  \bibinfo{volume}{33} (\bibinfo{year}{2020}),
  \bibinfo{pages}{9335--9345}.
\newblock


\bibitem[Chen et~al\mbox{.}(2021)]%
        {chen2021ethical}
\bibfield{author}{\bibinfo{person}{Irene~Y Chen}, \bibinfo{person}{Emma
  Pierson}, \bibinfo{person}{Sherri Rose}, \bibinfo{person}{Shalmali Joshi},
  \bibinfo{person}{Kadija Ferryman}, {and} \bibinfo{person}{Marzyeh Ghassemi}.}
  \bibinfo{year}{2021}\natexlab{}.
\newblock \showarticletitle{Ethical machine learning in healthcare}.
\newblock \bibinfo{journal}{\emph{Annual review of biomedical data science}}
  \bibinfo{volume}{4} (\bibinfo{year}{2021}), \bibinfo{pages}{123--144}.
\newblock


\bibitem[Dalal et~al\mbox{.}(2018)]%
        {dalal2018safe}
\bibfield{author}{\bibinfo{person}{Gal Dalal}, \bibinfo{person}{Krishnamurthy
  Dvijotham}, \bibinfo{person}{Matej Vecerik}, \bibinfo{person}{Todd Hester},
  \bibinfo{person}{Cosmin Paduraru}, {and} \bibinfo{person}{Yuval Tassa}.}
  \bibinfo{year}{2018}\natexlab{}.
\newblock \showarticletitle{Safe exploration in continuous action spaces}.
\newblock \bibinfo{journal}{\emph{arXiv preprint arXiv:1801.08757}}
  (\bibinfo{year}{2018}).
\newblock


\bibitem[Dereventsov et~al\mbox{.}(2021)]%
        {dereventsov2021offline}
\bibfield{author}{\bibinfo{person}{Anton Dereventsov},
  \bibinfo{person}{Joseph~D Daws~Jr}, {and} \bibinfo{person}{Clayton Webster}.}
  \bibinfo{year}{2021}\natexlab{}.
\newblock \showarticletitle{Offline policy comparison under limited historical
  agent-environment interactions}.
\newblock \bibinfo{journal}{\emph{arXiv preprint arXiv:2106.03934}}
  (\bibinfo{year}{2021}).
\newblock


\bibitem[Devarajan et~al\mbox{.}(2021)]%
        {devarajan2021dlio}
\bibfield{author}{\bibinfo{person}{Hariharan Devarajan},
  \bibinfo{person}{Huihuo Zheng}, \bibinfo{person}{Anthony Kougkas},
  \bibinfo{person}{Xian-He Sun}, {and} \bibinfo{person}{Venkatram Vishwanath}.}
  \bibinfo{year}{2021}\natexlab{}.
\newblock \showarticletitle{Dlio: A data-centric benchmark for scientific deep
  learning applications}. In \bibinfo{booktitle}{\emph{2021 IEEE/ACM 21st
  International Symposium on Cluster, Cloud and Internet Computing (CCGrid)}}.
  IEEE, \bibinfo{pages}{81--91}.
\newblock


\bibitem[Dozat(2016)]%
        {dozat2016incorporating}
\bibfield{author}{\bibinfo{person}{Timothy Dozat}.}
  \bibinfo{year}{2016}\natexlab{}.
\newblock \bibinfo{title}{Incorporating {N}esterov momentum into {A}dam}.
\newblock
  \bibinfo{howpublished}{\url{http://cs229.stanford.edu/proj2015/054_report.pdf}}.
\newblock


\bibitem[Dritsas and Trigka(2022)]%
        {dritsas2022data}
\bibfield{author}{\bibinfo{person}{Elias Dritsas} {and} \bibinfo{person}{Maria
  Trigka}.} \bibinfo{year}{2022}\natexlab{}.
\newblock \showarticletitle{Data-driven machine-learning methods for diabetes
  risk prediction}.
\newblock \bibinfo{journal}{\emph{Sensors}} \bibinfo{volume}{22},
  \bibinfo{number}{14} (\bibinfo{year}{2022}), \bibinfo{pages}{5304}.
\newblock


\bibitem[Duchi et~al\mbox{.}(2011)]%
        {duchi2011adaptive}
\bibfield{author}{\bibinfo{person}{John Duchi}, \bibinfo{person}{Elad Hazan},
  {and} \bibinfo{person}{Yoram Singer}.} \bibinfo{year}{2011}\natexlab{}.
\newblock \showarticletitle{Adaptive subgradient methods for online learning
  and stochastic optimization.}
\newblock \bibinfo{journal}{\emph{Journal of machine learning research}}
  \bibinfo{volume}{12}, \bibinfo{number}{7} (\bibinfo{year}{2011}).
\newblock


\bibitem[Dulac-Arnold et~al\mbox{.}(2021)]%
        {dulac2021challenges}
\bibfield{author}{\bibinfo{person}{Gabriel Dulac-Arnold}, \bibinfo{person}{Nir
  Levine}, \bibinfo{person}{Daniel~J Mankowitz}, \bibinfo{person}{Jerry Li},
  \bibinfo{person}{Cosmin Paduraru}, \bibinfo{person}{Sven Gowal}, {and}
  \bibinfo{person}{Todd Hester}.} \bibinfo{year}{2021}\natexlab{}.
\newblock \showarticletitle{Challenges of real-world reinforcement learning:
  definitions, benchmarks and analysis}.
\newblock \bibinfo{journal}{\emph{Machine Learning}} \bibinfo{volume}{110},
  \bibinfo{number}{9} (\bibinfo{year}{2021}), \bibinfo{pages}{2419--2468}.
\newblock


\bibitem[Emmert-Streib and Yli-Harja(2022)]%
        {emmert2022digital}
\bibfield{author}{\bibinfo{person}{Frank Emmert-Streib} {and}
  \bibinfo{person}{Olli Yli-Harja}.} \bibinfo{year}{2022}\natexlab{}.
\newblock \showarticletitle{What Is a Digital Twin? Experimental Design for a
  Data-Centric Machine Learning Perspective in Health}.
\newblock \bibinfo{journal}{\emph{International Journal of Molecular Sciences}}
  \bibinfo{volume}{23}, \bibinfo{number}{21} (\bibinfo{year}{2022}),
  \bibinfo{pages}{13149}.
\newblock


\bibitem[Fu et~al\mbox{.}(2021)]%
        {fu2021benchmarks}
\bibfield{author}{\bibinfo{person}{Justin Fu}, \bibinfo{person}{Mohammad
  Norouzi}, \bibinfo{person}{Ofir Nachum}, \bibinfo{person}{George Tucker},
  \bibinfo{person}{Ziyu Wang}, \bibinfo{person}{Alexander Novikov},
  \bibinfo{person}{Mengjiao Yang}, \bibinfo{person}{Michael~R Zhang},
  \bibinfo{person}{Yutian Chen}, \bibinfo{person}{Aviral Kumar},
  {et~al\mbox{.}}} \bibinfo{year}{2021}\natexlab{}.
\newblock \showarticletitle{Benchmarks for deep off-policy evaluation}.
\newblock \bibinfo{journal}{\emph{arXiv preprint arXiv:2103.16596}}
  (\bibinfo{year}{2021}).
\newblock


\bibitem[Gennatas et~al\mbox{.}(2020)]%
        {gennatas2020expert}
\bibfield{author}{\bibinfo{person}{Efstathios~D Gennatas},
  \bibinfo{person}{Jerome~H Friedman}, \bibinfo{person}{Lyle~H Ungar},
  \bibinfo{person}{Romain Pirracchio}, \bibinfo{person}{Eric Eaton},
  \bibinfo{person}{Lara~G Reichmann}, \bibinfo{person}{Yannet Interian},
  \bibinfo{person}{Jos{\'e}~Marcio Luna}, \bibinfo{person}{Charles~B Simone},
  \bibinfo{person}{Andrew Auerbach}, {et~al\mbox{.}}}
  \bibinfo{year}{2020}\natexlab{}.
\newblock \showarticletitle{Expert-augmented machine learning}.
\newblock \bibinfo{journal}{\emph{Proceedings of the National Academy of
  Sciences}} \bibinfo{volume}{117}, \bibinfo{number}{9} (\bibinfo{year}{2020}),
  \bibinfo{pages}{4571--4577}.
\newblock


\bibitem[Gonz{\'a}lez et~al\mbox{.}(2015)]%
        {gonzalez2015review}
\bibfield{author}{\bibinfo{person}{David Gonz{\'a}lez},
  \bibinfo{person}{Joshu{\'e} P{\'e}rez}, \bibinfo{person}{Vicente
  Milan{\'e}s}, {and} \bibinfo{person}{Fawzi Nashashibi}.}
  \bibinfo{year}{2015}\natexlab{}.
\newblock \showarticletitle{A review of motion planning techniques for
  automated vehicles}.
\newblock \bibinfo{journal}{\emph{IEEE Transactions on intelligent
  transportation systems}} \bibinfo{volume}{17}, \bibinfo{number}{4}
  (\bibinfo{year}{2015}), \bibinfo{pages}{1135--1145}.
\newblock


\bibitem[Goodfellow et~al\mbox{.}(2016)]%
        {goodfellow2016convolutional}
\bibfield{author}{\bibinfo{person}{Ian Goodfellow}, \bibinfo{person}{Yoshua
  Bengio}, {and} \bibinfo{person}{Aaron Courville}.}
  \bibinfo{year}{2016}\natexlab{}.
\newblock \showarticletitle{Convolutional networks}.
\newblock In \bibinfo{booktitle}{\emph{Deep learning}}.
  Vol.~\bibinfo{volume}{2016}. \bibinfo{publisher}{MIT press Cambridge, MA,
  USA}, \bibinfo{pages}{330--372}.
\newblock


\bibitem[Gori et~al\mbox{.}(2023)]%
        {gori2023machine}
\bibfield{author}{\bibinfo{person}{Marco Gori}, \bibinfo{person}{Alessandro
  Betti}, {and} \bibinfo{person}{Stefano Melacci}.}
  \bibinfo{year}{2023}\natexlab{}.
\newblock \bibinfo{booktitle}{\emph{Machine Learning: A constraint-based
  approach}}.
\newblock \bibinfo{publisher}{Elsevier}.
\newblock


\bibitem[Grattarola et~al\mbox{.}(2021)]%
        {grattarola2021learning}
\bibfield{author}{\bibinfo{person}{Daniele Grattarola},
  \bibinfo{person}{Lorenzo Livi}, {and} \bibinfo{person}{Cesare Alippi}.}
  \bibinfo{year}{2021}\natexlab{}.
\newblock \showarticletitle{Learning graph cellular automata}.
\newblock \bibinfo{journal}{\emph{Advances in Neural Information Processing
  Systems}}  \bibinfo{volume}{34} (\bibinfo{year}{2021}),
  \bibinfo{pages}{20983--20994}.
\newblock


\bibitem[Hinton et~al\mbox{.}(2012)]%
        {hinton2012neural}
\bibfield{author}{\bibinfo{person}{Geoffrey Hinton}, \bibinfo{person}{Nitish
  Srivastava}, {and} \bibinfo{person}{Kevin Swersky}.}
  \bibinfo{year}{2012}\natexlab{}.
\newblock \bibinfo{title}{Neural networks for machine learning lecture 6a
  overview of mini-batch gradient descent}.
\newblock
  \bibinfo{howpublished}{\url{http://www.cs.toronto.edu/~tijmen/csc321/slides/lecture_slides_lec6.pdf}}.
\newblock


\bibitem[Hirte(2022)]%
        {hirte2022john}
\bibfield{author}{\bibinfo{person}{Raphael Hirte}.}
  \bibinfo{year}{2022}\natexlab{}.
\newblock \bibinfo{title}{John Horton Conway’s Game of Life: An overview and
  examples}.
\newblock
\newblock


\bibitem[Horvatha et~al\mbox{.}(2023)]%
        {horvatha2023harnessing}
\bibfield{author}{\bibinfo{person}{Blanka Horvatha},
  \bibinfo{person}{Aitor~Muguruza Gonzalezb}, {and} \bibinfo{person}{Mikko~S
  Pakkanenc}.} \bibinfo{year}{2023}\natexlab{}.
\newblock \showarticletitle{Harnessing Quantitative Finance by Data-Centric
  Methods}.
\newblock \bibinfo{journal}{\emph{Machine Learning and Data Sciences for
  Financial Markets: A Guide to Contemporary Practices}}
  (\bibinfo{year}{2023}), \bibinfo{pages}{265}.
\newblock


\bibitem[Isola et~al\mbox{.}(2017)]%
        {isola2017image}
\bibfield{author}{\bibinfo{person}{Phillip Isola}, \bibinfo{person}{Jun-Yan
  Zhu}, \bibinfo{person}{Tinghui Zhou}, {and} \bibinfo{person}{Alexei~A
  Efros}.} \bibinfo{year}{2017}\natexlab{}.
\newblock \showarticletitle{Image-to-image translation with conditional
  adversarial networks}. In \bibinfo{booktitle}{\emph{Proceedings of the IEEE
  conference on computer vision and pattern recognition}}.
  \bibinfo{pages}{1125--1134}.
\newblock


\bibitem[Jo and Gebru(2020)]%
        {jo2020lessons}
\bibfield{author}{\bibinfo{person}{Eun~Seo Jo} {and} \bibinfo{person}{Timnit
  Gebru}.} \bibinfo{year}{2020}\natexlab{}.
\newblock \showarticletitle{Lessons from archives: Strategies for collecting
  sociocultural data in machine learning}. In
  \bibinfo{booktitle}{\emph{Proceedings of the 2020 conference on fairness,
  accountability, and transparency}}. \bibinfo{pages}{306--316}.
\newblock


\bibitem[Kingma and Ba(2014)]%
        {kingma2014adam}
\bibfield{author}{\bibinfo{person}{Diederik~P Kingma} {and}
  \bibinfo{person}{Jimmy Ba}.} \bibinfo{year}{2014}\natexlab{}.
\newblock \showarticletitle{Adam: A method for stochastic optimization}.
\newblock \bibinfo{journal}{\emph{arXiv preprint arXiv:1412.6980}}
  (\bibinfo{year}{2014}).
\newblock


\bibitem[Krechetov(2021)]%
        {krechetov2021game}
\bibfield{author}{\bibinfo{person}{Mikhail Krechetov}.}
  \bibinfo{year}{2021}\natexlab{}.
\newblock \showarticletitle{Game of life on graphs}.
\newblock \bibinfo{journal}{\emph{arXiv preprint arXiv:2111.01780}}
  (\bibinfo{year}{2021}).
\newblock


\bibitem[Li and Dong(2022)]%
        {li2022big}
\bibfield{author}{\bibinfo{person}{Zhichao Li} {and} \bibinfo{person}{Jinwei
  Dong}.} \bibinfo{year}{2022}\natexlab{}.
\newblock \showarticletitle{Big Geospatial Data and Data-Driven Methods for
  Urban Dengue Risk Forecasting: A Review}.
\newblock \bibinfo{journal}{\emph{Remote Sensing}} \bibinfo{volume}{14},
  \bibinfo{number}{19} (\bibinfo{year}{2022}), \bibinfo{pages}{5052}.
\newblock


\bibitem[Liu et~al\mbox{.}(2022)]%
        {liu2022finrl}
\bibfield{author}{\bibinfo{person}{Xiao-Yang Liu}, \bibinfo{person}{Ziyi Xia},
  \bibinfo{person}{Jingyang Rui}, \bibinfo{person}{Jiechao Gao},
  \bibinfo{person}{Hongyang Yang}, \bibinfo{person}{Ming Zhu},
  \bibinfo{person}{Christina Wang}, \bibinfo{person}{Zhaoran Wang}, {and}
  \bibinfo{person}{Jian Guo}.} \bibinfo{year}{2022}\natexlab{}.
\newblock \showarticletitle{FinRL-Meta: Market environments and benchmarks for
  data-driven financial reinforcement learning}.
\newblock \bibinfo{journal}{\emph{Advances in Neural Information Processing
  Systems}}  \bibinfo{volume}{35} (\bibinfo{year}{2022}),
  \bibinfo{pages}{1835--1849}.
\newblock


\bibitem[Loshchilov and Hutter(2017)]%
        {loshchilov2017decoupled}
\bibfield{author}{\bibinfo{person}{Ilya Loshchilov} {and}
  \bibinfo{person}{Frank Hutter}.} \bibinfo{year}{2017}\natexlab{}.
\newblock \showarticletitle{Decoupled weight decay regularization}.
\newblock \bibinfo{journal}{\emph{arXiv preprint arXiv:1711.05101}}
  (\bibinfo{year}{2017}).
\newblock


\bibitem[Majeed and Hwang(2023)]%
        {majeed2023data}
\bibfield{author}{\bibinfo{person}{Abdul Majeed} {and}
  \bibinfo{person}{Seong~Oun Hwang}.} \bibinfo{year}{2023}\natexlab{}.
\newblock \showarticletitle{Data-Centric Artificial Intelligence,
  Preprocessing, and the Quest for Transformative Artificial Intelligence
  Systems Development}.
\newblock \bibinfo{journal}{\emph{Computer}} \bibinfo{volume}{56},
  \bibinfo{number}{5} (\bibinfo{year}{2023}), \bibinfo{pages}{109--115}.
\newblock


\bibitem[McMahan et~al\mbox{.}(2013)]%
        {mcmahan2013ad}
\bibfield{author}{\bibinfo{person}{H~Brendan McMahan}, \bibinfo{person}{Gary
  Holt}, \bibinfo{person}{David Sculley}, \bibinfo{person}{Michael Young},
  \bibinfo{person}{Dietmar Ebner}, \bibinfo{person}{Julian Grady},
  \bibinfo{person}{Lan Nie}, \bibinfo{person}{Todd Phillips},
  \bibinfo{person}{Eugene Davydov}, \bibinfo{person}{Daniel Golovin},
  {et~al\mbox{.}}} \bibinfo{year}{2013}\natexlab{}.
\newblock \showarticletitle{Ad click prediction: a view from the trenches}. In
  \bibinfo{booktitle}{\emph{Proceedings of the 19th ACM SIGKDD international
  conference on Knowledge discovery and data mining}}.
  \bibinfo{pages}{1222--1230}.
\newblock


\bibitem[Miranda(2021)]%
        {miranda2021towards}
\bibfield{author}{\bibinfo{person}{Lester~James Miranda}.}
  \bibinfo{year}{2021}\natexlab{}.
\newblock \showarticletitle{Towards data-centric machine learning: a short
  review}.
\newblock \bibinfo{journal}{\emph{ljvmiranda921.github.io}}
  (\bibinfo{year}{2021}).
\newblock


\bibitem[Motamedi et~al\mbox{.}(2021)]%
        {motamedi2021data}
\bibfield{author}{\bibinfo{person}{Mohammad Motamedi}, \bibinfo{person}{Nikolay
  Sakharnykh}, {and} \bibinfo{person}{Tim Kaldewey}.}
  \bibinfo{year}{2021}\natexlab{}.
\newblock \showarticletitle{A data-centric approach for training deep neural
  networks with less data}.
\newblock \bibinfo{journal}{\emph{arXiv preprint arXiv:2110.03613}}
  (\bibinfo{year}{2021}).
\newblock


\bibitem[Murez et~al\mbox{.}(2018)]%
        {murez2018image}
\bibfield{author}{\bibinfo{person}{Zak Murez}, \bibinfo{person}{Soheil
  Kolouri}, \bibinfo{person}{David Kriegman}, \bibinfo{person}{Ravi
  Ramamoorthi}, {and} \bibinfo{person}{Kyungnam Kim}.}
  \bibinfo{year}{2018}\natexlab{}.
\newblock \showarticletitle{Image to image translation for domain adaptation}.
  In \bibinfo{booktitle}{\emph{Proceedings of the IEEE conference on computer
  vision and pattern recognition}}. \bibinfo{pages}{4500--4509}.
\newblock


\bibitem[Nguyen et~al\mbox{.}(2021)]%
        {nguyen2021budget}
\bibfield{author}{\bibinfo{person}{Sam Nguyen}, \bibinfo{person}{Ryan Chan},
  \bibinfo{person}{Jose Cadena}, \bibinfo{person}{Braden Soper},
  \bibinfo{person}{Paul Kiszka}, \bibinfo{person}{Lucas Womack},
  \bibinfo{person}{Mark Work}, \bibinfo{person}{Joan~M Duggan},
  \bibinfo{person}{Steven~T Haller}, \bibinfo{person}{Jennifer~A Hanrahan},
  {et~al\mbox{.}}} \bibinfo{year}{2021}\natexlab{}.
\newblock \showarticletitle{Budget constrained machine learning for early
  prediction of adverse outcomes for COVID-19 patients}.
\newblock \bibinfo{journal}{\emph{Scientific Reports}} \bibinfo{volume}{11},
  \bibinfo{number}{1} (\bibinfo{year}{2021}), \bibinfo{pages}{19543}.
\newblock


\bibitem[Nye and Saxe(2018)]%
        {nye2018efficient}
\bibfield{author}{\bibinfo{person}{Maxwell Nye} {and} \bibinfo{person}{Andrew
  Saxe}.} \bibinfo{year}{2018}\natexlab{}.
\newblock \showarticletitle{Are efficient deep representations learnable?}
\newblock \bibinfo{journal}{\emph{arXiv preprint arXiv:1807.06399}}
  (\bibinfo{year}{2018}).
\newblock


\bibitem[Pan et~al\mbox{.}(2022)]%
        {pan2022data}
\bibfield{author}{\bibinfo{person}{Indranil Pan}, \bibinfo{person}{Lachlan~R
  Mason}, {and} \bibinfo{person}{Omar~K Matar}.}
  \bibinfo{year}{2022}\natexlab{}.
\newblock \showarticletitle{Data-centric Engineering: integrating simulation,
  machine learning and statistics. Challenges and opportunities}.
\newblock \bibinfo{journal}{\emph{Chemical Engineering Science}}
  \bibinfo{volume}{249} (\bibinfo{year}{2022}), \bibinfo{pages}{117271}.
\newblock


\bibitem[Pang et~al\mbox{.}(2021)]%
        {pang2021image}
\bibfield{author}{\bibinfo{person}{Yingxue Pang}, \bibinfo{person}{Jianxin
  Lin}, \bibinfo{person}{Tao Qin}, {and} \bibinfo{person}{Zhibo Chen}.}
  \bibinfo{year}{2021}\natexlab{}.
\newblock \showarticletitle{Image-to-image translation: Methods and
  applications}.
\newblock \bibinfo{journal}{\emph{IEEE Transactions on Multimedia}}
  \bibinfo{volume}{24} (\bibinfo{year}{2021}), \bibinfo{pages}{3859--3881}.
\newblock


\bibitem[Peper et~al\mbox{.}(2010)]%
        {peper2010variations}
\bibfield{author}{\bibinfo{person}{Ferdinand Peper}, \bibinfo{person}{Susumu
  Adachi}, {and} \bibinfo{person}{Jia Lee}.} \bibinfo{year}{2010}\natexlab{}.
\newblock \showarticletitle{Variations on the game of life}.
\newblock In \bibinfo{booktitle}{\emph{Game of Life Cellular Automata}}.
  \bibinfo{publisher}{Springer}, \bibinfo{pages}{235--255}.
\newblock


\bibitem[Perez et~al\mbox{.}(2021)]%
        {perez2021constrained}
\bibfield{author}{\bibinfo{person}{Guillaume Perez}, \bibinfo{person}{Sebastian
  Ament}, \bibinfo{person}{Carla Gomes}, {and} \bibinfo{person}{Arnaud
  Lallouet}.} \bibinfo{year}{2021}\natexlab{}.
\newblock \showarticletitle{Constrained Machine Learning: The Bagel Framework}.
\newblock \bibinfo{journal}{\emph{arXiv preprint arXiv:2112.01088}}
  (\bibinfo{year}{2021}).
\newblock


\bibitem[Rafler(2011)]%
        {rafler2011generalization}
\bibfield{author}{\bibinfo{person}{Stephan Rafler}.}
  \bibinfo{year}{2011}\natexlab{}.
\newblock \showarticletitle{Generalization of Conway's" Game of Life" to a
  continuous domain-SmoothLife}.
\newblock \bibinfo{journal}{\emph{arXiv preprint arXiv:1111.1567}}
  (\bibinfo{year}{2011}).
\newblock


\bibitem[Ratner et~al\mbox{.}(2016)]%
        {ratner2016data}
\bibfield{author}{\bibinfo{person}{Alexander~J Ratner},
  \bibinfo{person}{Christopher~M De~Sa}, \bibinfo{person}{Sen Wu},
  \bibinfo{person}{Daniel Selsam}, {and} \bibinfo{person}{Christopher R{\'e}}.}
  \bibinfo{year}{2016}\natexlab{}.
\newblock \showarticletitle{Data programming: Creating large training sets,
  quickly}.
\newblock \bibinfo{journal}{\emph{Advances in neural information processing
  systems}}  \bibinfo{volume}{29} (\bibinfo{year}{2016}).
\newblock


\bibitem[Rendell(2011)]%
        {rendell2011universal}
\bibfield{author}{\bibinfo{person}{Paul Rendell}.}
  \bibinfo{year}{2011}\natexlab{}.
\newblock \showarticletitle{A universal turing machine in conway's game of
  life}. In \bibinfo{booktitle}{\emph{2011 International Conference on High
  Performance Computing \& Simulation}}. IEEE, \bibinfo{pages}{764--772}.
\newblock


\bibitem[Rennard(2002)]%
        {rennard2002implementation}
\bibfield{author}{\bibinfo{person}{Jean-Philippe Rennard}.}
  \bibinfo{year}{2002}\natexlab{}.
\newblock \showarticletitle{Implementation of logical functions in the Game of
  Life}.
\newblock In \bibinfo{booktitle}{\emph{Collision-based computing}}.
  \bibinfo{publisher}{Springer}, \bibinfo{pages}{491--512}.
\newblock


\bibitem[Roh et~al\mbox{.}(2019)]%
        {roh2019survey}
\bibfield{author}{\bibinfo{person}{Yuji Roh}, \bibinfo{person}{Geon Heo}, {and}
  \bibinfo{person}{Steven~Euijong Whang}.} \bibinfo{year}{2019}\natexlab{}.
\newblock \showarticletitle{A survey on data collection for machine learning: a
  big data-ai integration perspective}.
\newblock \bibinfo{journal}{\emph{IEEE Transactions on Knowledge and Data
  Engineering}} \bibinfo{volume}{33}, \bibinfo{number}{4}
  (\bibinfo{year}{2019}), \bibinfo{pages}{1328--1347}.
\newblock


\bibitem[Seedat et~al\mbox{.}(2022)]%
        {seedat2022dc}
\bibfield{author}{\bibinfo{person}{Nabeel Seedat}, \bibinfo{person}{Fergus
  Imrie}, {and} \bibinfo{person}{Mihaela van~der Schaar}.}
  \bibinfo{year}{2022}\natexlab{}.
\newblock \showarticletitle{DC-Check: A Data-Centric AI checklist to guide the
  development of reliable machine learning systems}.
\newblock \bibinfo{journal}{\emph{arXiv preprint arXiv:2211.05764}}
  (\bibinfo{year}{2022}).
\newblock


\bibitem[Shapiro(2001)]%
        {shapiro2001genetic}
\bibfield{author}{\bibinfo{person}{Jonathan Shapiro}.}
  \bibinfo{year}{2001}\natexlab{}.
\newblock \showarticletitle{Genetic algorithms in machine learning}.
\newblock \bibinfo{journal}{\emph{Machine Learning and Its Applications:
  Advanced Lectures}} (\bibinfo{year}{2001}), \bibinfo{pages}{146--168}.
\newblock


\bibitem[Shazeer and Stern(2018)]%
        {shazeer2018adafactor}
\bibfield{author}{\bibinfo{person}{Noam Shazeer} {and}
  \bibinfo{person}{Mitchell Stern}.} \bibinfo{year}{2018}\natexlab{}.
\newblock \showarticletitle{Adafactor: Adaptive learning rates with sublinear
  memory cost}. In \bibinfo{booktitle}{\emph{International Conference on
  Machine Learning}}. PMLR, \bibinfo{pages}{4596--4604}.
\newblock


\bibitem[Springer and Kenyon(2021)]%
        {springer2021s}
\bibfield{author}{\bibinfo{person}{Jacob~M Springer} {and}
  \bibinfo{person}{Garrett~T Kenyon}.} \bibinfo{year}{2021}\natexlab{}.
\newblock \showarticletitle{It's hard for neural networks to learn the game of
  life}. In \bibinfo{booktitle}{\emph{2021 International Joint Conference on
  Neural Networks (IJCNN)}}. IEEE, \bibinfo{pages}{1--8}.
\newblock


\bibitem[Sutskever et~al\mbox{.}(2013)]%
        {sutskever2013importance}
\bibfield{author}{\bibinfo{person}{Ilya Sutskever}, \bibinfo{person}{James
  Martens}, \bibinfo{person}{George Dahl}, {and} \bibinfo{person}{Geoffrey
  Hinton}.} \bibinfo{year}{2013}\natexlab{}.
\newblock \showarticletitle{On the importance of initialization and momentum in
  deep learning}. In \bibinfo{booktitle}{\emph{International conference on
  machine learning}}. PMLR, \bibinfo{pages}{1139--1147}.
\newblock


\bibitem[Tennenholtz et~al\mbox{.}(2020)]%
        {tennenholtz2020off}
\bibfield{author}{\bibinfo{person}{Guy Tennenholtz}, \bibinfo{person}{Uri
  Shalit}, {and} \bibinfo{person}{Shie Mannor}.}
  \bibinfo{year}{2020}\natexlab{}.
\newblock \showarticletitle{Off-policy evaluation in partially observable
  environments}. In \bibinfo{booktitle}{\emph{Proceedings of the AAAI
  Conference on Artificial Intelligence}}, Vol.~\bibinfo{volume}{34}.
\newblock


\bibitem[Winoto et~al\mbox{.}(2020)]%
        {winoto2020small}
\bibfield{author}{\bibinfo{person}{Amadeus~Suryo Winoto},
  \bibinfo{person}{Michael Kristianus}, {and} \bibinfo{person}{Chinthaka
  Premachandra}.} \bibinfo{year}{2020}\natexlab{}.
\newblock \showarticletitle{Small and slim deep convolutional neural network
  for mobile device}.
\newblock \bibinfo{journal}{\emph{IEEE Access}}  \bibinfo{volume}{8}
  (\bibinfo{year}{2020}), \bibinfo{pages}{125210--125222}.
\newblock


\bibitem[Yang et~al\mbox{.}(2021)]%
        {yang2021safe}
\bibfield{author}{\bibinfo{person}{Tsung-Yen Yang}, \bibinfo{person}{Michael~Y
  Hu}, \bibinfo{person}{Yinlam Chow}, \bibinfo{person}{Peter~J Ramadge}, {and}
  \bibinfo{person}{Karthik Narasimhan}.} \bibinfo{year}{2021}\natexlab{}.
\newblock \showarticletitle{Safe reinforcement learning with natural language
  constraints}.
\newblock \bibinfo{journal}{\emph{Advances in Neural Information Processing
  Systems}}  \bibinfo{volume}{34} (\bibinfo{year}{2021}),
  \bibinfo{pages}{13794--13808}.
\newblock


\bibitem[Yao et~al\mbox{.}(2021)]%
        {yao2021power}
\bibfield{author}{\bibinfo{person}{Jiayu Yao}, \bibinfo{person}{Emma
  Brunskill}, \bibinfo{person}{Weiwei Pan}, \bibinfo{person}{Susan Murphy},
  {and} \bibinfo{person}{Finale Doshi-Velez}.} \bibinfo{year}{2021}\natexlab{}.
\newblock \showarticletitle{Power constrained bandits}. In
  \bibinfo{booktitle}{\emph{Machine Learning for Healthcare Conference}}. PMLR,
  \bibinfo{pages}{209--259}.
\newblock


\bibitem[Zahid et~al\mbox{.}(2021)]%
        {zahid2021systematic}
\bibfield{author}{\bibinfo{person}{Arnob Zahid}, \bibinfo{person}{Jennifer~Kay
  Poulsen}, \bibinfo{person}{Ravi Sharma}, {and} \bibinfo{person}{Stephen~C
  Wingreen}.} \bibinfo{year}{2021}\natexlab{}.
\newblock \showarticletitle{A systematic review of emerging information
  technologies for sustainable data-centric health-care}.
\newblock \bibinfo{journal}{\emph{International Journal of Medical
  Informatics}}  \bibinfo{volume}{149} (\bibinfo{year}{2021}),
  \bibinfo{pages}{104420}.
\newblock


\bibitem[Zeiler(2012)]%
        {zeiler2012adadelta}
\bibfield{author}{\bibinfo{person}{Matthew~D Zeiler}.}
  \bibinfo{year}{2012}\natexlab{}.
\newblock \showarticletitle{Adadelta: an adaptive learning rate method}.
\newblock \bibinfo{journal}{\emph{arXiv preprint arXiv:1212.5701}}
  (\bibinfo{year}{2012}).
\newblock


\bibitem[Zha et~al\mbox{.}(2023)]%
        {zha2023data}
\bibfield{author}{\bibinfo{person}{Daochen Zha}, \bibinfo{person}{Zaid~Pervaiz
  Bhat}, \bibinfo{person}{Kwei-Herng Lai}, \bibinfo{person}{Fan Yang}, {and}
  \bibinfo{person}{Xia Hu}.} \bibinfo{year}{2023}\natexlab{}.
\newblock \showarticletitle{Data-centric AI: Perspectives and Challenges}.
\newblock \bibinfo{journal}{\emph{arXiv preprint arXiv:2301.04819}}
  (\bibinfo{year}{2023}).
\newblock


\bibitem[Zhang et~al\mbox{.}(2022)]%
        {zhang2022survey}
\bibfield{author}{\bibinfo{person}{Hanqing Zhang}, \bibinfo{person}{Haolin
  Song}, \bibinfo{person}{Shaoyu Li}, \bibinfo{person}{Ming Zhou}, {and}
  \bibinfo{person}{Dawei Song}.} \bibinfo{year}{2022}\natexlab{}.
\newblock \showarticletitle{A survey of controllable text generation using
  transformer-based pre-trained language models}.
\newblock \bibinfo{journal}{\emph{arXiv preprint arXiv:2201.05337}}
  (\bibinfo{year}{2022}).
\newblock


\bibitem[Zhang et~al\mbox{.}(2020)]%
        {zhang2020scalable}
\bibfield{author}{\bibinfo{person}{Jiaxin Zhang}, \bibinfo{person}{Hoang Tran},
  \bibinfo{person}{Dan Lu}, {and} \bibinfo{person}{Guannan Zhang}.}
  \bibinfo{year}{2020}\natexlab{}.
\newblock \showarticletitle{A scalable evolution strategy with directional
  Gaussian smoothing for blackbox optimization}.
\newblock \bibinfo{journal}{\emph{arXiv preprint}} (\bibinfo{year}{2020}).
\newblock


\end{thebibliography}

%%%%%%%%%%%%%%%%%%%%%%%%%%%%%%%%%%%%%%%%%%%%%%%%%%%%%%%%%%%%%%%%%%%%%%%%%%%%%%%%%%%%%%%%%%%%%%%%%%%
\clearpage
\appendix

%%%%%%%%%%%%%%%%%%%%%%%%%%%%%%%%%%%%%%%%%%%%%%%%%%%%%%%%%%%%%%%%%%%%%%%%%%%%%%%%%%%%%%%%%%%%%%%%%%%
\section{Hyperparameter Search}\label{sec:algorithms}
For each algorithm and environment we perform an extensive hyperparameter search to find the appropriate value of the learning rate $\lambda$.
Specifically, we perform an exhaustive grid search over the values
\[
    \lambda \in \{\texttt{1e-1}, \texttt{3e-2}, \texttt{1e-2}, \texttt{3e-3},
    \texttt{1e-3}, \texttt{3e-4}, \texttt{1e-4}\}
\]
by running $100$ tests and measuring their corresponding convergence rates.
In the case of a tie, the smaller learning rate was used.

\begin{figure*}[ht!]
    \centering
    \begin{subfigure}{.49\linewidth}
        \includegraphics[width=\linewidth]{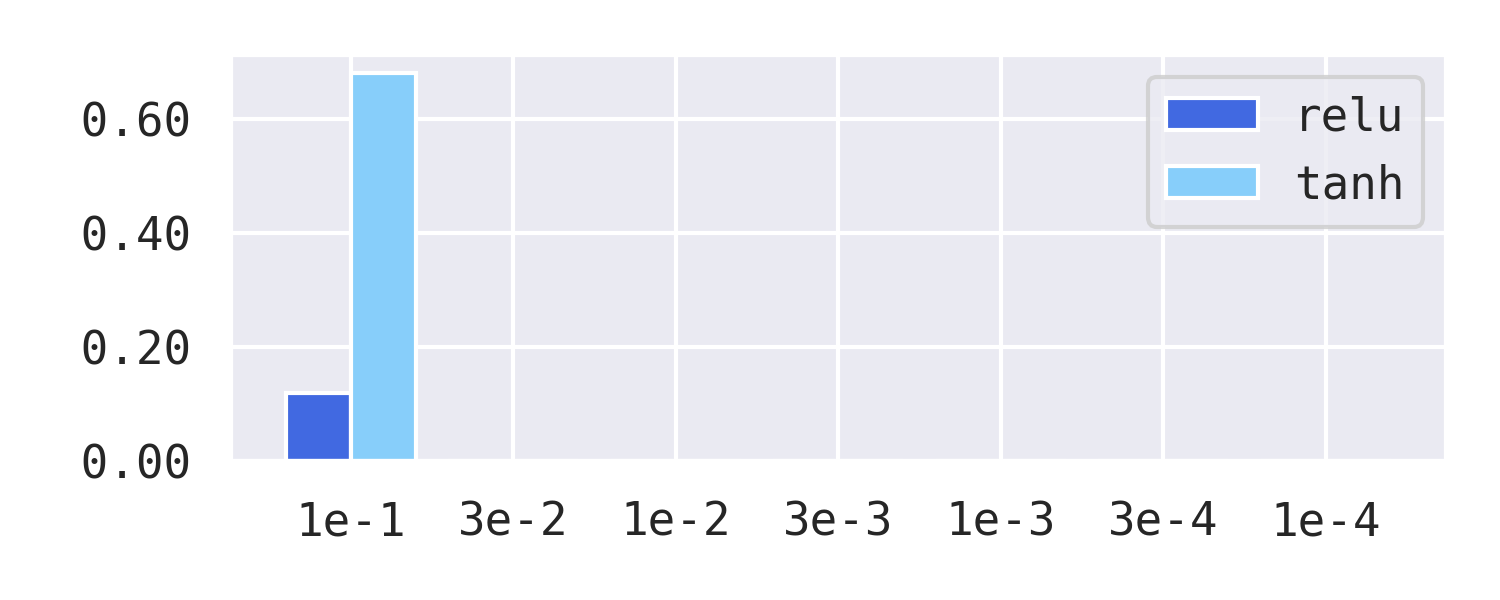}
        \caption{Adadelta}
    \end{subfigure}
    \begin{subfigure}{.49\linewidth}
        \includegraphics[width=\linewidth]{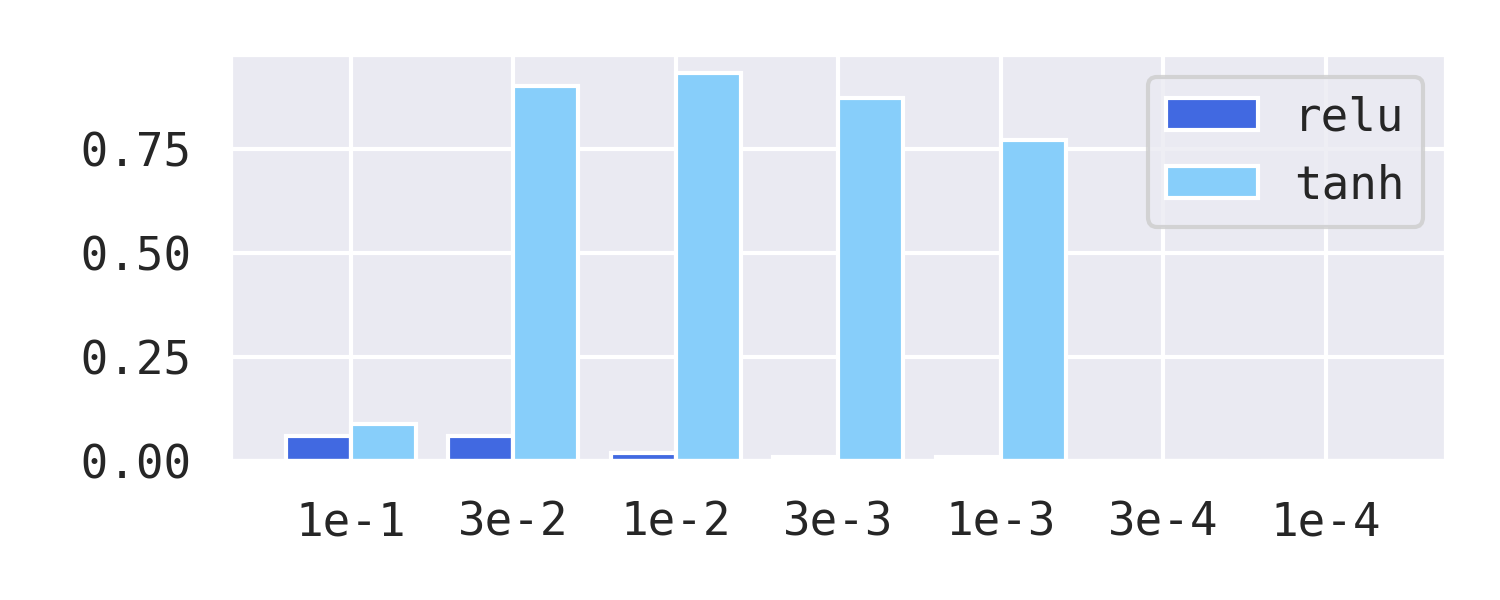}
        \caption{Adafactor}
    \end{subfigure}
    \\
    \begin{subfigure}{.49\linewidth}
        \includegraphics[width=\linewidth]{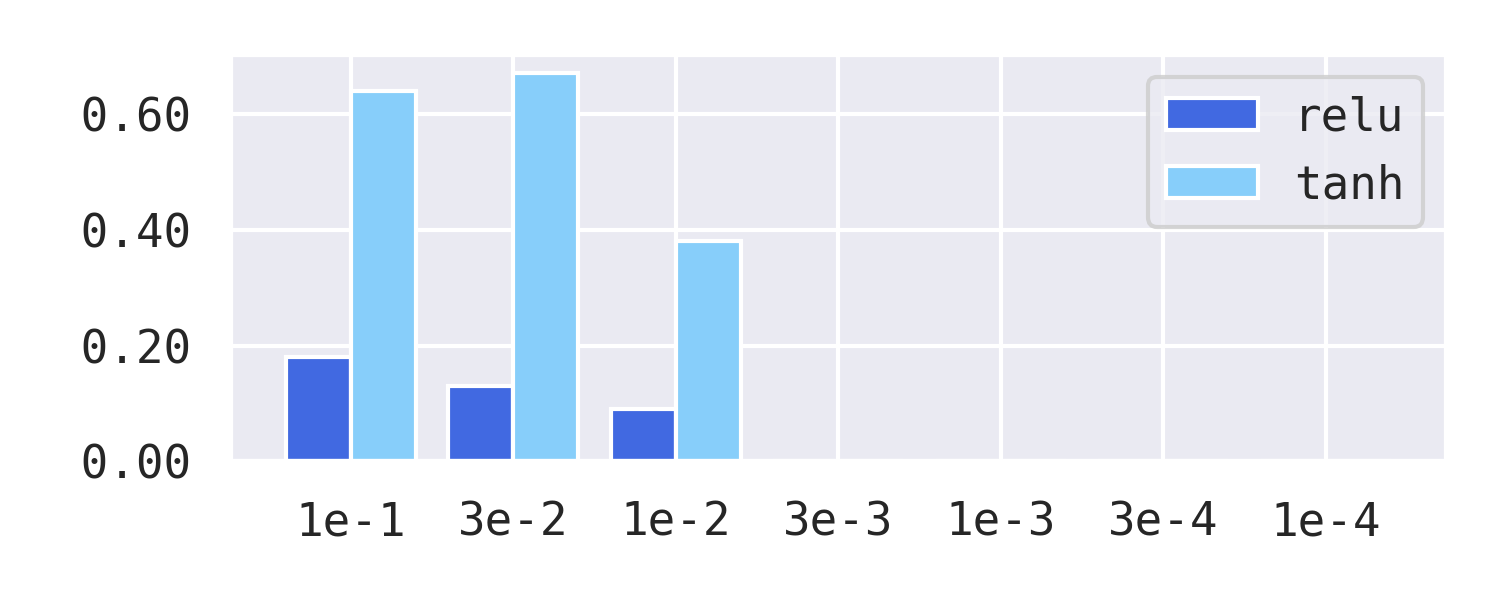}
        \caption{Adagrad}
    \end{subfigure}
    \begin{subfigure}{.49\linewidth}
        \includegraphics[width=\linewidth]{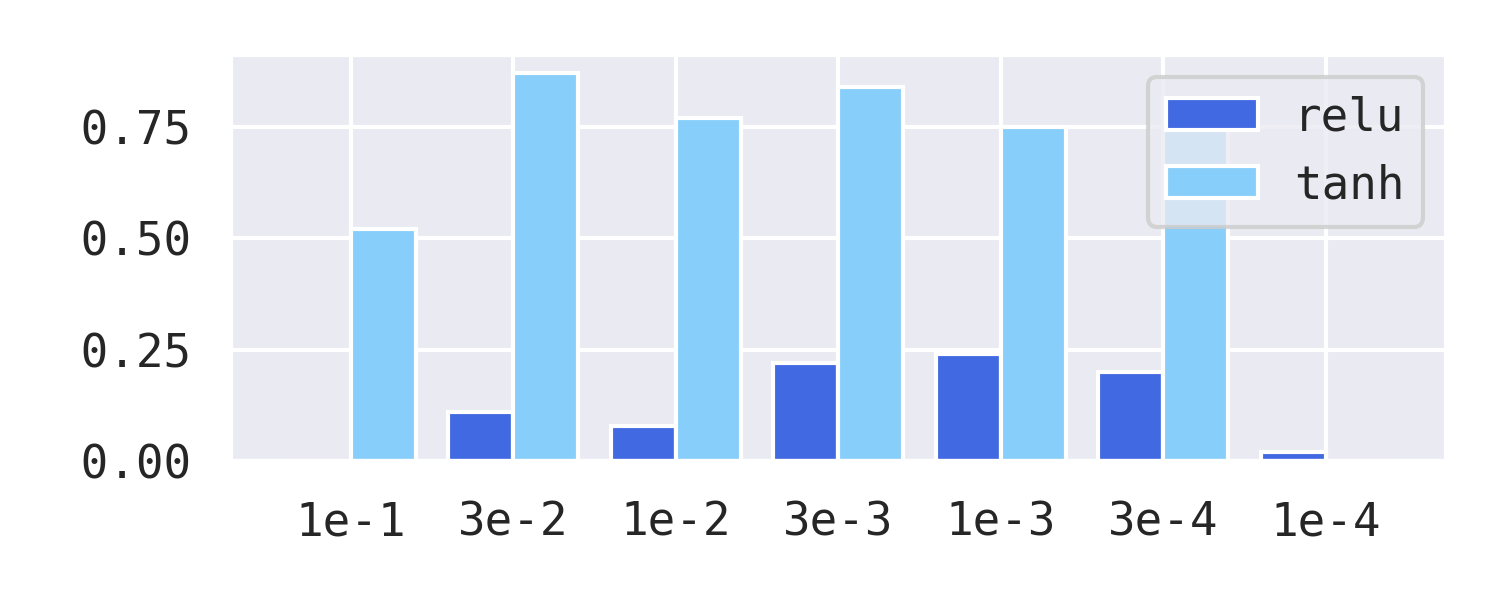}
        \caption{Adam}
    \end{subfigure}
    \\
    \begin{subfigure}{.49\linewidth}
        \includegraphics[width=\linewidth]{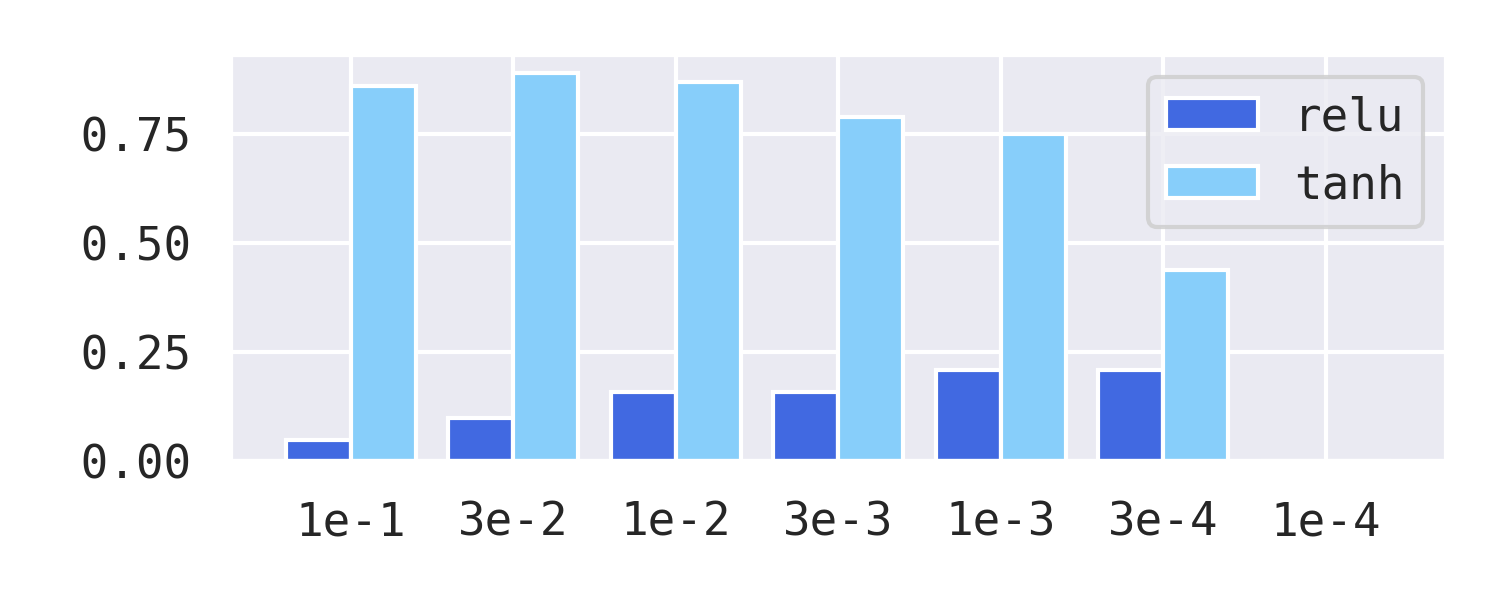}
        \caption{Adamax}
    \end{subfigure}
    \begin{subfigure}{.49\linewidth}
        \includegraphics[width=\linewidth]{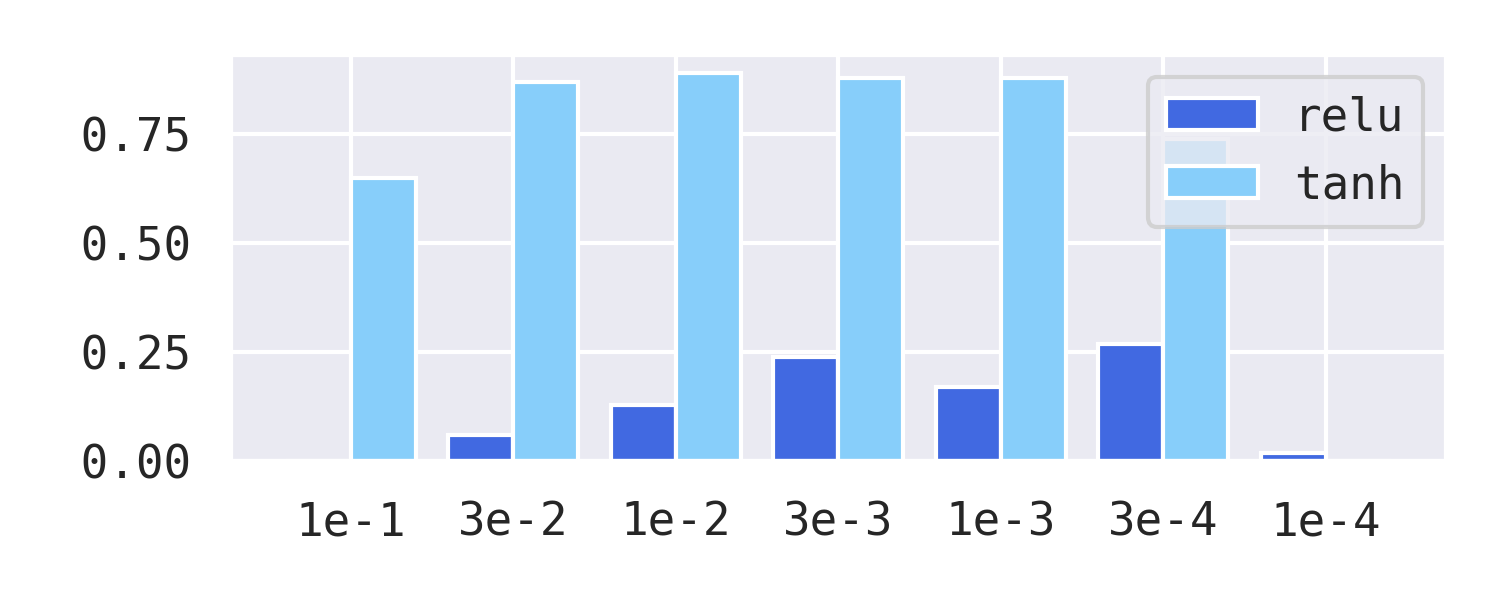}
        \caption{AdamW}
    \end{subfigure}
    \\
    \begin{subfigure}{.49\linewidth}
        \includegraphics[width=\linewidth]{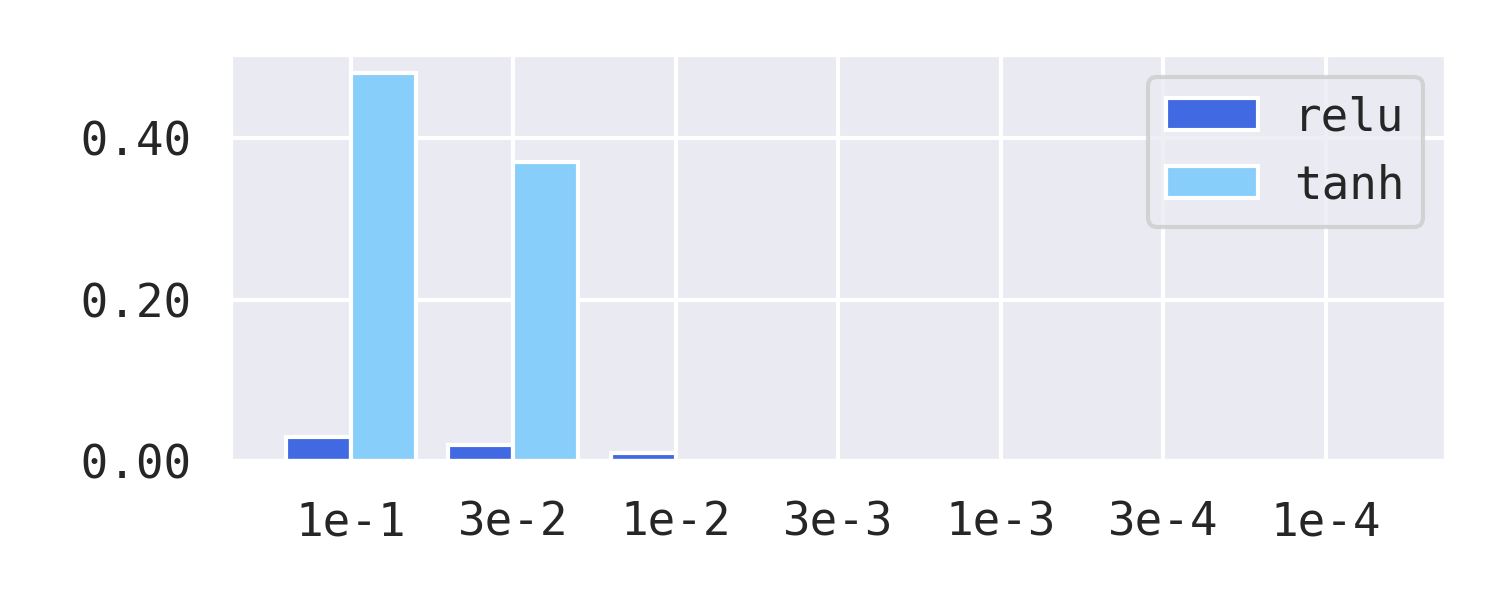}
        \caption{Ftrl}
    \end{subfigure}
    \begin{subfigure}{.49\linewidth}
        \includegraphics[width=\linewidth]{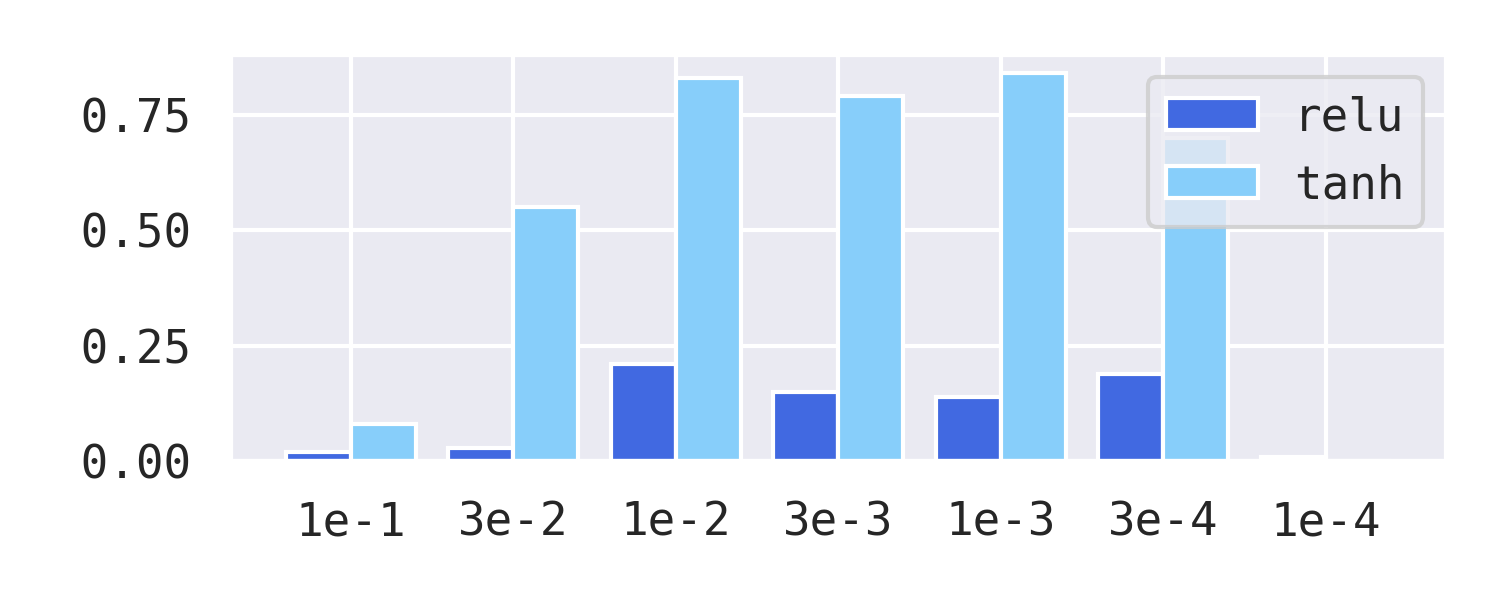}
        \caption{Nadam}
    \end{subfigure}
    \\
    \begin{subfigure}{.49\linewidth}
        \includegraphics[width=\linewidth]{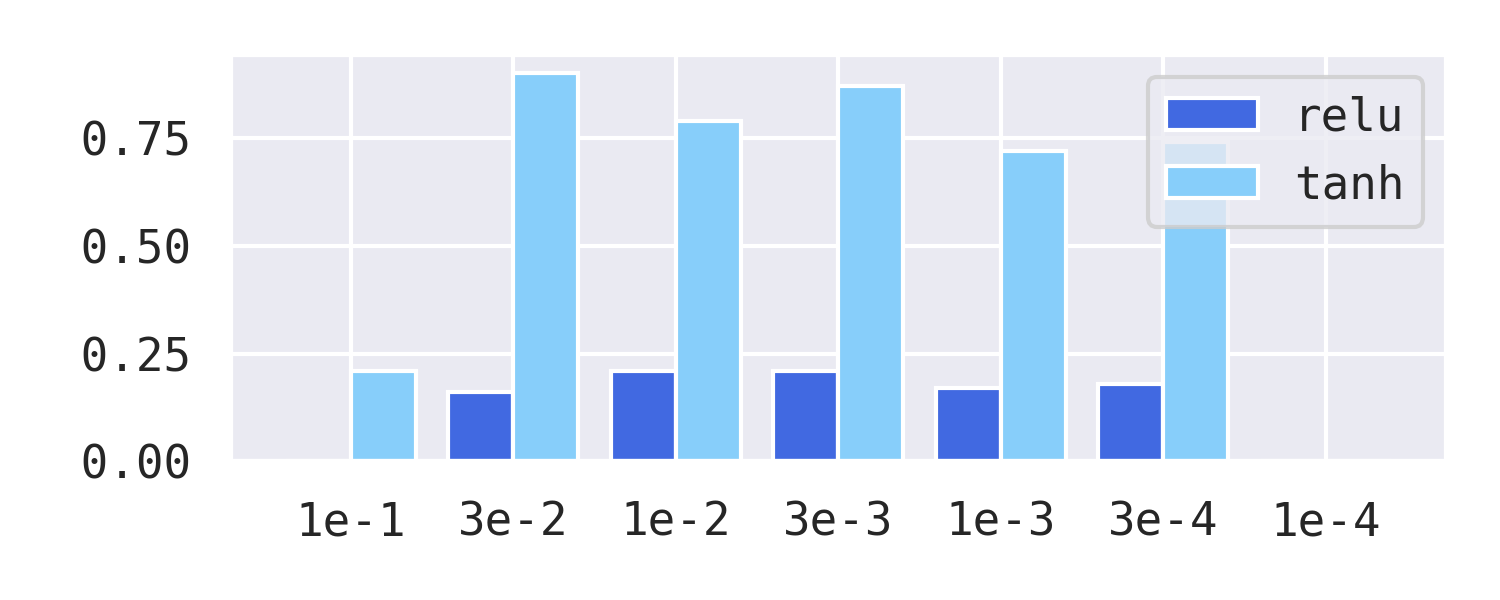}
        \caption{RMSprop}
    \end{subfigure}
    \begin{subfigure}{.49\linewidth}
        \includegraphics[width=\linewidth]{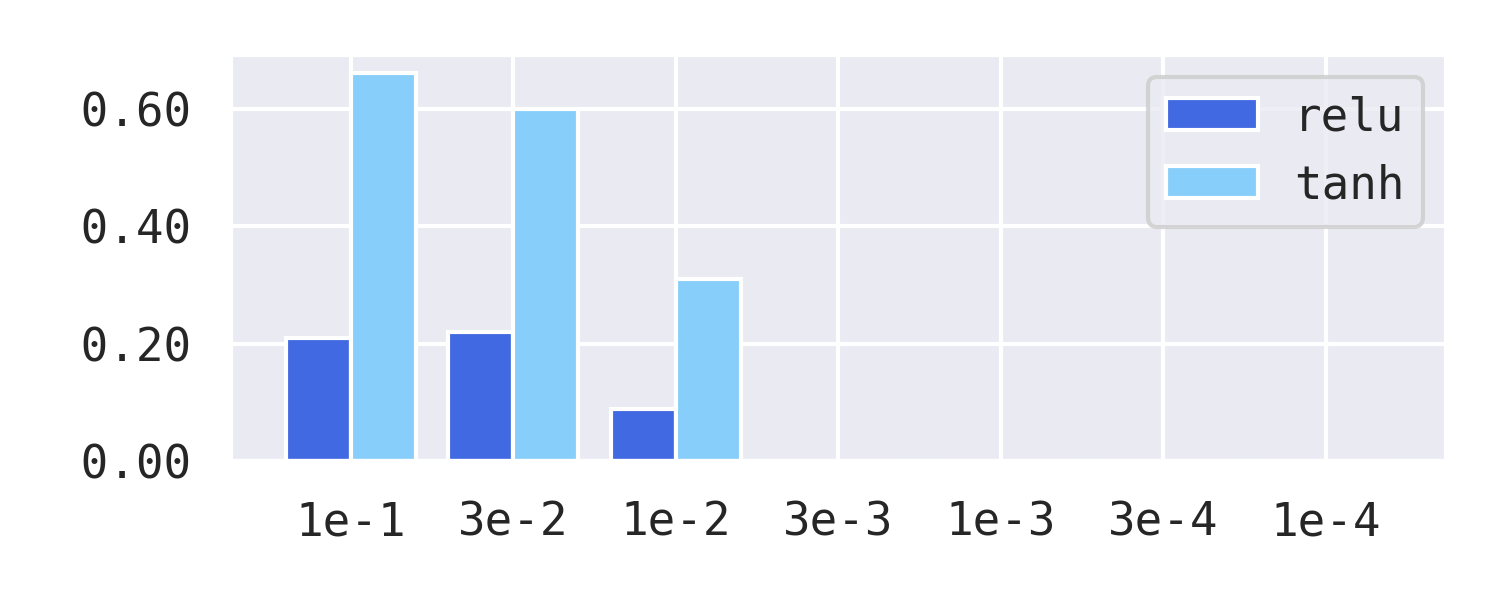}
        \caption{SGD}
    \end{subfigure}
    \caption{Hyperparameter search for random dataset on 1-step Game of Life.}
    \label{fig:search_1_random}
\end{figure*}

\begin{figure*}[ht!]
    \centering
    \begin{subfigure}{.49\linewidth}
        \includegraphics[width=\linewidth]{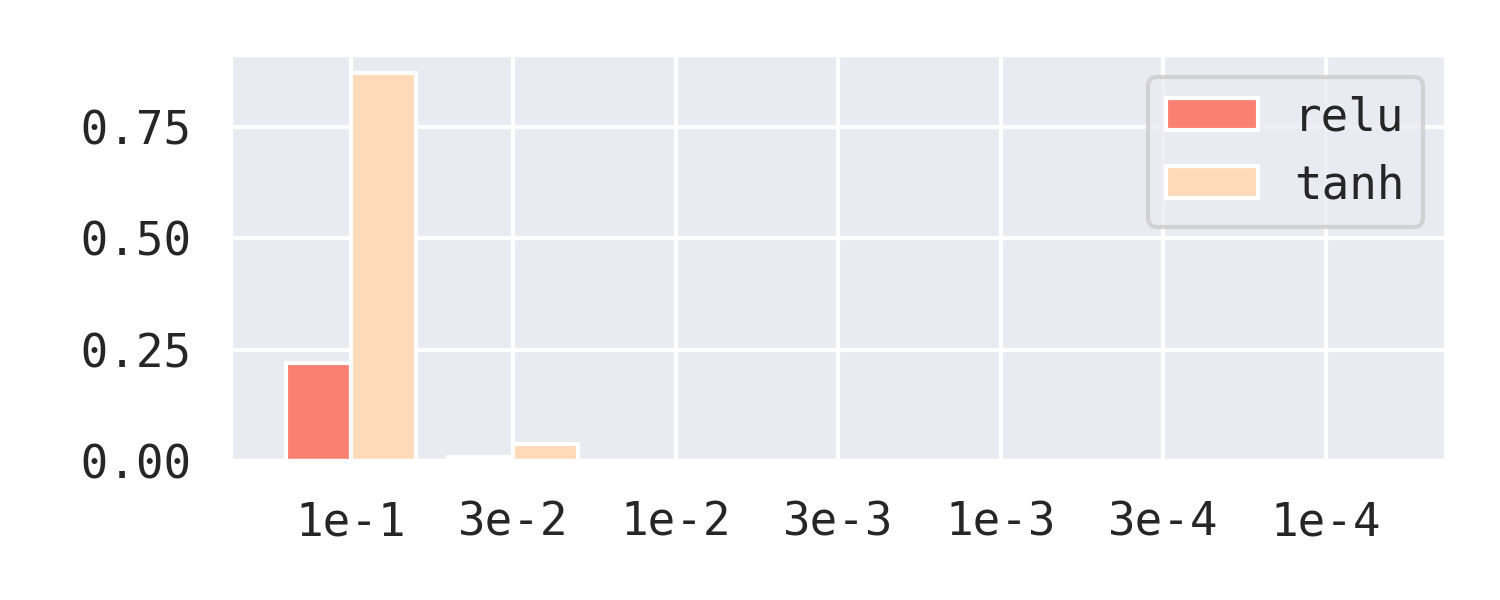}
        \caption{Adadelta}
    \end{subfigure}
    \begin{subfigure}{.49\linewidth}
        \includegraphics[width=\linewidth]{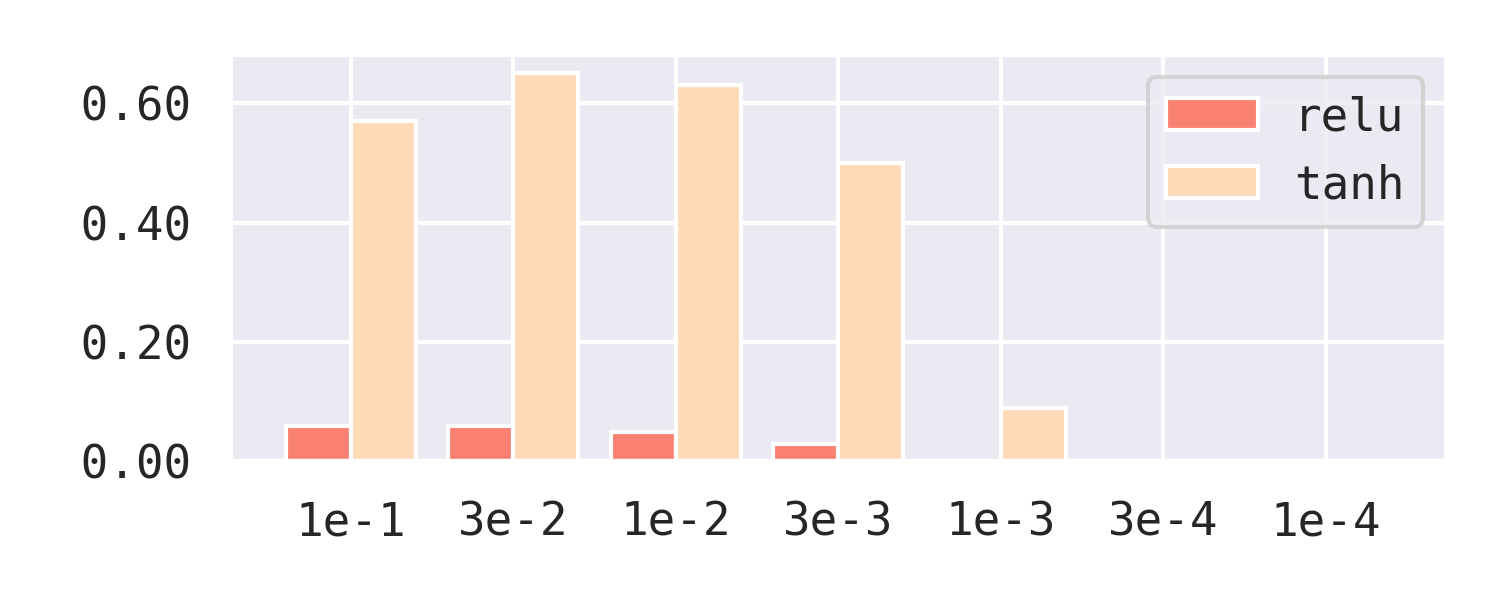}
        \caption{Adafactor}
    \end{subfigure}
    \\
    \begin{subfigure}{.49\linewidth}
        \includegraphics[width=\linewidth]{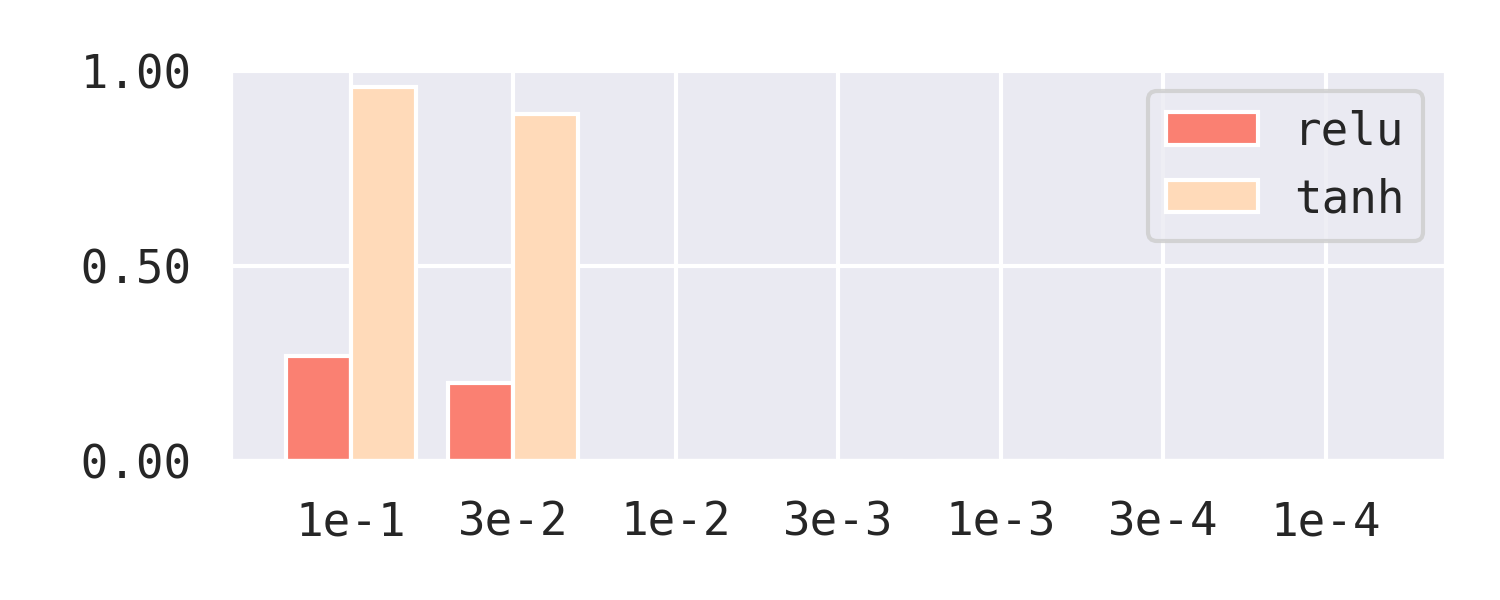}
        \caption{Adagrad}
    \end{subfigure}
    \begin{subfigure}{.49\linewidth}
        \includegraphics[width=\linewidth]{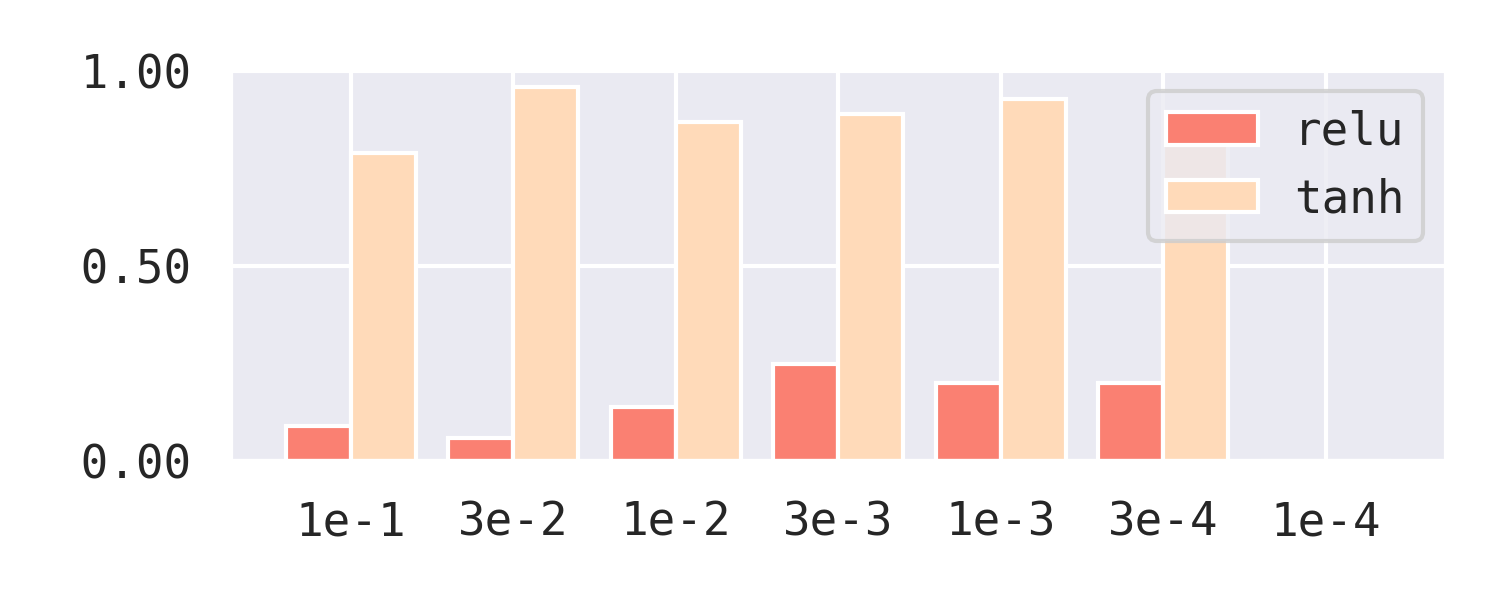}
        \caption{Adam}
    \end{subfigure}
    \\
    \begin{subfigure}{.49\linewidth}
        \includegraphics[width=\linewidth]{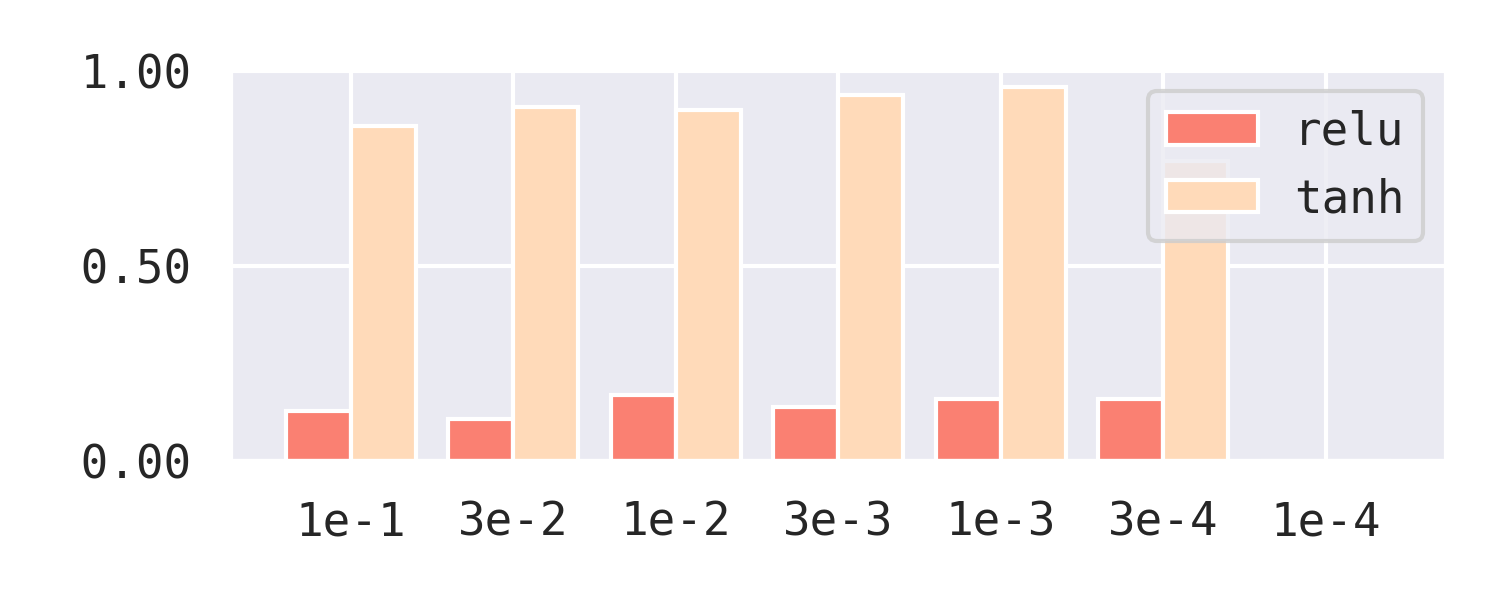}
        \caption{Adamax}
    \end{subfigure}
    \begin{subfigure}{.49\linewidth}
        \includegraphics[width=\linewidth]{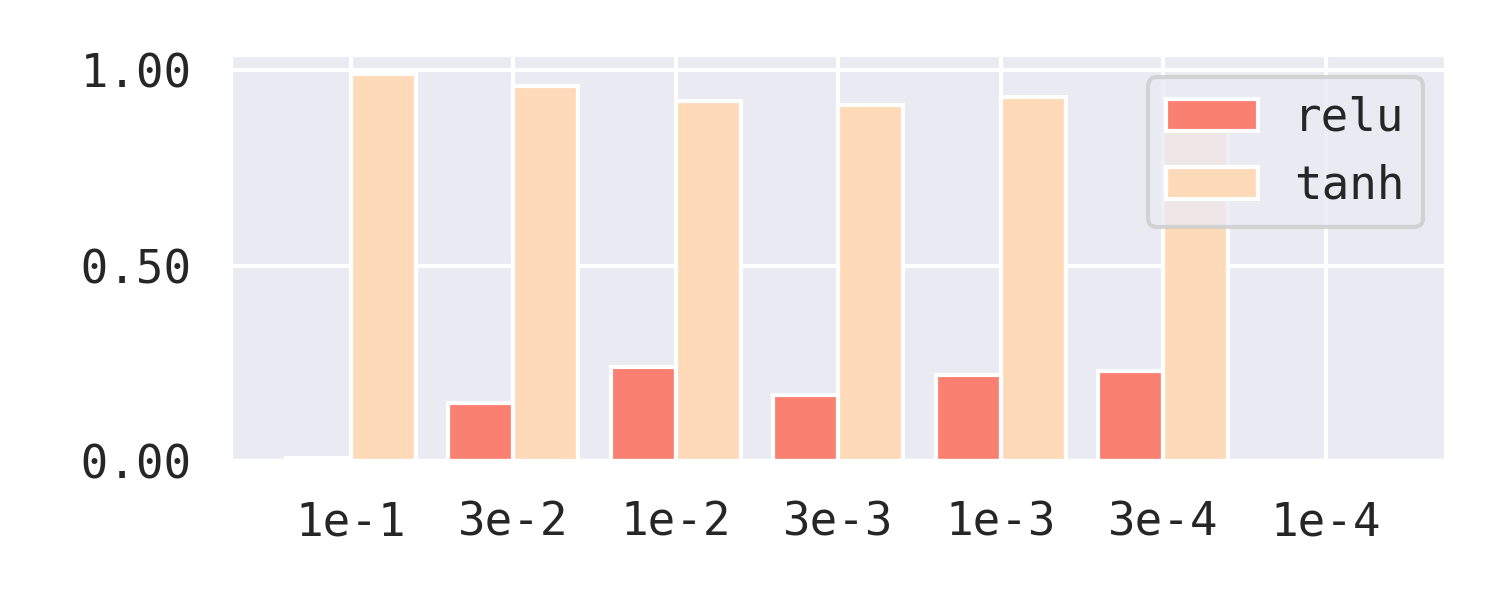}
        \caption{AdamW}
    \end{subfigure}
    \\
    \begin{subfigure}{.49\linewidth}
        \includegraphics[width=\linewidth]{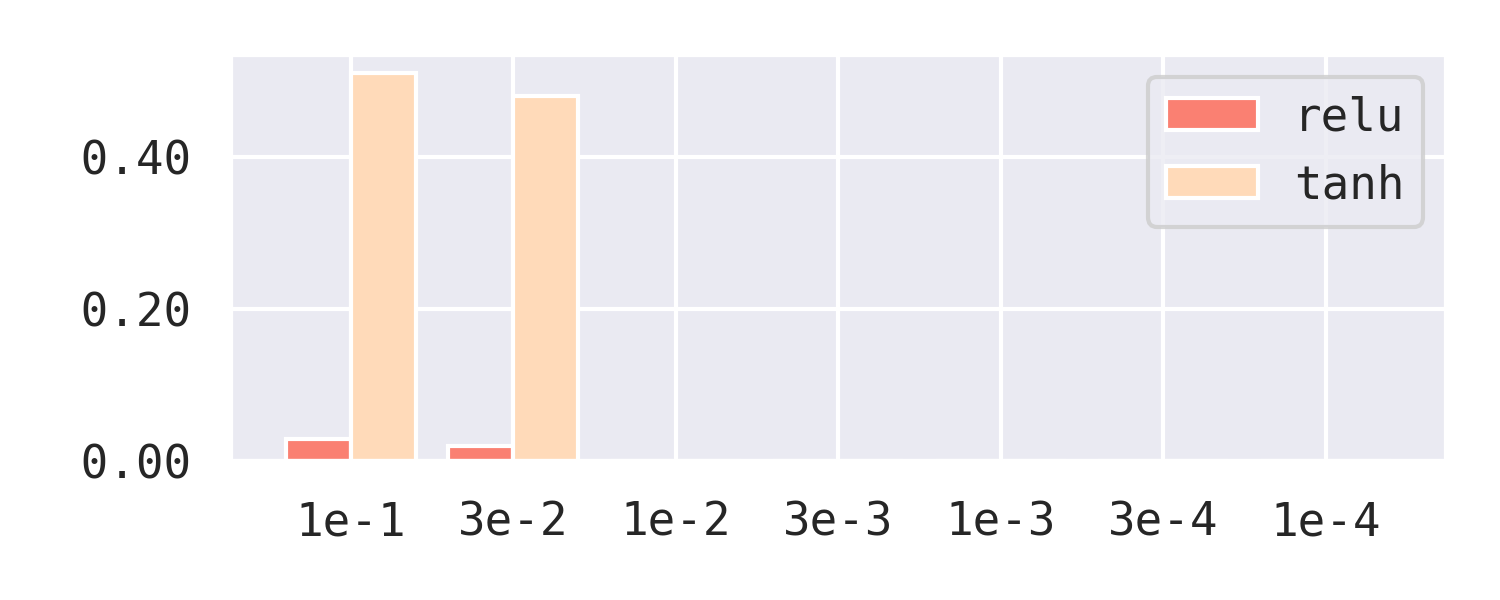}
        \caption{Ftrl}
    \end{subfigure}
    \begin{subfigure}{.49\linewidth}
        \includegraphics[width=\linewidth]{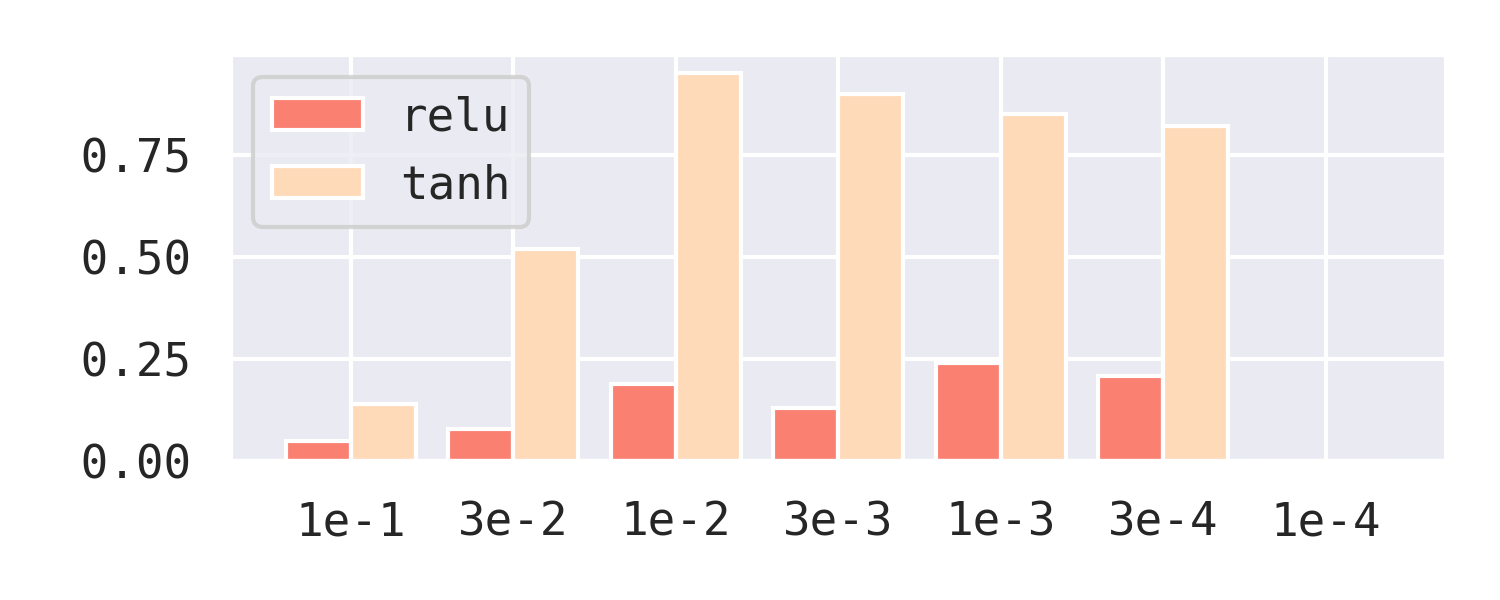}
        \caption{Nadam}
    \end{subfigure}
    \\
    \begin{subfigure}{.49\linewidth}
        \includegraphics[width=\linewidth]{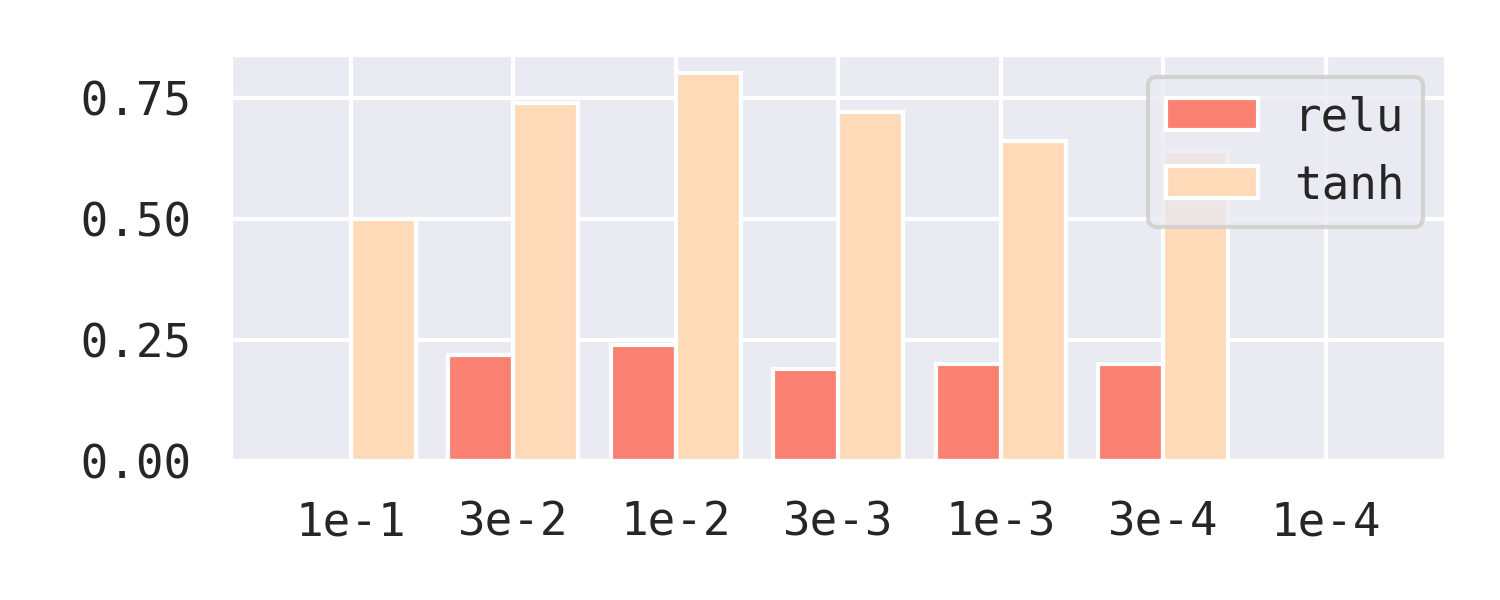}
        \caption{RMSprop}
    \end{subfigure}
    \begin{subfigure}{.49\linewidth}
        \includegraphics[width=\linewidth]{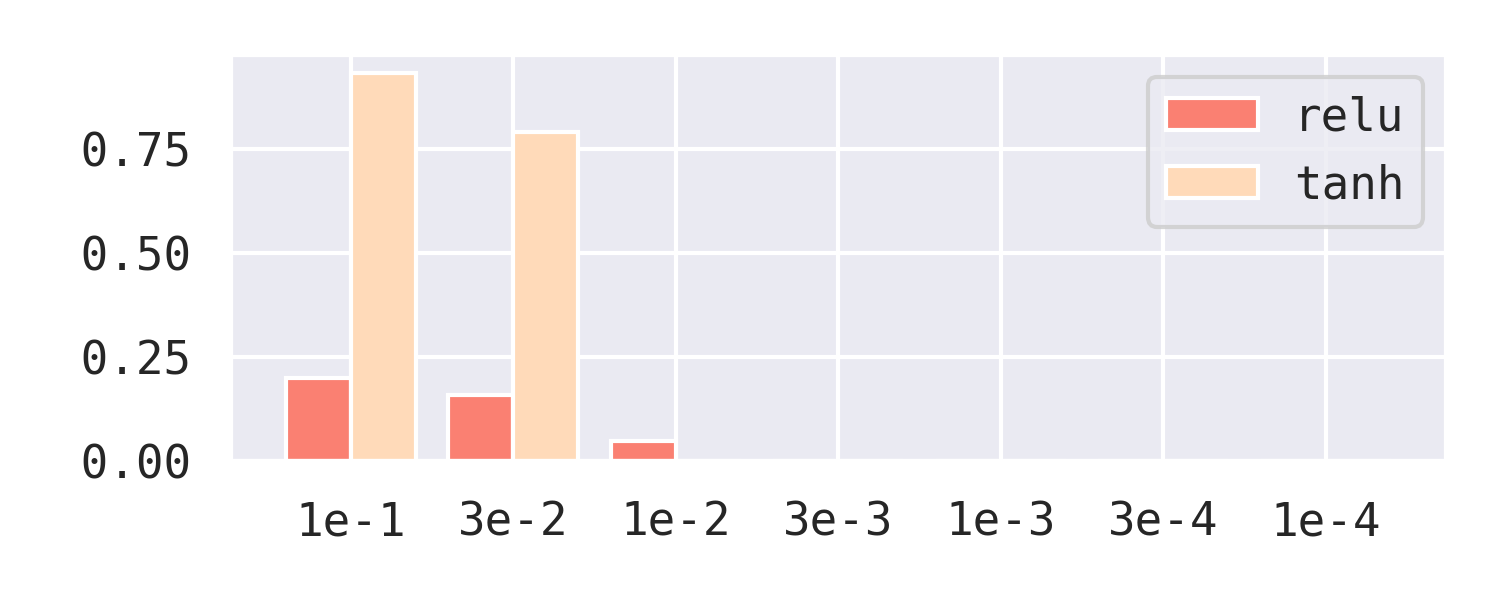}
        \caption{SGD}
    \end{subfigure}
    \caption{Hyperparameter search for fixed dataset on 1-step Game of Life.}
    \label{fig:search_1_fixed}
\end{figure*}

\begin{figure*}[ht!]
    \centering
    \begin{subfigure}{.49\linewidth}
        \includegraphics[width=\linewidth]{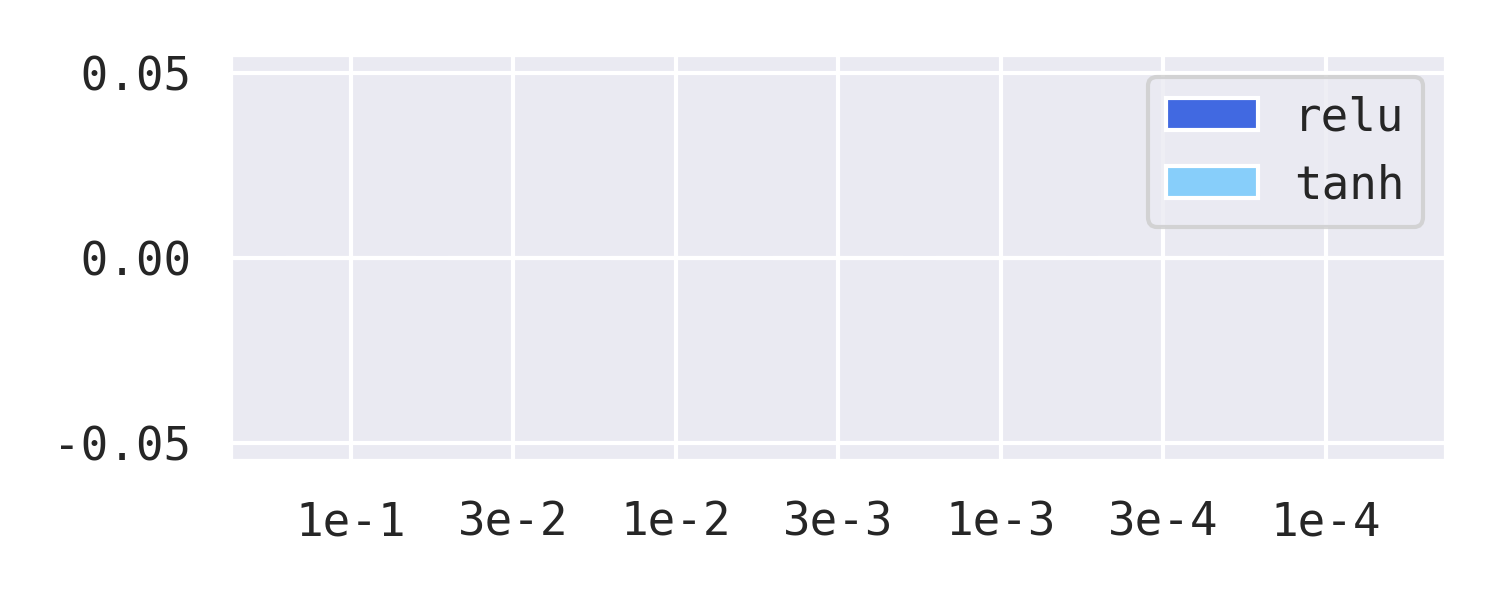}
        \caption{Adadelta}
    \end{subfigure}
    \begin{subfigure}{.49\linewidth}
        \includegraphics[width=\linewidth]{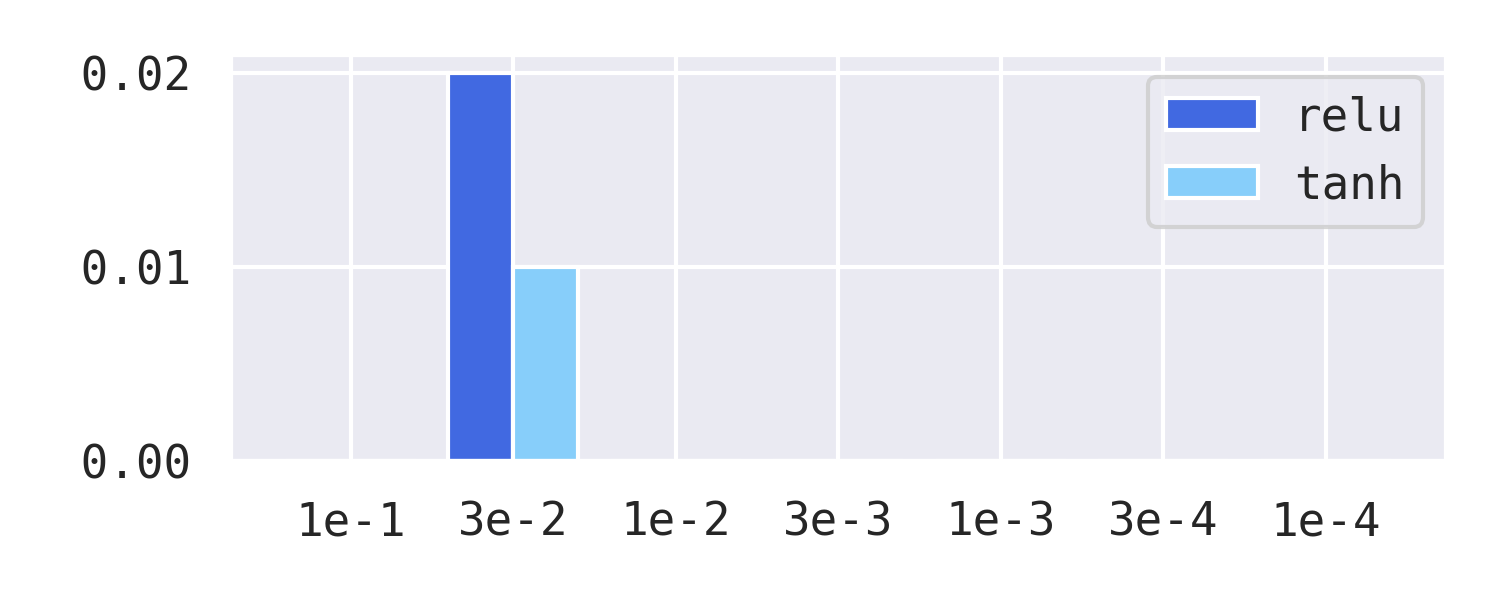}
        \caption{Adafactor}
    \end{subfigure}
    \\
    \begin{subfigure}{.49\linewidth}
        \includegraphics[width=\linewidth]{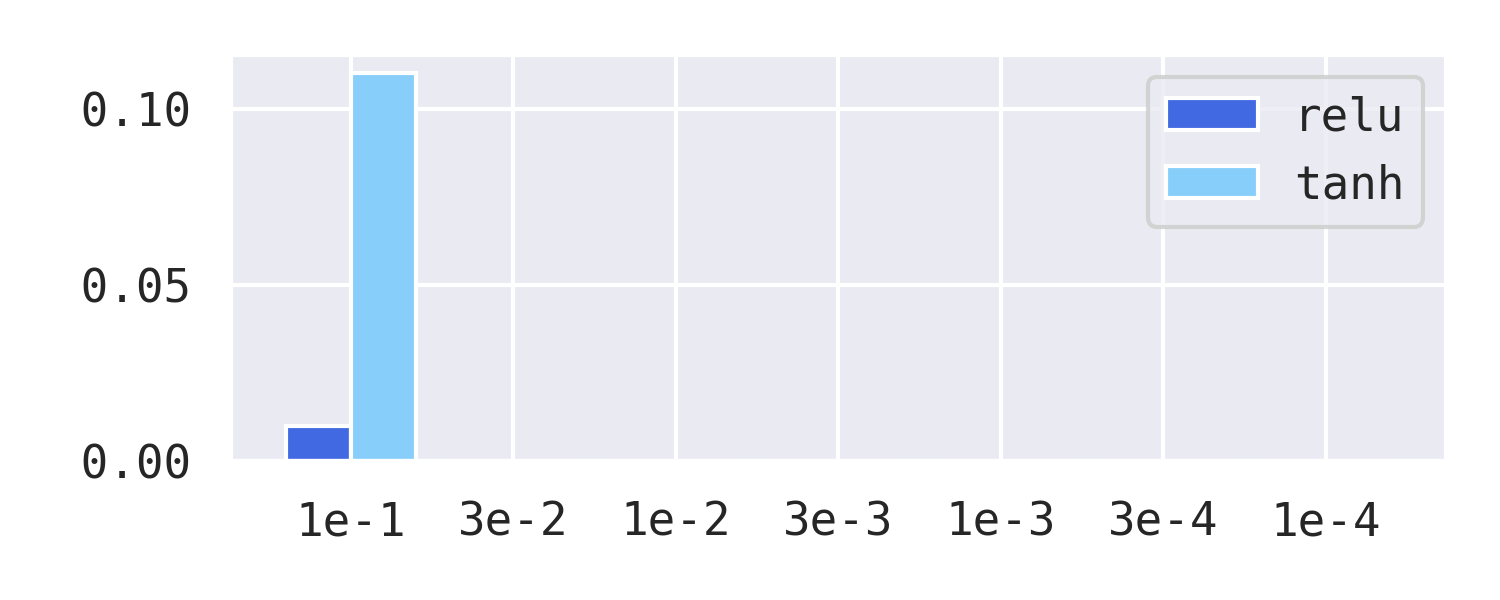}
        \caption{Adagrad}
    \end{subfigure}
    \begin{subfigure}{.49\linewidth}
        \includegraphics[width=\linewidth]{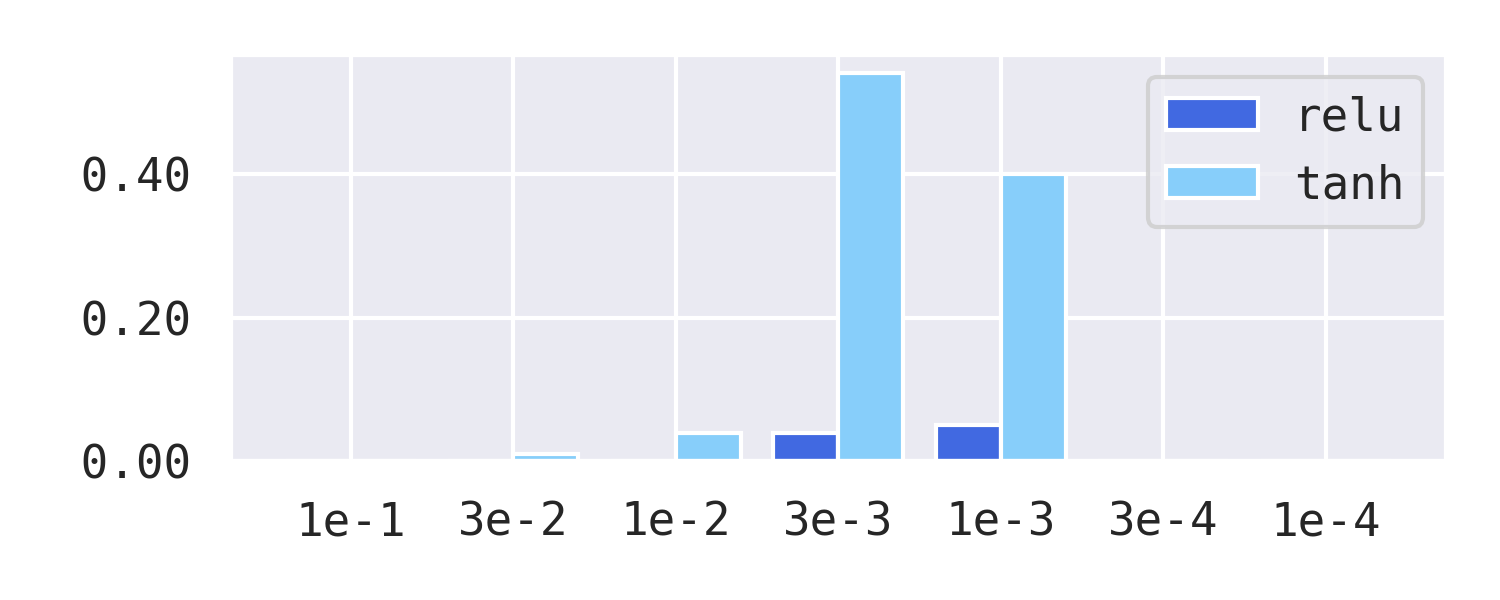}
        \caption{Adam}
    \end{subfigure}
    \\
    \begin{subfigure}{.49\linewidth}
        \includegraphics[width=\linewidth]{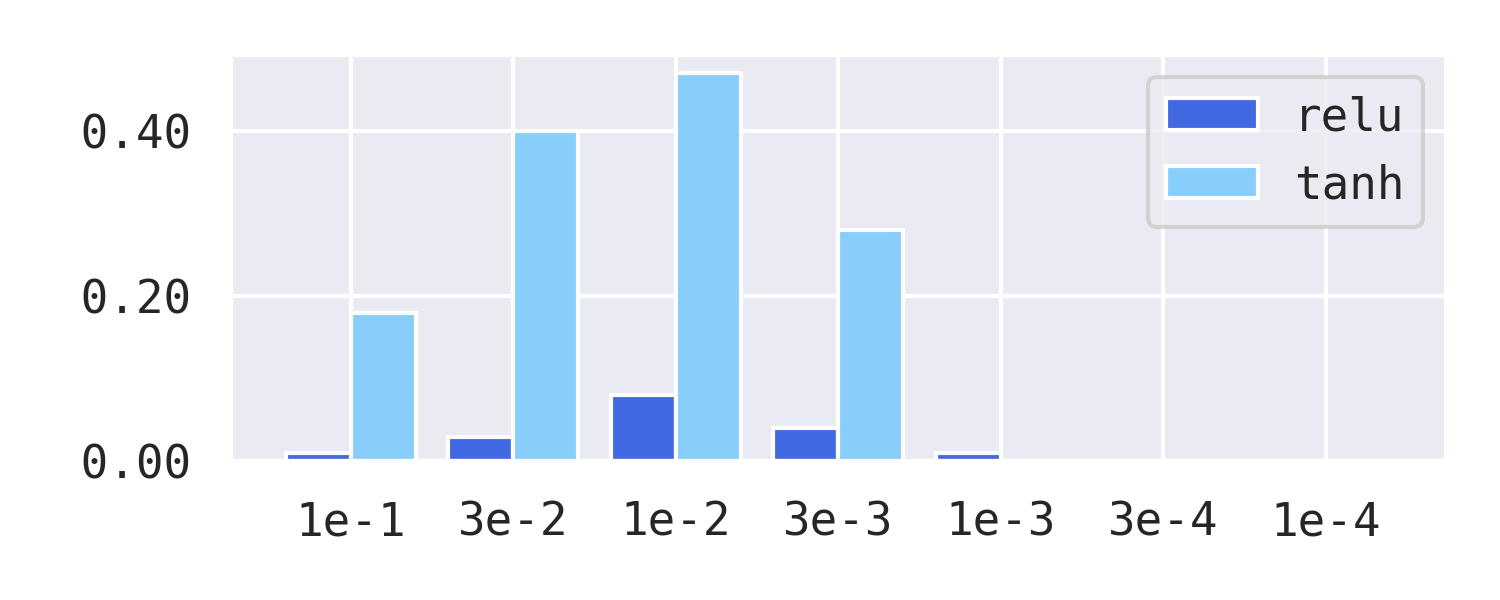}
        \caption{Adamax}
    \end{subfigure}
    \begin{subfigure}{.49\linewidth}
        \includegraphics[width=\linewidth]{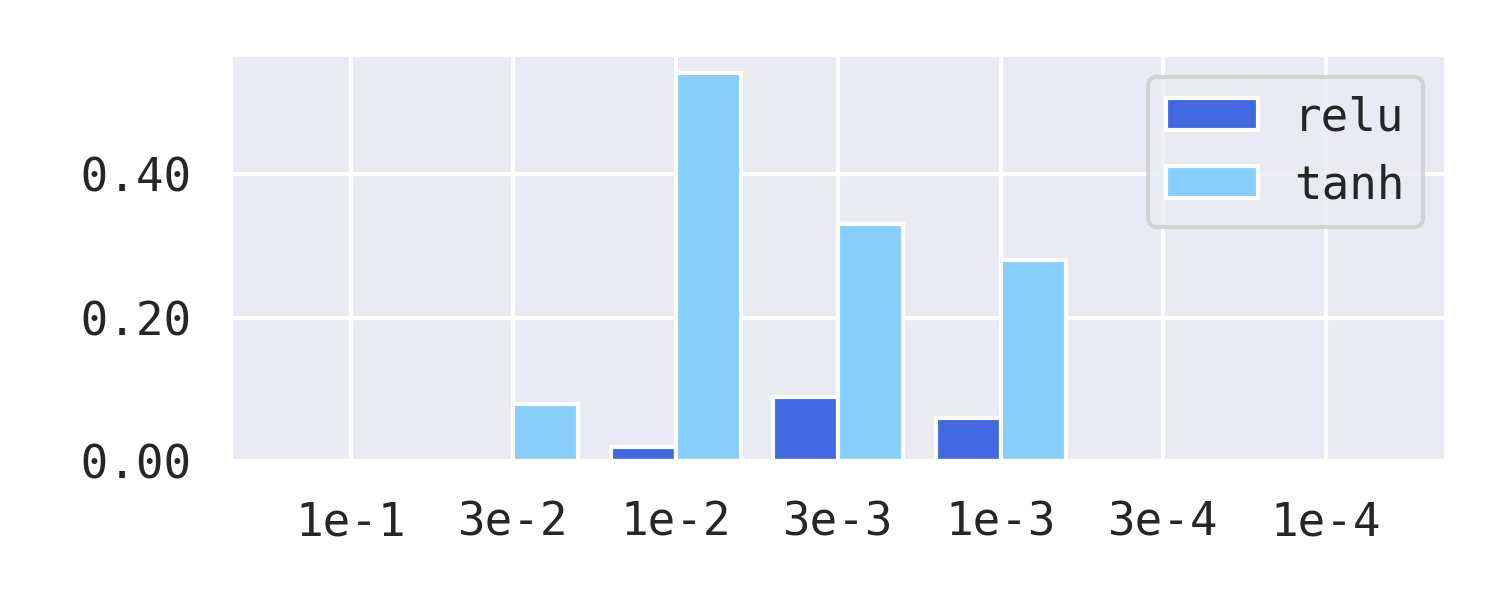}
        \caption{AdamW}
    \end{subfigure}
    \\
    \begin{subfigure}{.49\linewidth}
        \includegraphics[width=\linewidth]{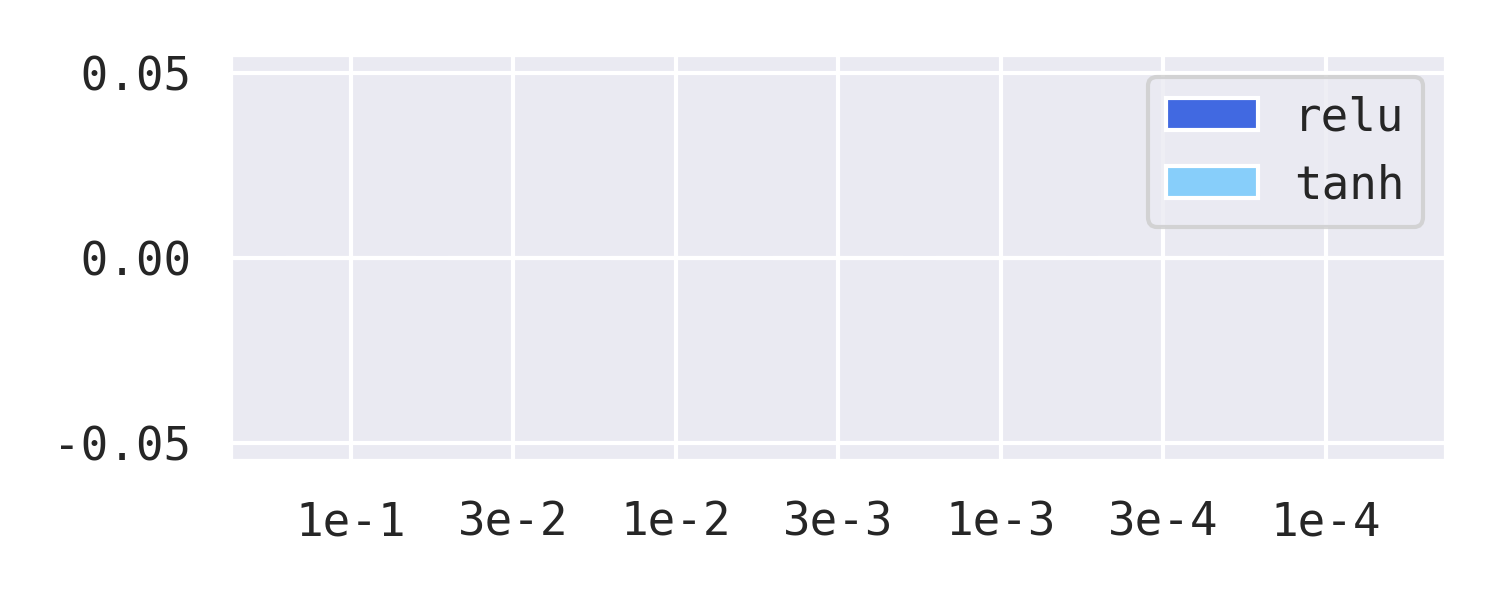}
        \caption{Ftrl}
    \end{subfigure}
    \begin{subfigure}{.49\linewidth}
        \includegraphics[width=\linewidth]{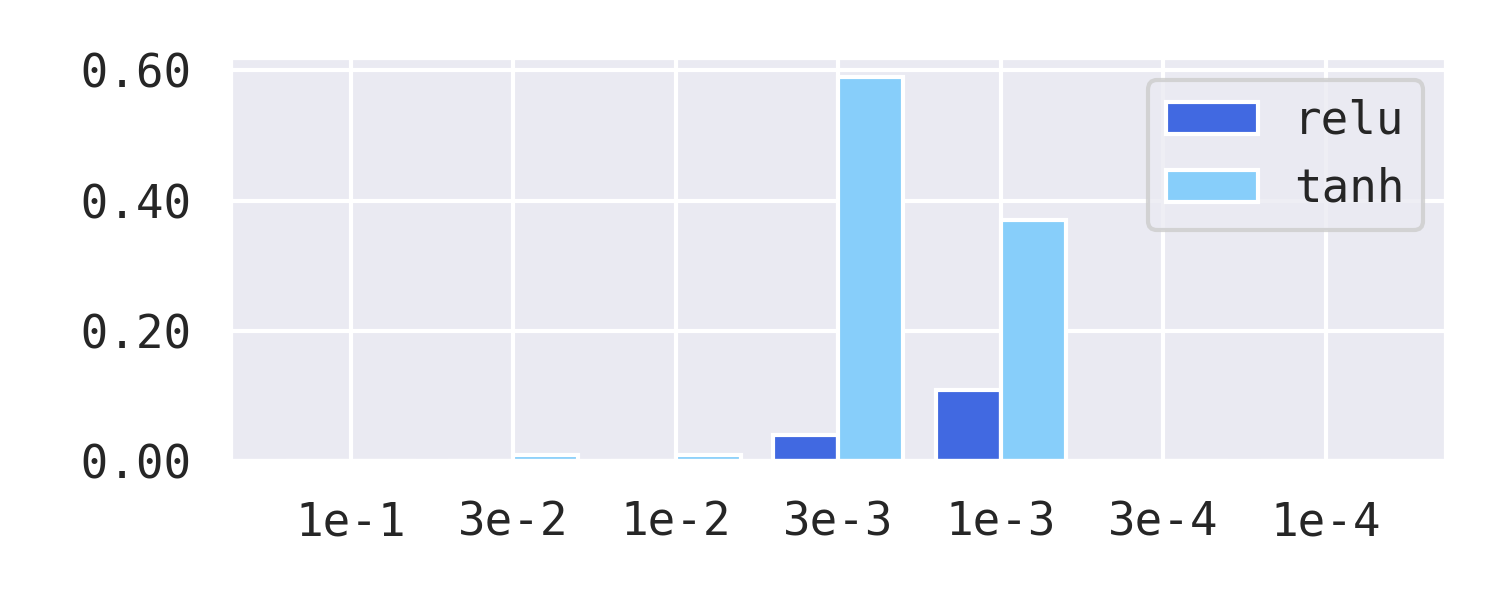}
        \caption{Nadam}
    \end{subfigure}
    \\
    \begin{subfigure}{.49\linewidth}
        \includegraphics[width=\linewidth]{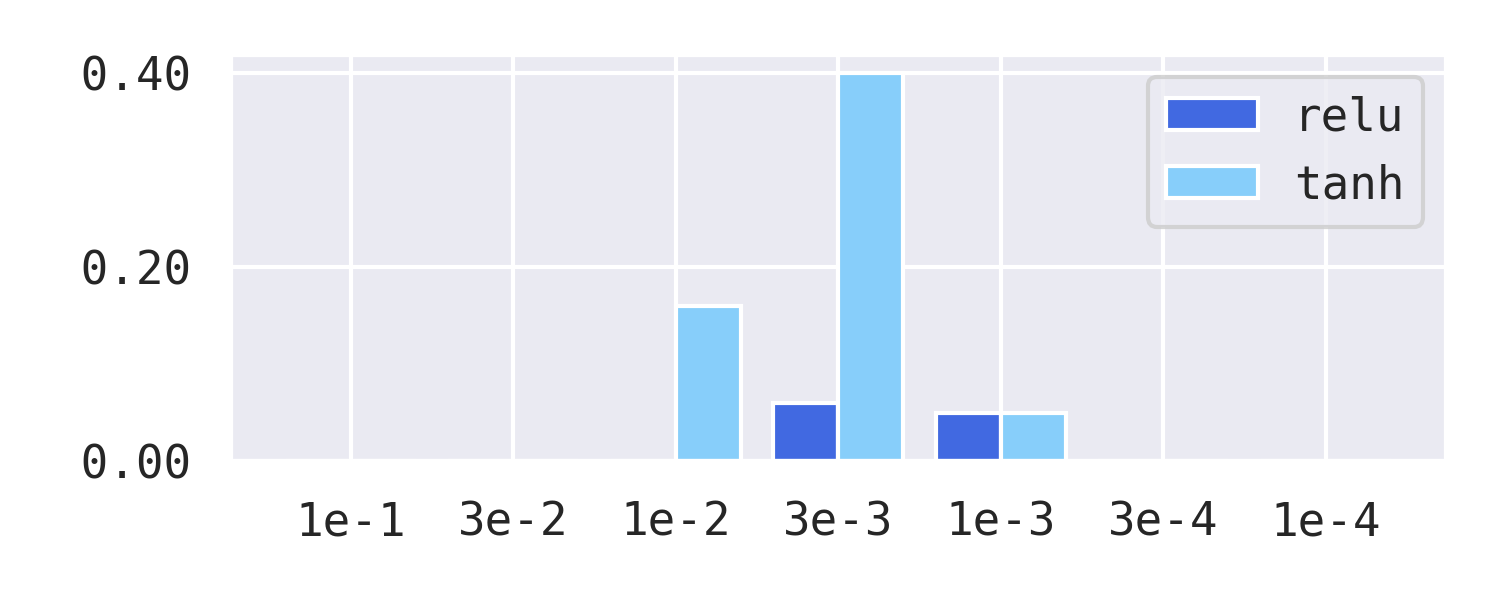}
        \caption{RMSprop}
    \end{subfigure}
    \begin{subfigure}{.49\linewidth}
        \includegraphics[width=\linewidth]{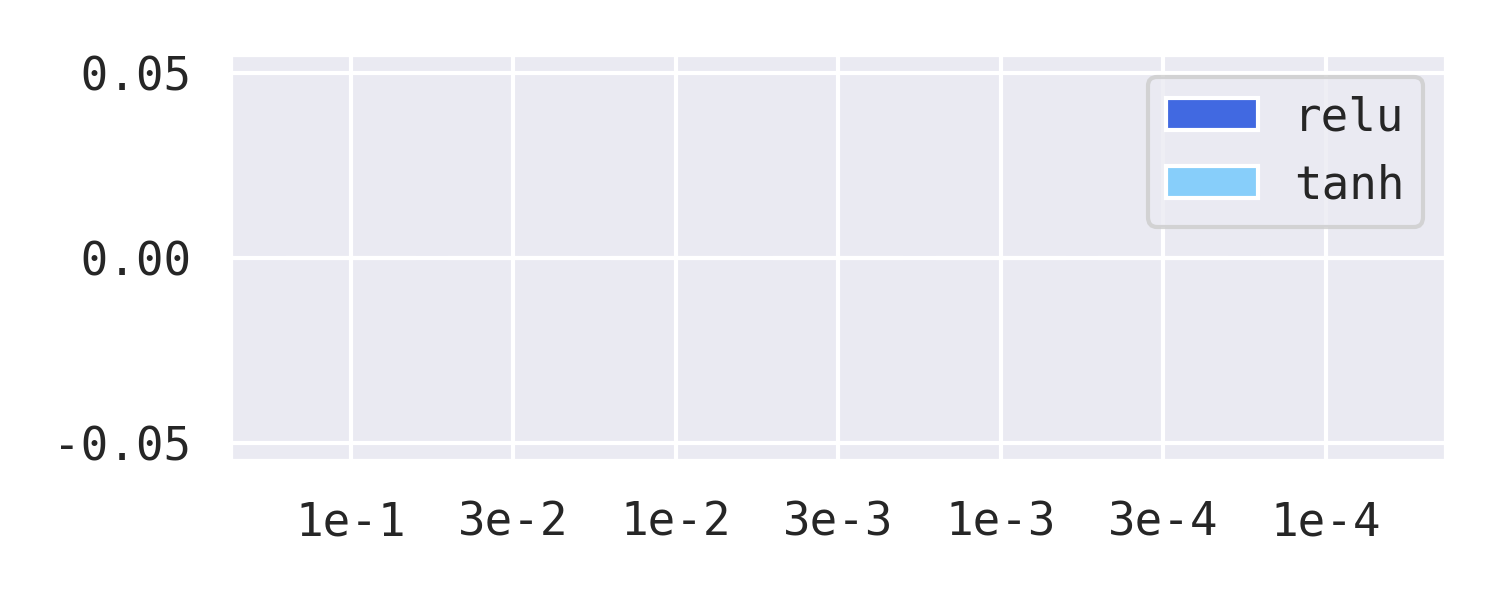}
        \caption{SGD}
    \end{subfigure}
    \caption{Hyperparameter search for random dataset on 2-step Game of Life with recursive network.}
    \label{fig:search_2_rec_random}
\end{figure*}

\begin{figure*}[ht!]
    \centering
    \begin{subfigure}{.49\linewidth}
        \includegraphics[width=\linewidth]{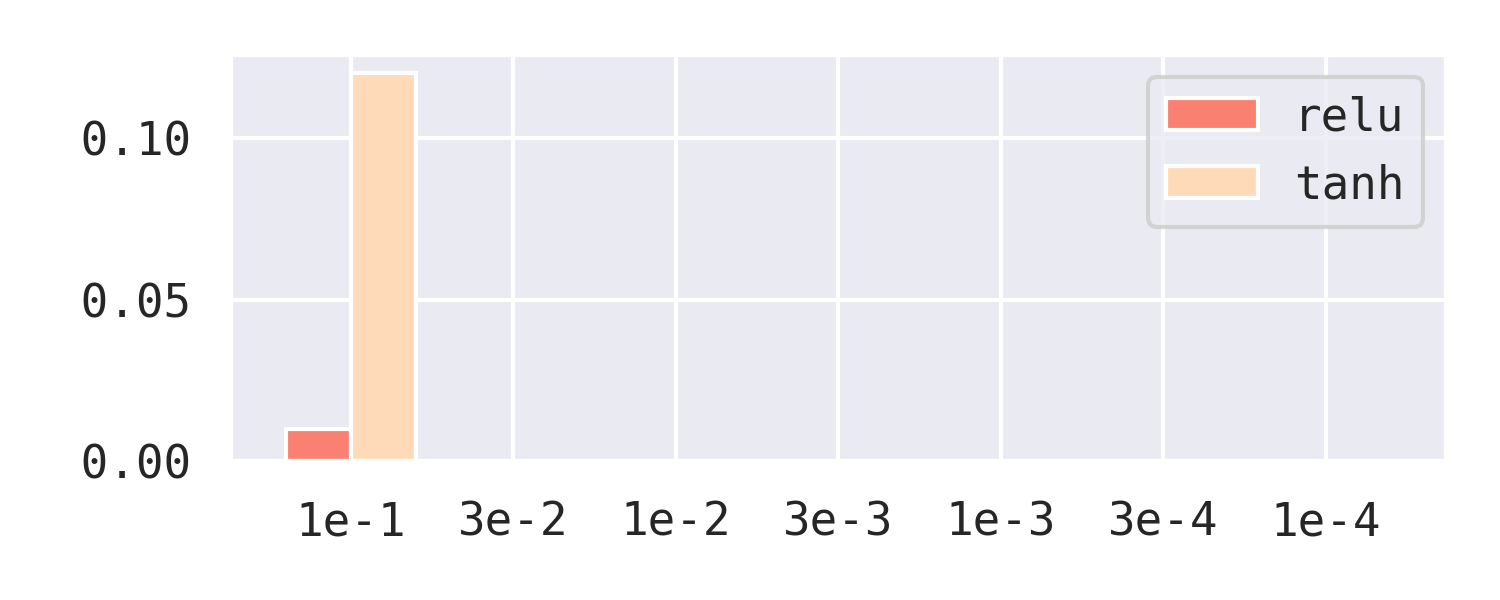}
        \caption{Adadelta}
    \end{subfigure}
    \begin{subfigure}{.49\linewidth}
        \includegraphics[width=\linewidth]{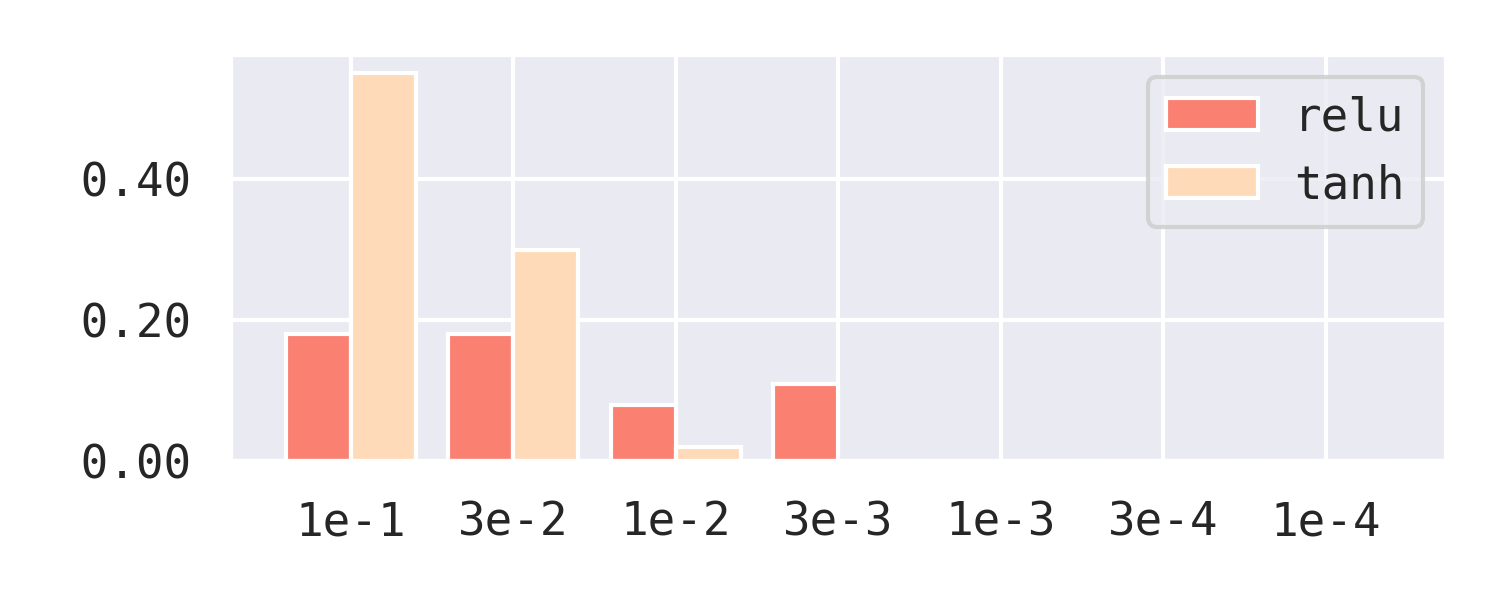}
        \caption{Adafactor}
    \end{subfigure}
    \\
    \begin{subfigure}{.49\linewidth}
        \includegraphics[width=\linewidth]{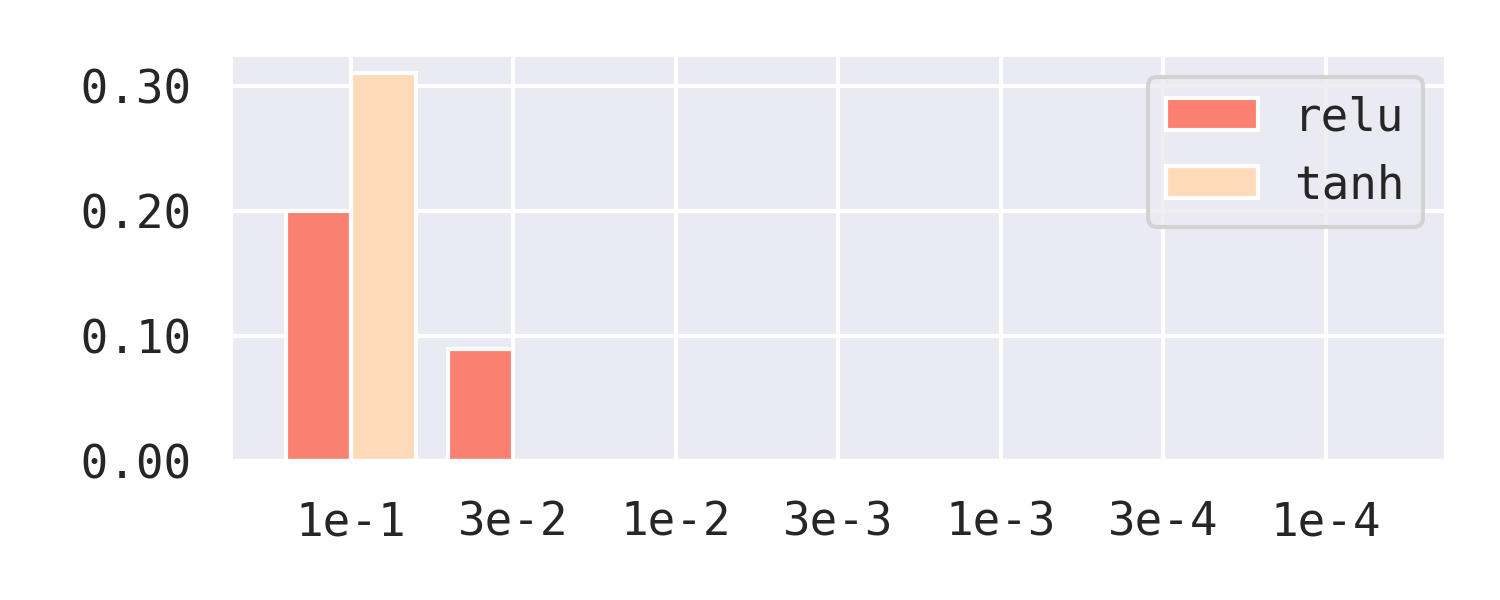}
        \caption{Adagrad}
    \end{subfigure}
    \begin{subfigure}{.49\linewidth}
        \includegraphics[width=\linewidth]{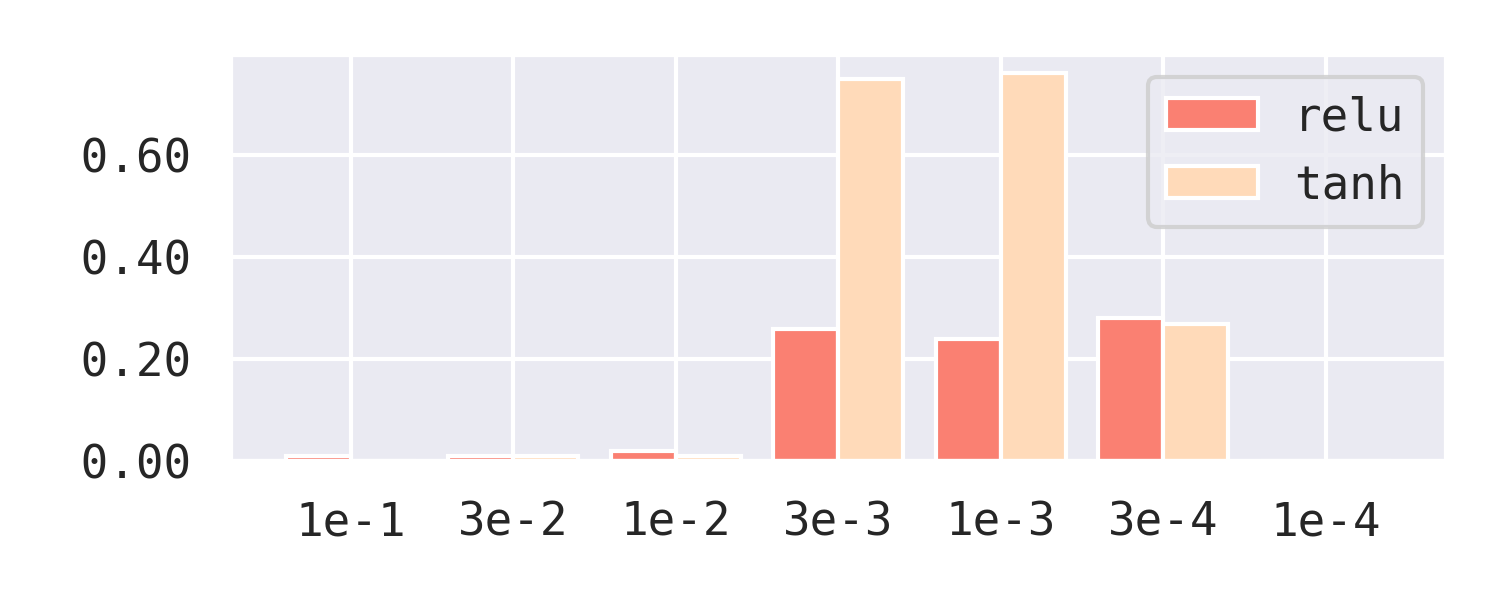}
        \caption{Adam}
    \end{subfigure}
    \\
    \begin{subfigure}{.49\linewidth}
        \includegraphics[width=\linewidth]{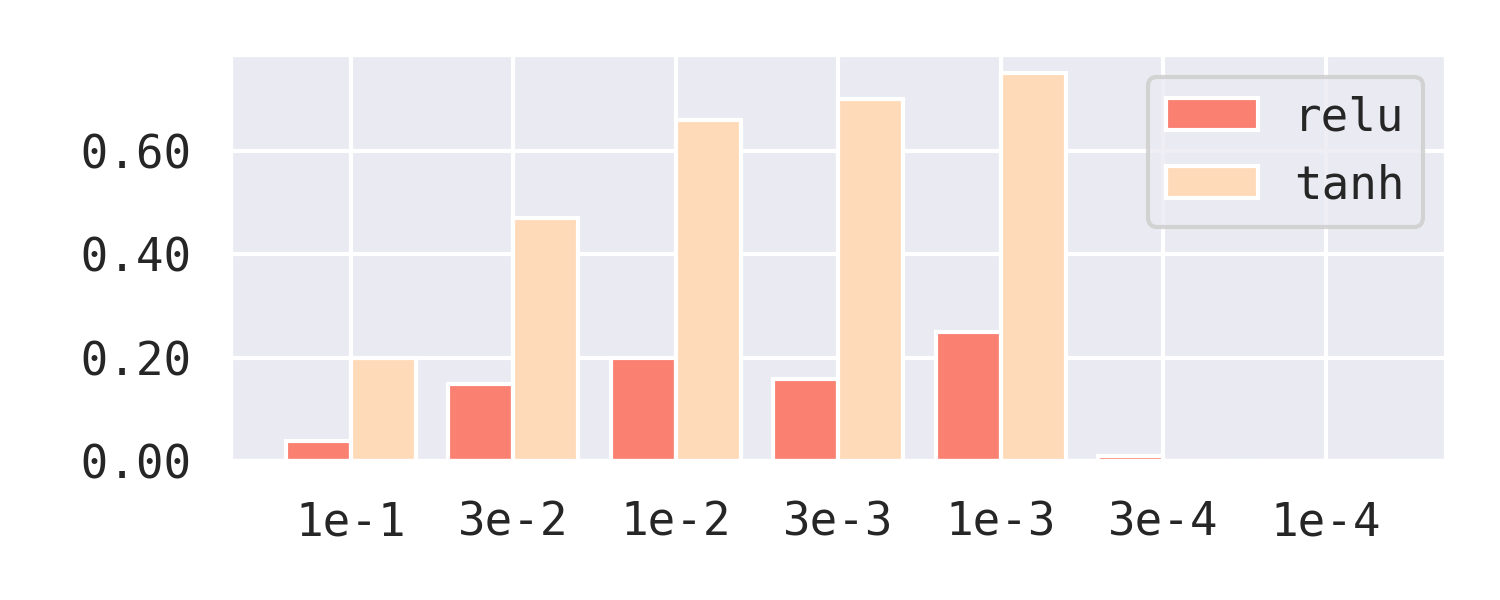}
        \caption{Adamax}
    \end{subfigure}
    \begin{subfigure}{.49\linewidth}
        \includegraphics[width=\linewidth]{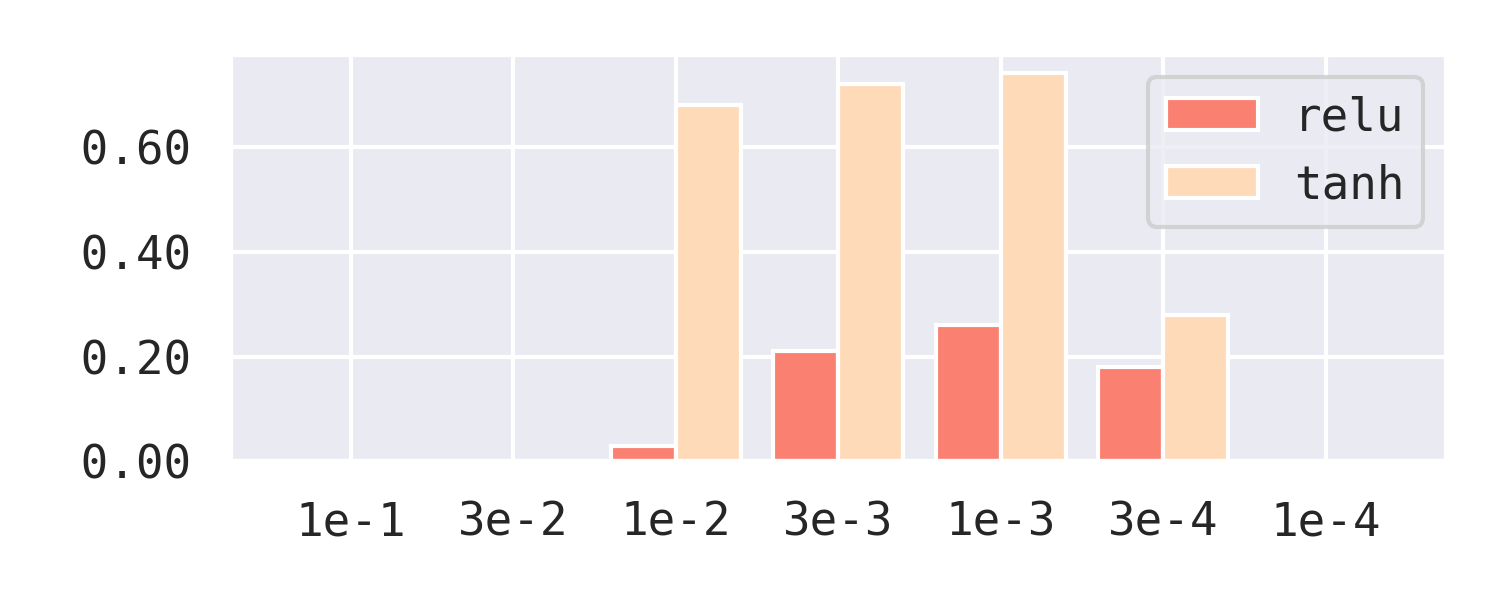}
        \caption{AdamW}
    \end{subfigure}
    \\
    \begin{subfigure}{.49\linewidth}
        \includegraphics[width=\linewidth]{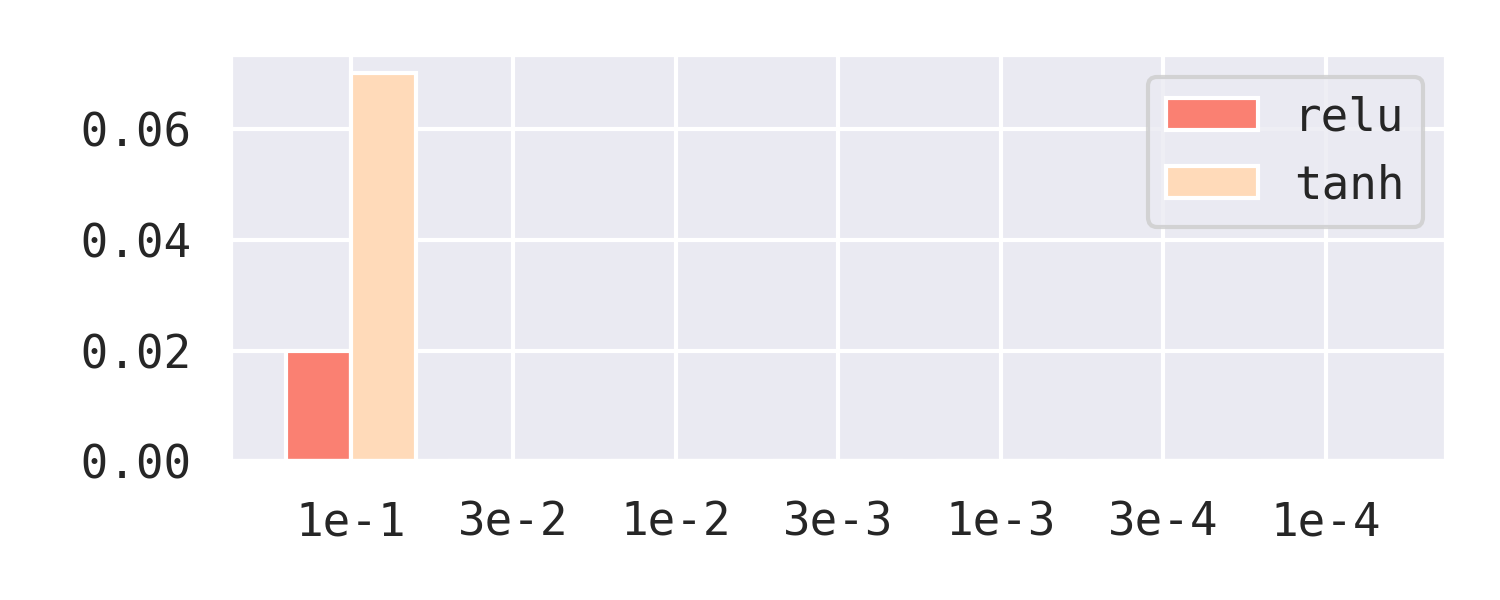}
        \caption{Ftrl}
    \end{subfigure}
    \begin{subfigure}{.49\linewidth}
        \includegraphics[width=\linewidth]{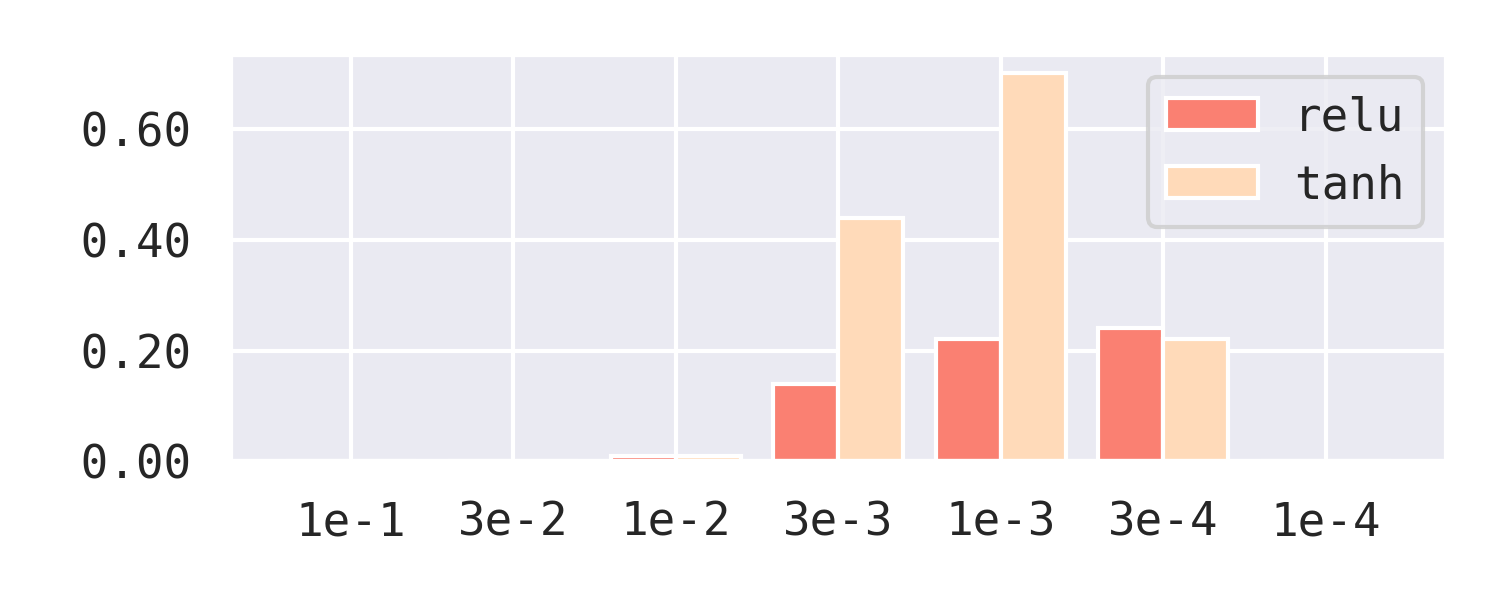}
        \caption{Nadam}
    \end{subfigure}
    \\
    \begin{subfigure}{.49\linewidth}
        \includegraphics[width=\linewidth]{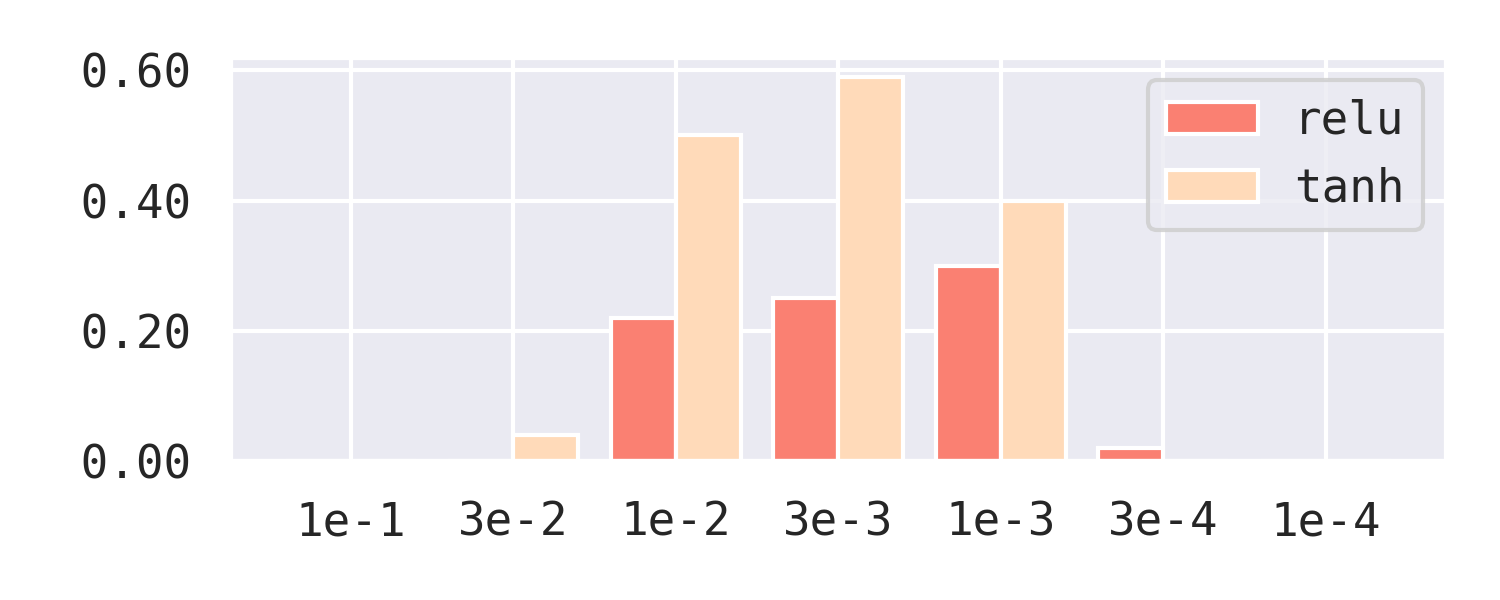}
        \caption{RMSprop}
    \end{subfigure}
    \begin{subfigure}{.49\linewidth}
        \includegraphics[width=\linewidth]{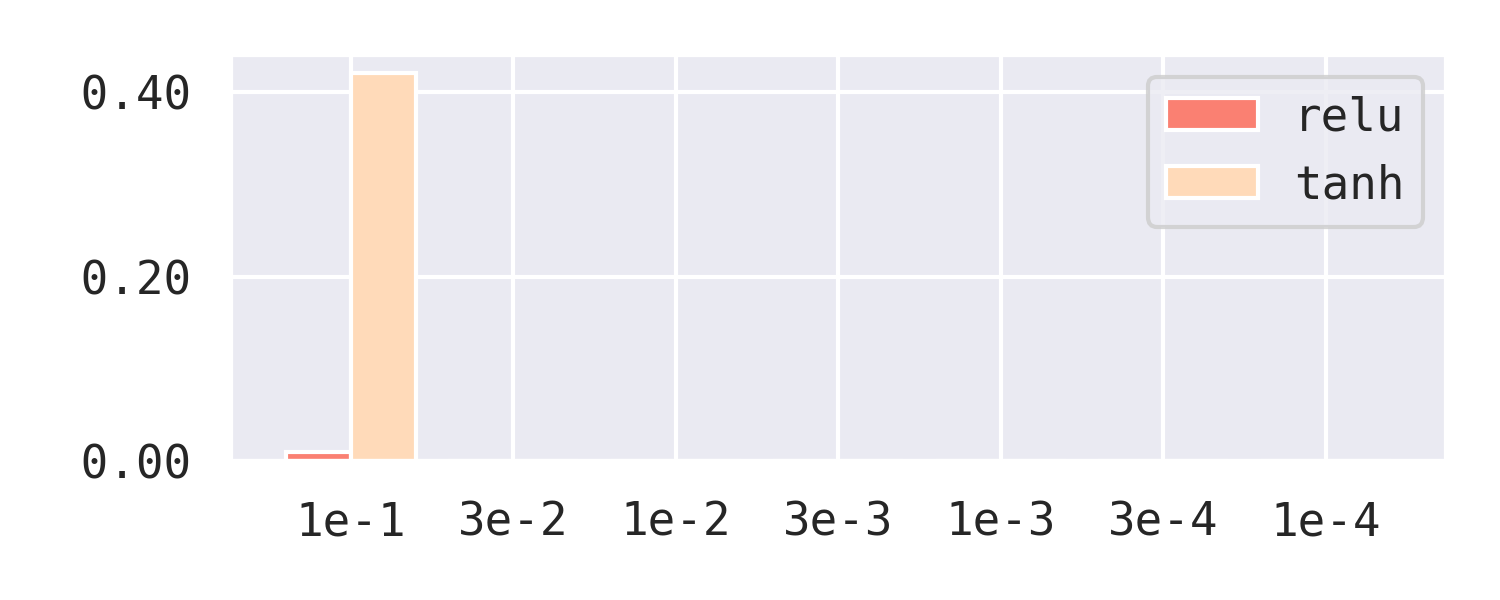}
        \caption{SGD}
    \end{subfigure}
    \caption{Hyperparameter search for fixed dataset on 2-step Game of Life with recursive network.}
    \label{fig:search_2_rec_fixed}
\end{figure*}

\begin{figure*}[ht!]
    \centering
    \begin{subfigure}{.49\linewidth}
        \includegraphics[width=\linewidth]{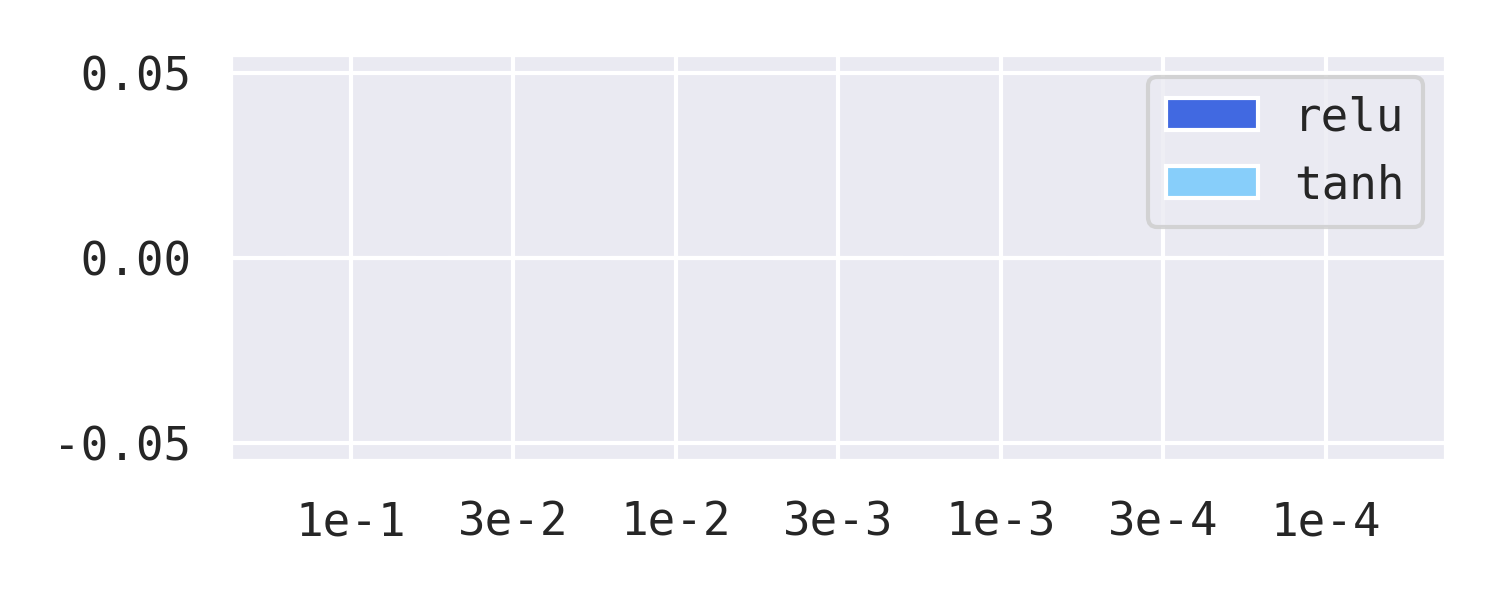}
        \caption{Adadelta}
    \end{subfigure}
    \begin{subfigure}{.49\linewidth}
        \includegraphics[width=\linewidth]{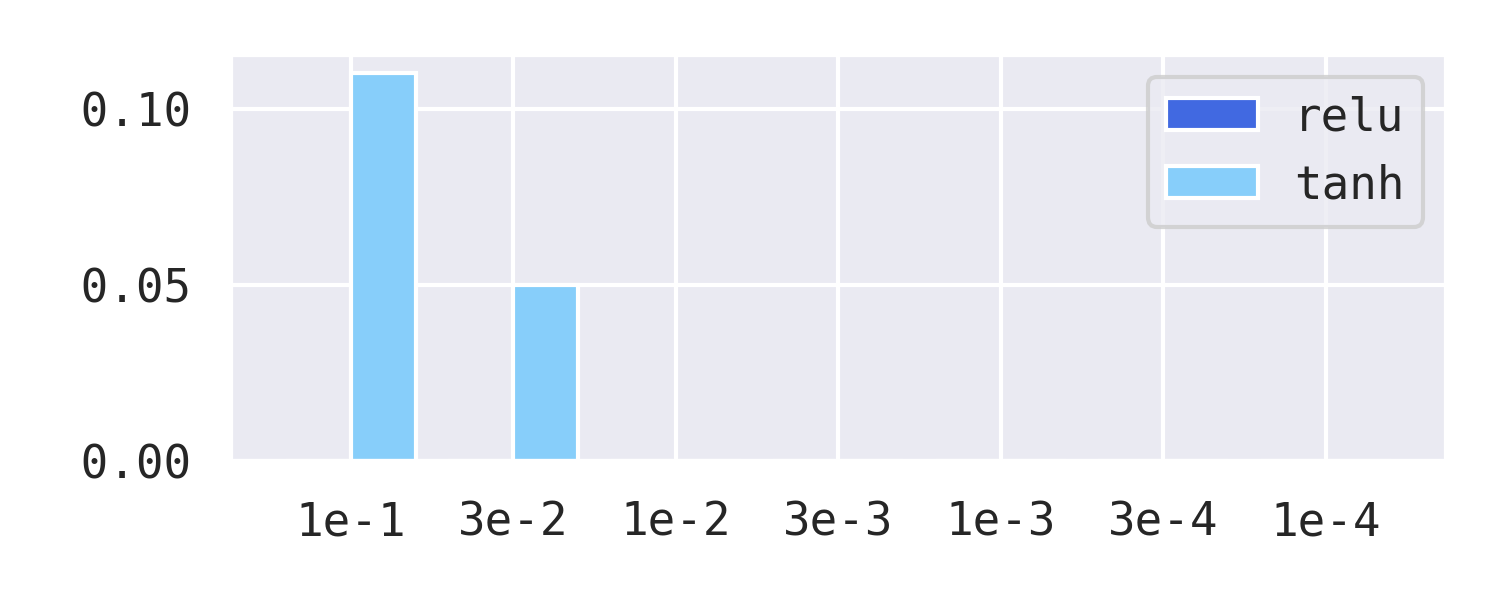}
        \caption{Adafactor}
    \end{subfigure}
    \\
    \begin{subfigure}{.49\linewidth}
        \includegraphics[width=\linewidth]{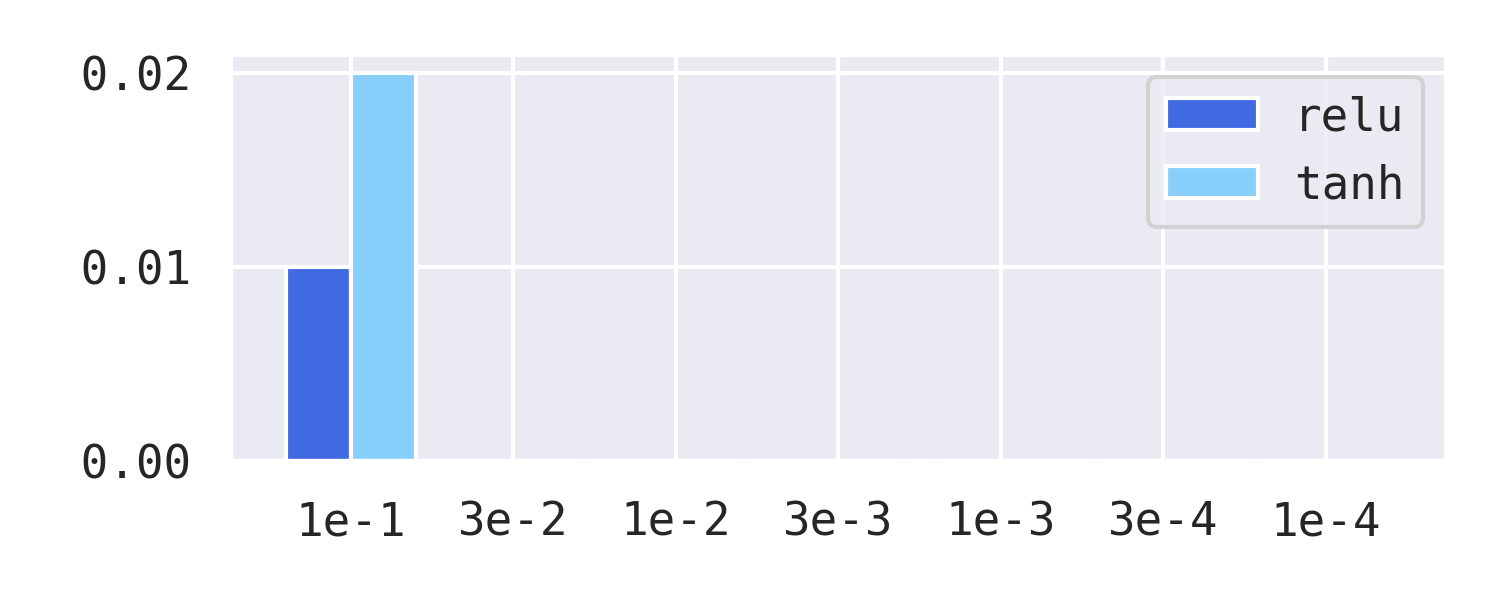}
        \caption{Adagrad}
    \end{subfigure}
    \begin{subfigure}{.49\linewidth}
        \includegraphics[width=\linewidth]{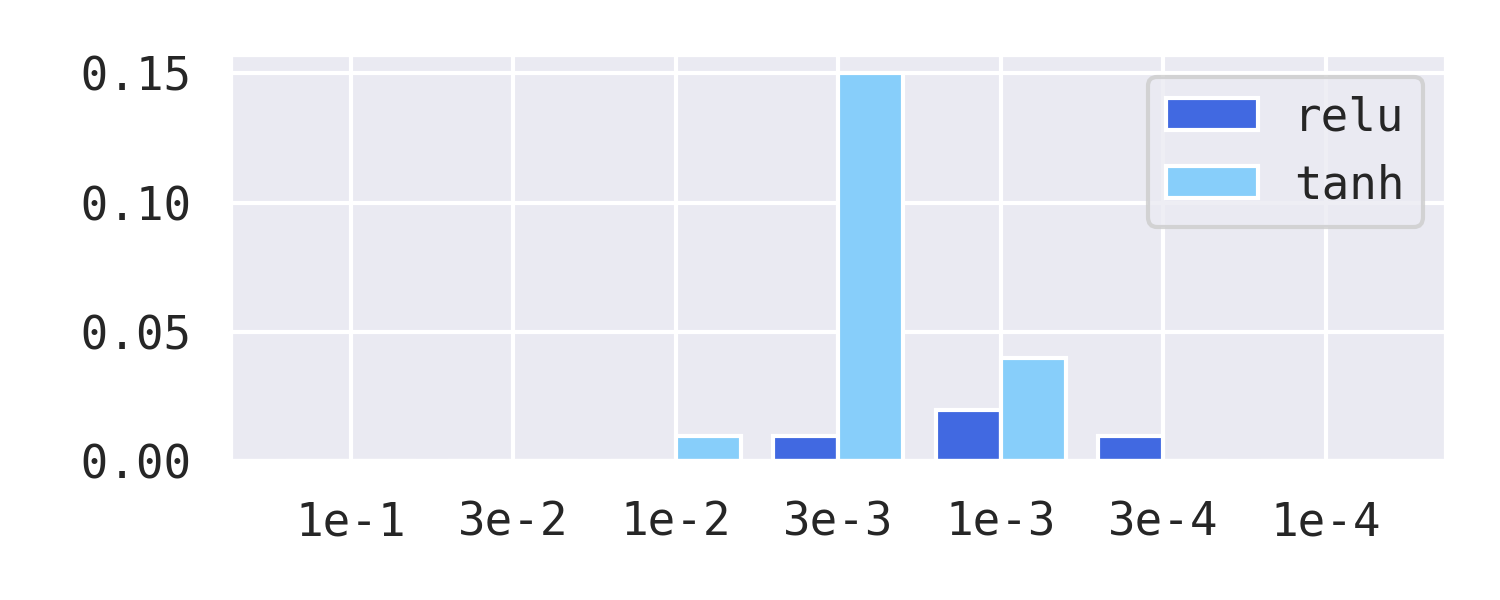}
        \caption{Adam}
    \end{subfigure}
    \\
    \begin{subfigure}{.49\linewidth}
        \includegraphics[width=\linewidth]{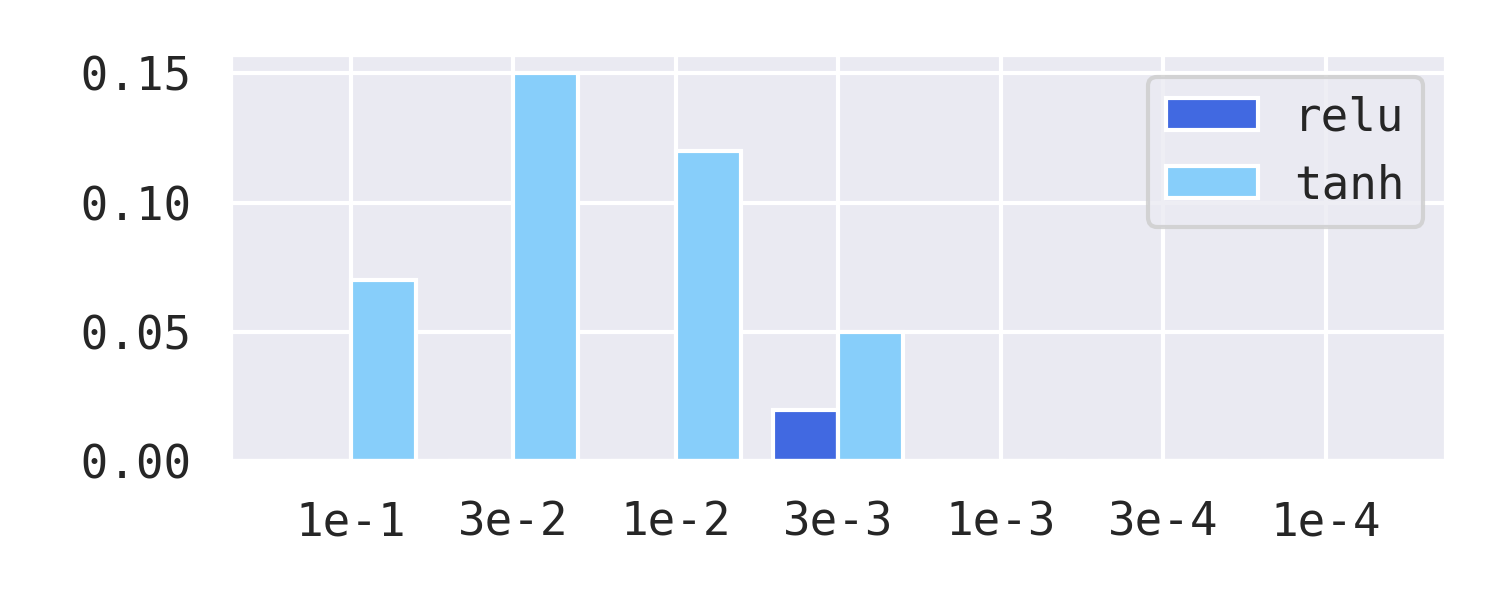}
        \caption{Adamax}
    \end{subfigure}
    \begin{subfigure}{.49\linewidth}
        \includegraphics[width=\linewidth]{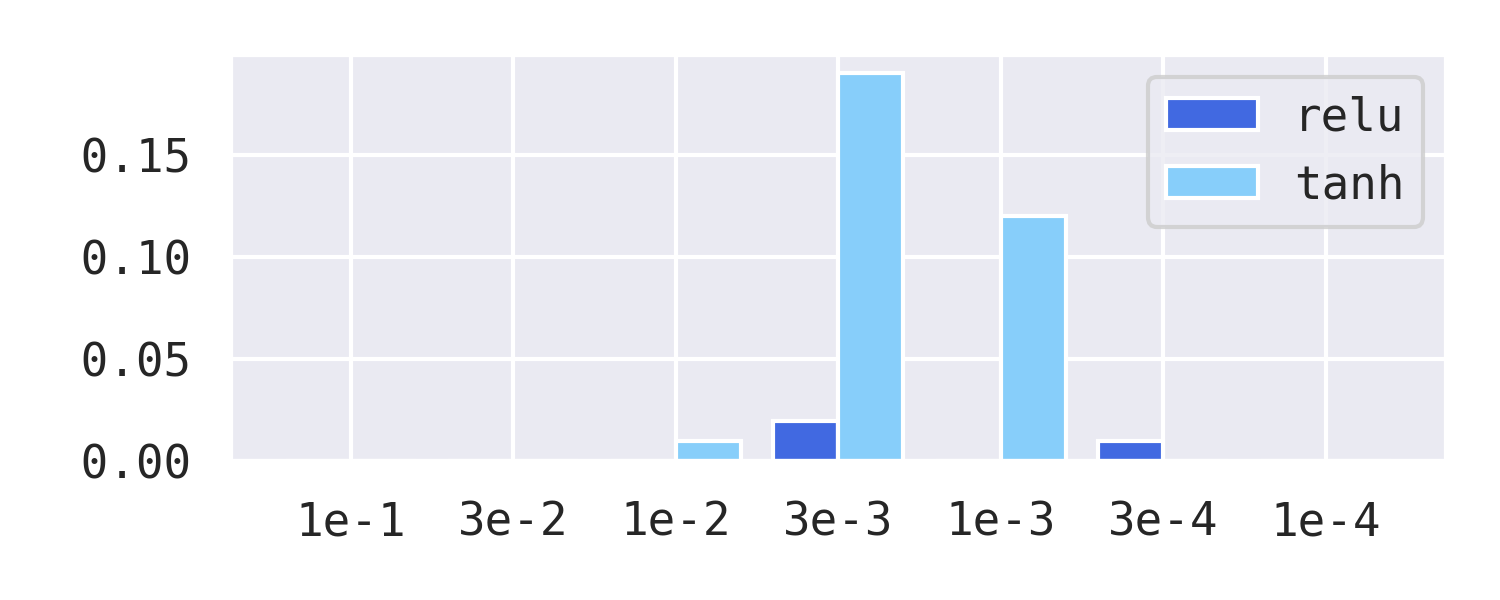}
        \caption{AdamW}
    \end{subfigure}
    \\
    \begin{subfigure}{.49\linewidth}
        \includegraphics[width=\linewidth]{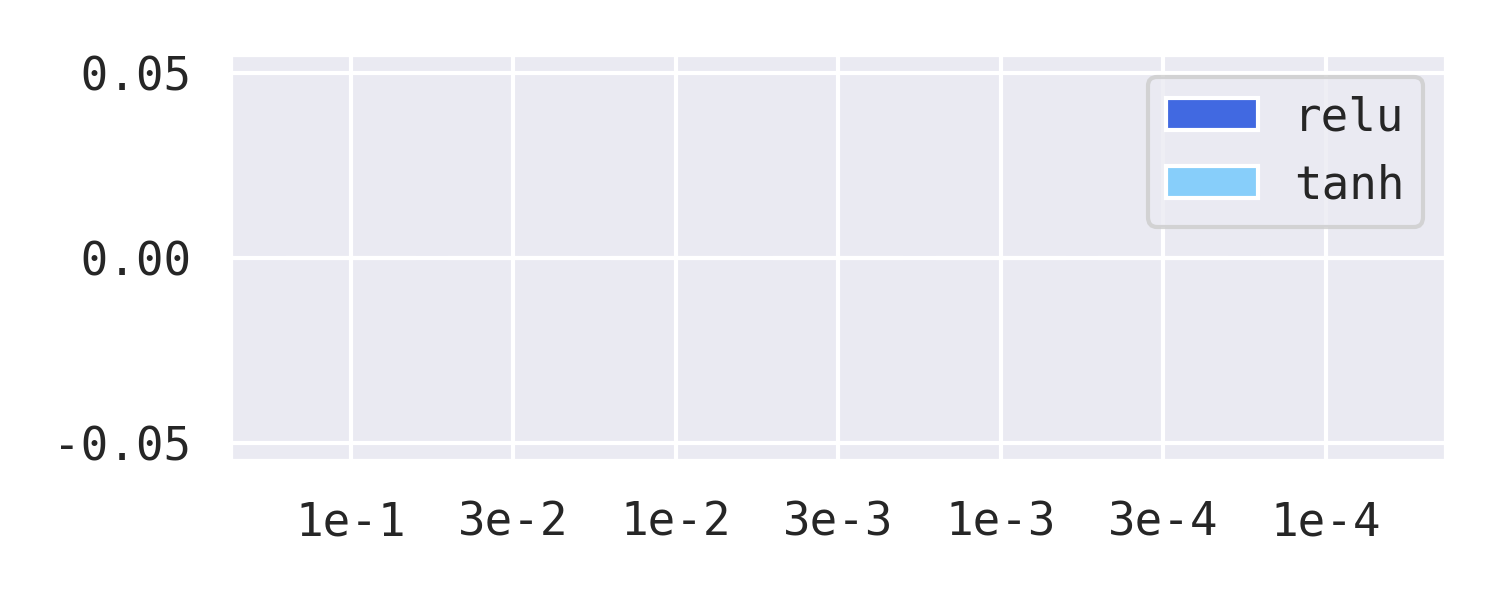}
        \caption{Ftrl}
    \end{subfigure}
    \begin{subfigure}{.49\linewidth}
        \includegraphics[width=\linewidth]{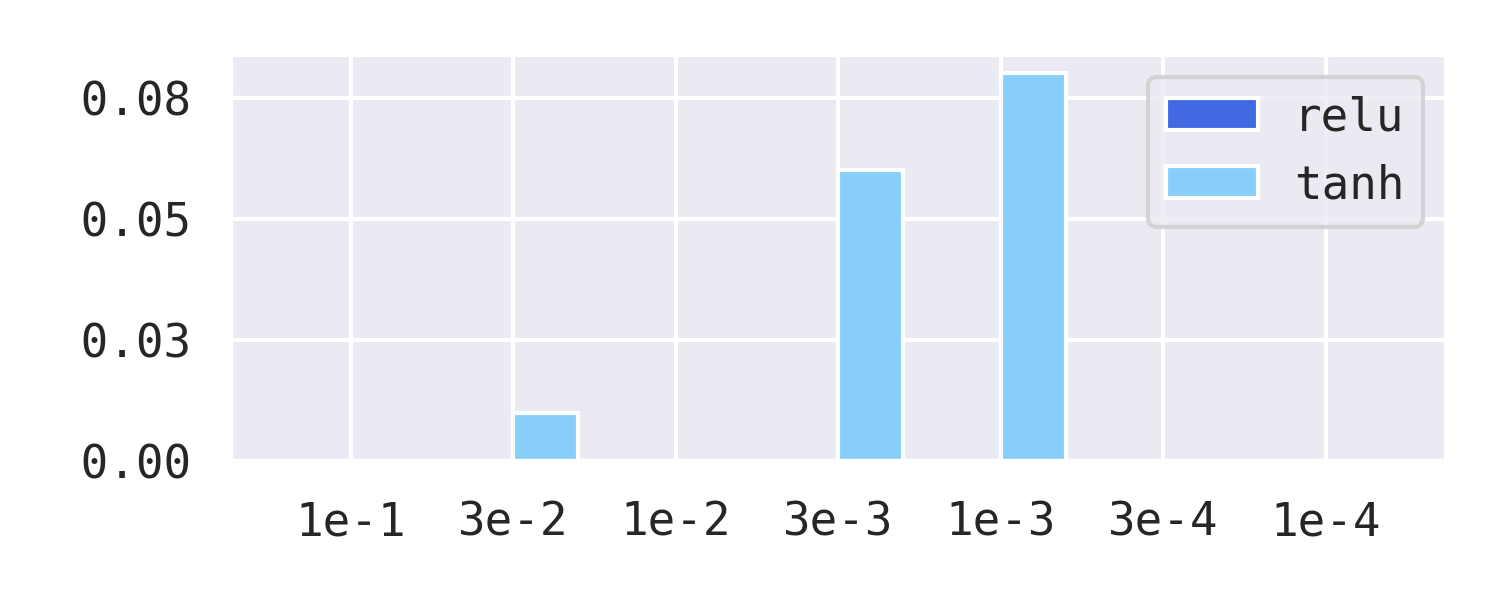}
        \caption{Nadam}
    \end{subfigure}
    \\
    \begin{subfigure}{.49\linewidth}
        \includegraphics[width=\linewidth]{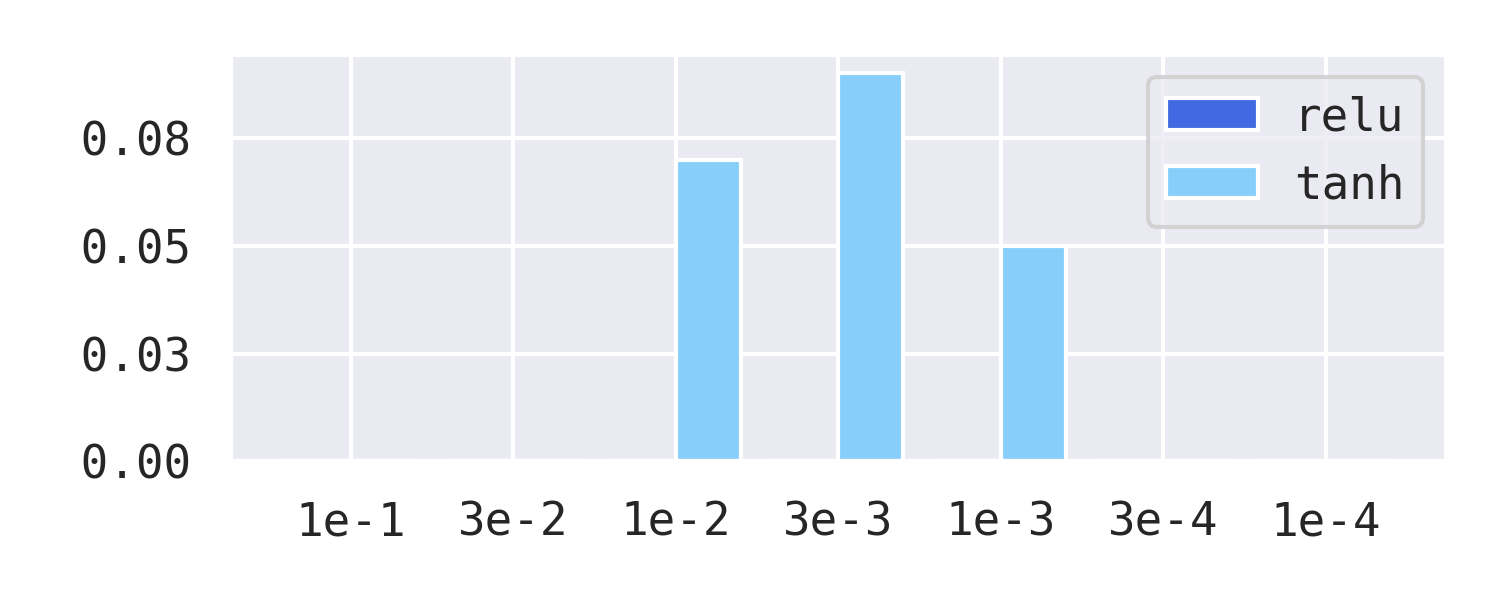}
        \caption{RMSprop}
    \end{subfigure}
    \begin{subfigure}{.49\linewidth}
        \includegraphics[width=\linewidth]{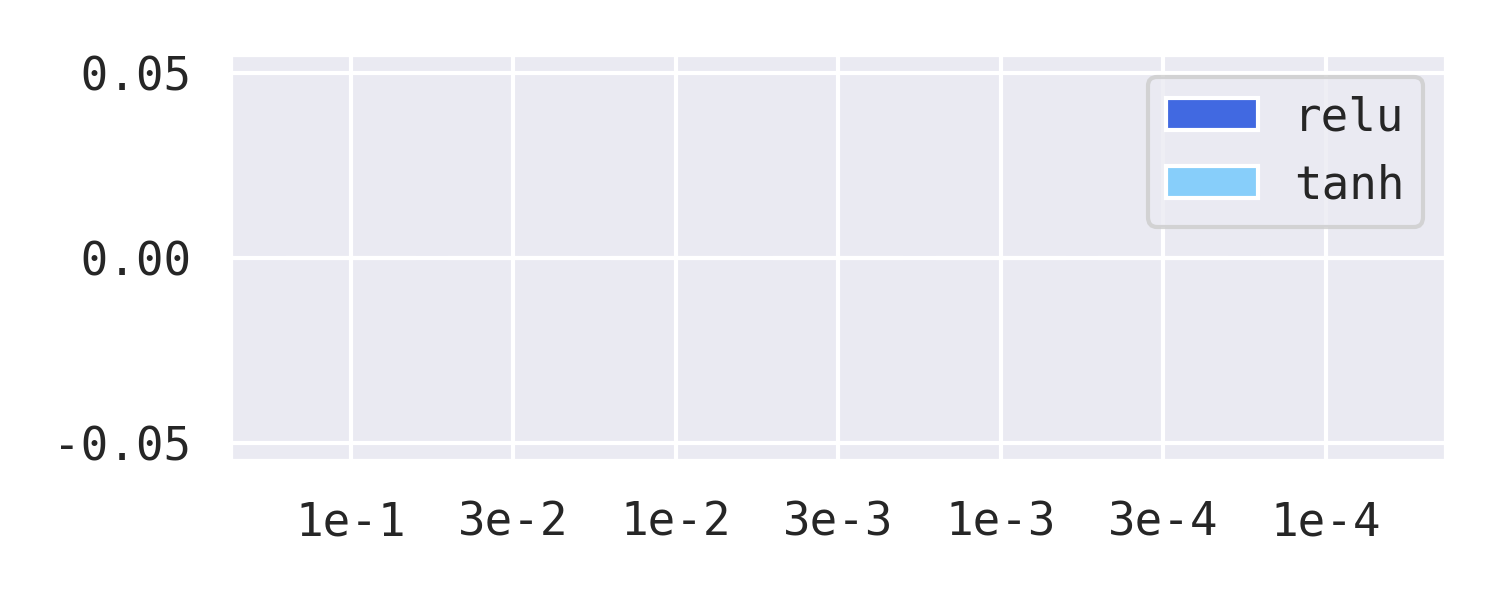}
        \caption{SGD}
    \end{subfigure}
    \caption{Hyperparameter search for random dataset on 2-step Game of Life with sequential network.}
    \label{fig:search_2_seq_random}
\end{figure*}

\begin{figure*}[ht!]
    \centering
    \begin{subfigure}{.49\linewidth}
        \includegraphics[width=\linewidth]{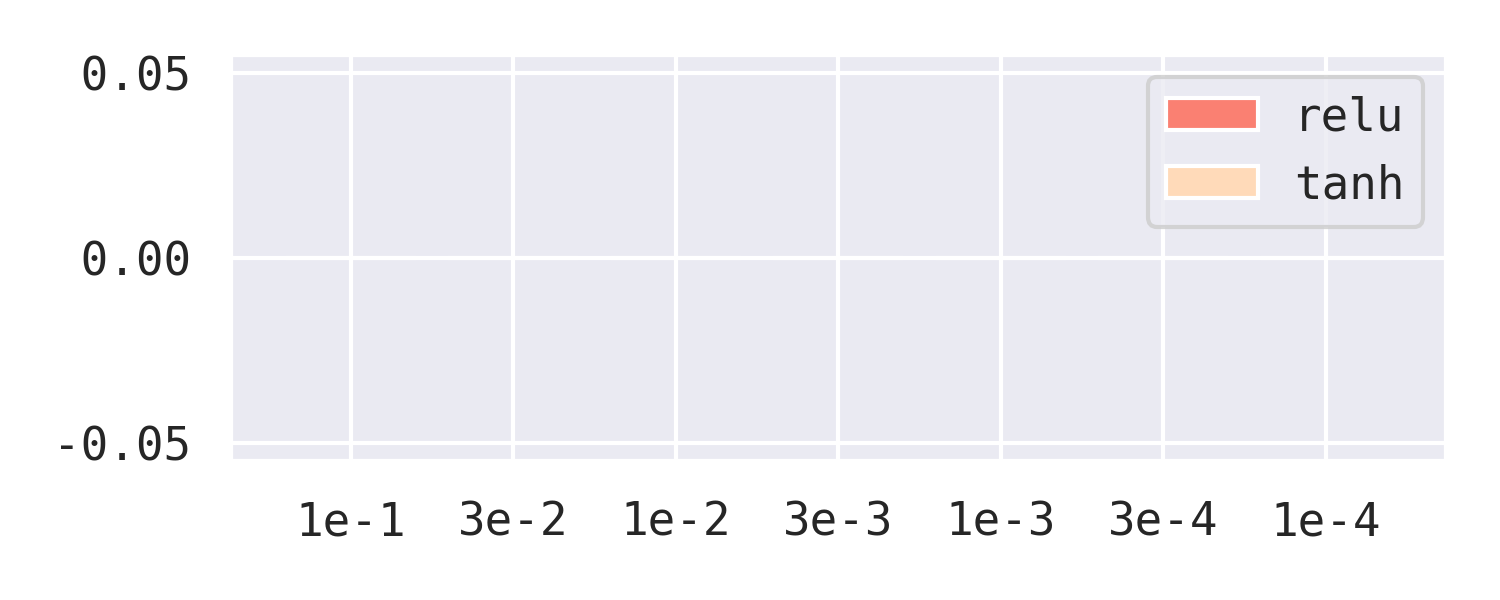}
        \caption{Adadelta}
    \end{subfigure}
    \begin{subfigure}{.49\linewidth}
        \includegraphics[width=\linewidth]{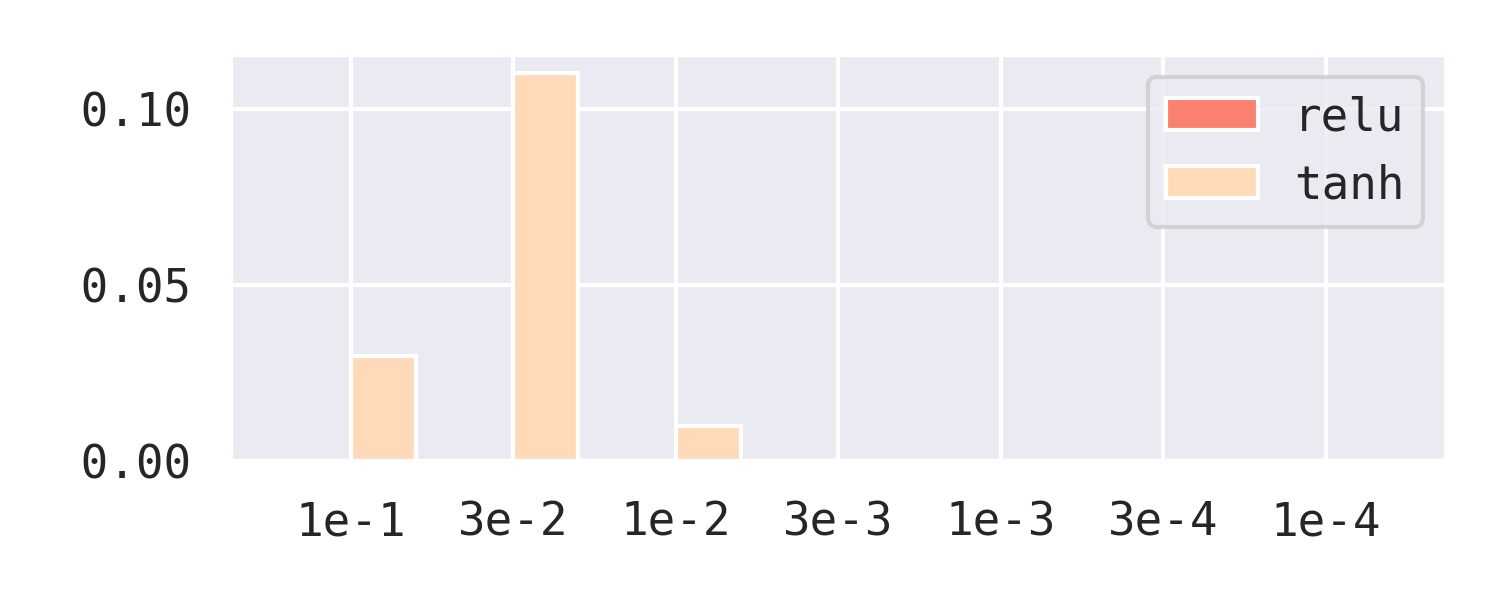}
        \caption{Adafactor}
    \end{subfigure}
    \\
    \begin{subfigure}{.49\linewidth}
        \includegraphics[width=\linewidth]{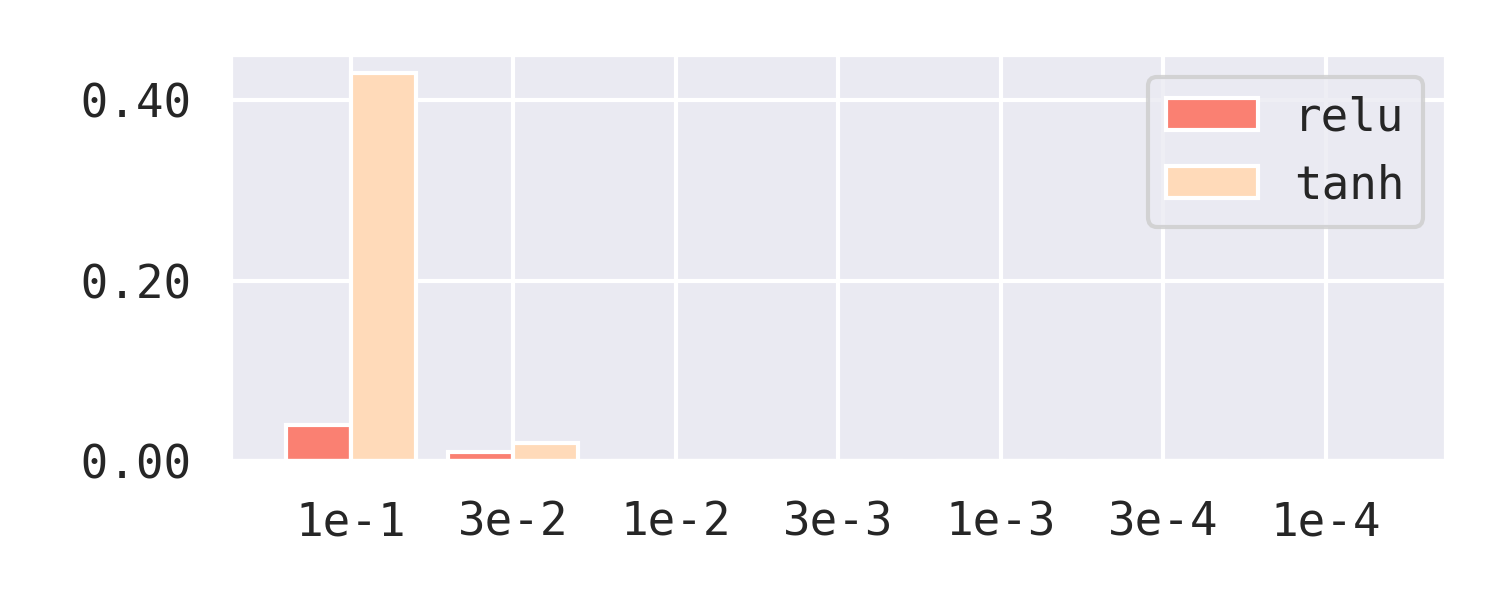}
        \caption{Adagrad}
    \end{subfigure}
    \begin{subfigure}{.49\linewidth}
        \includegraphics[width=\linewidth]{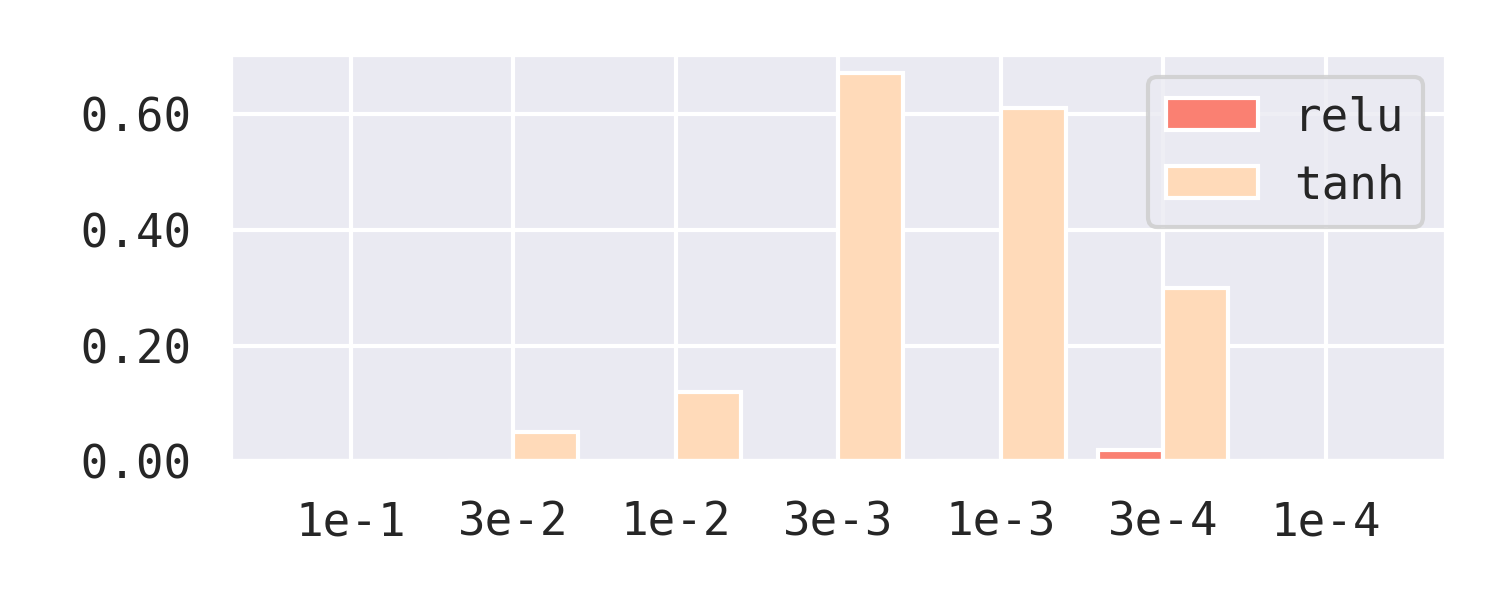}
        \caption{Adam}
    \end{subfigure}
    \\
    \begin{subfigure}{.49\linewidth}
        \includegraphics[width=\linewidth]{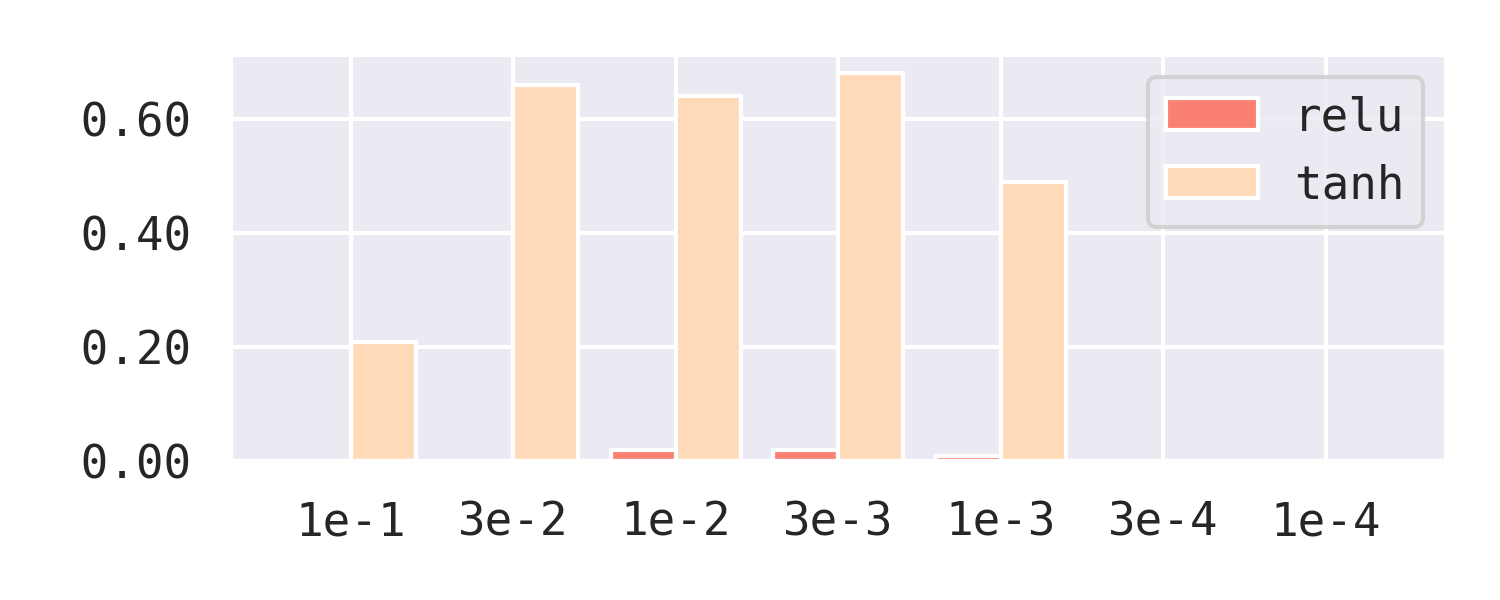}
        \caption{Adamax}
    \end{subfigure}
    \begin{subfigure}{.49\linewidth}
        \includegraphics[width=\linewidth]{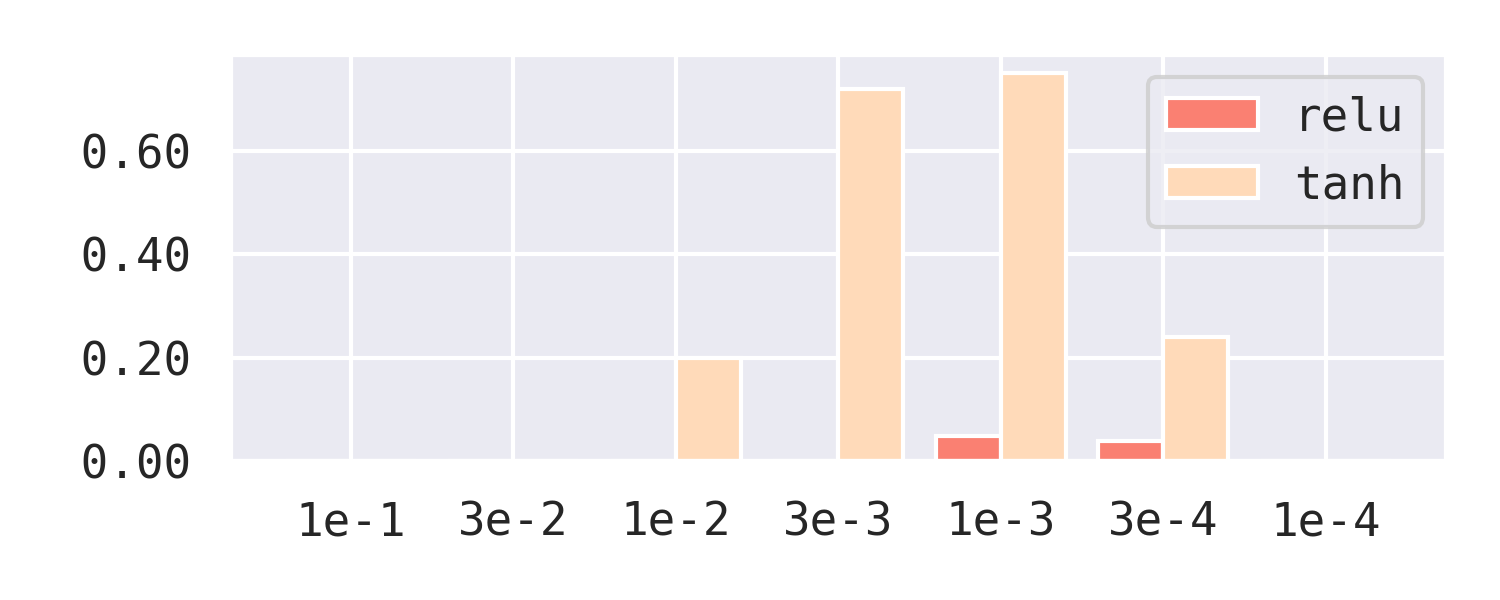}
        \caption{AdamW}
    \end{subfigure}
    \\
    \begin{subfigure}{.49\linewidth}
        \includegraphics[width=\linewidth]{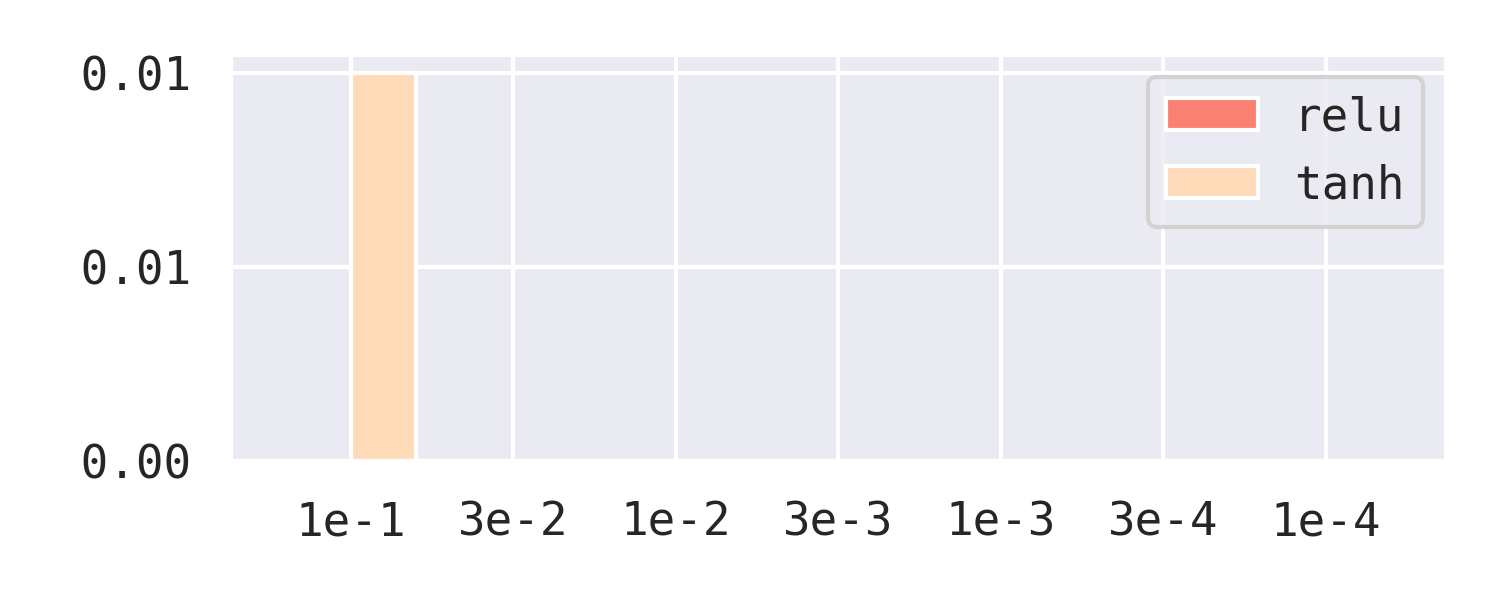}
        \caption{Ftrl}
    \end{subfigure}
    \begin{subfigure}{.49\linewidth}
        \includegraphics[width=\linewidth]{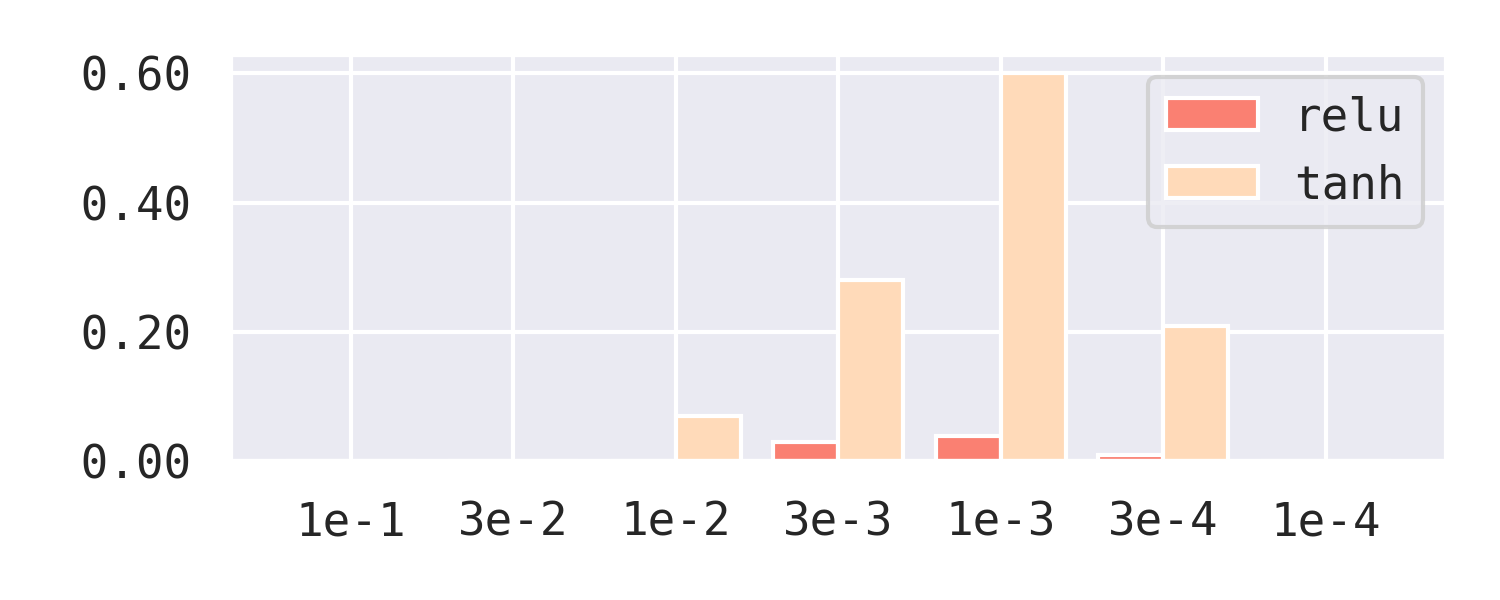}
        \caption{Nadam}
    \end{subfigure}
    \\
    \begin{subfigure}{.49\linewidth}
        \includegraphics[width=\linewidth]{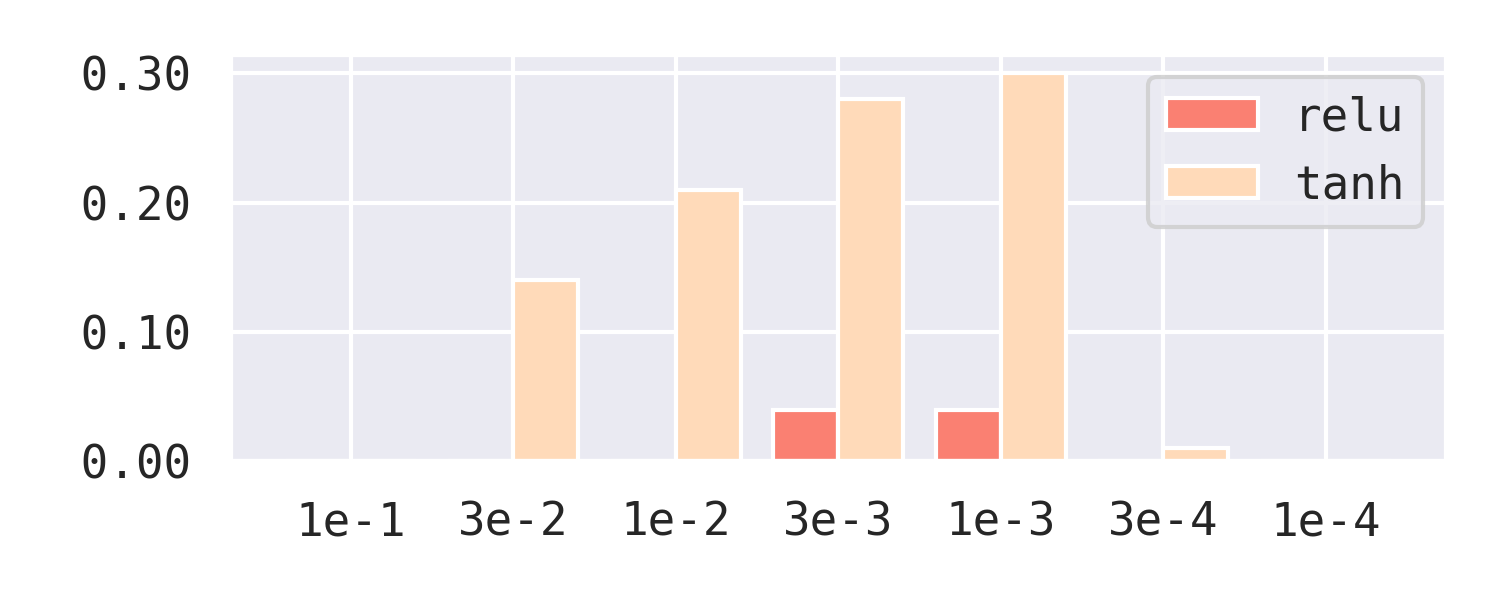}
        \caption{RMSprop}
    \end{subfigure}
    \begin{subfigure}{.49\linewidth}
        \includegraphics[width=\linewidth]{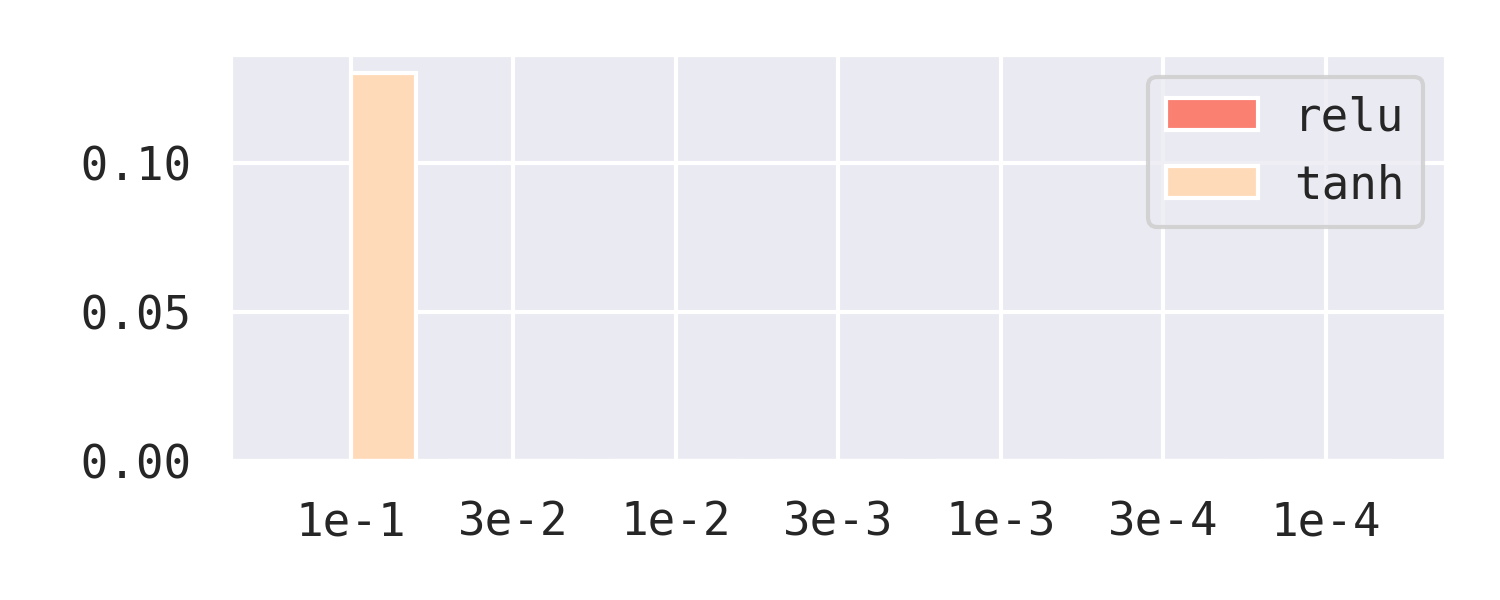}
        \caption{SGD}
    \end{subfigure}
    \caption{Hyperparameter search for fixed dataset on 2-step Game of Life with sequential network.}
    \label{fig:search_2_seq_fixed}
\end{figure*}

We note that, since the random dataset generation is seedable, the hyperparameter search is performed on exactly the same datasets that the algorithms are evaluated on.
The results of all tests are provided in Figures~\ref{fig:search_1_random}--\ref{fig:search_2_seq_fixed}.
The value of the learning rate $\lambda$ providing the highest convergence rate is selected for each algorithm/environment pair, see Tables~\ref{tab:search_1_step}, \ref{tab:search_2_step_rec}, and~\ref{tab:search_2_step_seq}, where the entry "\texttt{----}" indicates the absence of successful simulations for any of the values of $\lambda$.

\vspace*{4in}

\begin{table}[h]
    \caption{Hyperparameters for 1-step Game of Life.}
    \label{tab:search_1_step}
    \centering\small
    \begin{tabular}{lrrrr}
        \toprule
        & \multicolumn{2}{c}{Random dataset} & \multicolumn{2}{c}{Fixed dataset}
        \\\cmidrule(lr){2-3}\cmidrule(lr){4-5}
        algorithm & relu & tanh & relu & tanh
        \\\midrule
        Adadelta & \texttt{1e-1} & \texttt{1e-1} & \texttt{1e-1} & \texttt{1e-1}
        \\
        Adafactor & \texttt{3e-2} & \texttt{1e-2} & \texttt{3e-2} & \texttt{3e-2}
        \\
        Adagrad & \texttt{1e-1} & \texttt{3e-2} & \texttt{1e-1} & \texttt{1e-1}
        \\
        Adam & \texttt{1e-3} & \texttt{3e-2} & \texttt{3e-3} & \texttt{3e-2}
        \\
        Adamax & \texttt{3e-4} & \texttt{3e-2} & \texttt{1e-2} & \texttt{1e-3}
        \\
        AdamW & \texttt{3e-4} & \texttt{1e-2} & \texttt{1e-2} & \texttt{1e-1}
        \\
        Ftrl & \texttt{1e-1} & \texttt{1e-1} & \texttt{1e-1} & \texttt{1e-1}
        \\
        Nadam & \texttt{1e-2} & \texttt{1e-3} & \texttt{1e-3} & \texttt{1e-2}
        \\
        RMSprop & \texttt{3e-3} & \texttt{3e-2} & \texttt{1e-2} & \texttt{1e-2}
        \\
        SGD & \texttt{3e-2} & \texttt{1e-1} & \texttt{1e-1} & \texttt{1e-1}
        \\\bottomrule
    \end{tabular}
\end{table}

\begin{table}[h]
    \caption{Hyperparameters for 2-step Game of Life with recursive network.}
    \label{tab:search_2_step_rec}
    \centering\small
    \begin{tabular}{lrrrr}
        \toprule
        & \multicolumn{2}{c}{Random dataset} & \multicolumn{2}{c}{Fixed dataset}
        \\\cmidrule(lr){2-3}\cmidrule(lr){4-5}
        algorithm & relu & tanh & relu & tanh
        \\\midrule
        Adadelta & \texttt{----} & \texttt{----} & \texttt{1e-1} & \texttt{1e-1}
        \\
        Adafactor & \texttt{3e-2} & \texttt{3e-2} & \texttt{3e-2} & \texttt{1e-1}
        \\
        Adagrad & \texttt{1e-1} & \texttt{1e-1} & \texttt{1e-1} & \texttt{1e-1}
        \\
        Adam & \texttt{1e-3} & \texttt{3e-3} & \texttt{3e-4} & \texttt{1e-3}
        \\
        Adamax & \texttt{1e-2} & \texttt{1e-2} & \texttt{1e-3} & \texttt{1e-3}
        \\
        AdamW & \texttt{3e-3} & \texttt{1e-2} & \texttt{1e-3} & \texttt{1e-3}
        \\
        Ftrl & \texttt{----} & \texttt{----} & \texttt{1e-1} & \texttt{1e-1}
        \\
        Nadam & \texttt{1e-3} & \texttt{3e-3} & \texttt{3e-4} & \texttt{1e-3}
        \\
        RMSprop & \texttt{1e-3} & \texttt{3e-3} & \texttt{1e-3} & \texttt{3e-3}
        \\
        SGD & \texttt{----} & \texttt{----} & \texttt{1e-1} & \texttt{1e-1}
        \\\bottomrule
    \end{tabular}
\end{table}

\begin{table}[h]
    \caption{Hyperparameters for 2-step Game of Life with sequential network.}
    \label{tab:search_2_step_seq}
    \centering\small
    \begin{tabular}{lrrrr}
        \toprule
        & \multicolumn{2}{c}{Random dataset} & \multicolumn{2}{c}{Fixed dataset}
        \\\cmidrule(lr){2-3}\cmidrule(lr){4-5}
        algorithm & relu & tanh & relu & tanh
        \\\midrule
        Adadelta & \texttt{----} & \texttt{----} & \texttt{----} & \texttt{----}
        \\
        Adafactor & \texttt{----} & \texttt{1e-1} & \texttt{----} & \texttt{3e-2}
        \\
        Adagrad & \texttt{1e-1} & \texttt{1e-1} & \texttt{1e-1} & \texttt{1e-1}
        \\
        Adam & \texttt{1e-3} & \texttt{3e-3} & \texttt{3e-4} & \texttt{3e-3}
        \\
        Adamax & \texttt{3e-3} & \texttt{3e-2} & \texttt{3e-3} & \texttt{3e-3}
        \\
        AdamW & \texttt{3e-3} & \texttt{3e-3} & \texttt{1e-3} & \texttt{1e-3}
        \\
        Ftrl & \texttt{----} & \texttt{----} & \texttt{----} & \texttt{1e-1}
        \\
        Nadam & \texttt{----} & \texttt{1e-3} & \texttt{1e-3} & \texttt{1e-3}
        \\
        RMSprop & \texttt{----} & \texttt{3e-3} & \texttt{1e-3} & \texttt{1e-3}
        \\
        SGD & \texttt{----} & \texttt{----} & \texttt{----} & \texttt{1e-1}
        \\\bottomrule
    \end{tabular}
\end{table}

\clearpage

\clearpage

\clearpage

\clearpage

\clearpage

\clearpage

\end{document}